\pdfoutput=1
\PassOptionsToPackage{verbose}{wrapfig}

\documentclass[10pt,logo,copyright]{nvidia_release_template/nvidia_report}

\usepackage[numbers,sort&compress]{natbib}
\usepackage{environ}

\definecolor{weakgray}{HTML}{888888}
\definecolor{stronggreen}{HTML}{76B900}

\usepackage{cuted} %
\usepackage{currfile} %
\usepackage{caption} %

\usepackage[normalem]{ulem}
\usepackage{adjustbox} %

\newcommand{\supp}[1]{Appendix~\ref{#1}}

\usepackage{multirow}
\usepackage{enumitem} %
\usepackage{colortbl} %
\usepackage{float}
\usepackage{wrapfig}
\usepackage{tikz}
\usepackage{sansmath}
\usetikzlibrary{arrows.meta,positioning,shapes.geometric,decorations.pathreplacing}
\definecolor{nvidiagreen}{HTML}{76B900}
\definecolor{figimgA}{HTML}{bc80bd} %
\definecolor{figclsLynx}{HTML}{fdb462} %

\definecolor{nvgreenlight}{HTML}{E2F2C6}
\newcommand{\oursrow}{\rowcolor{nvgreenlight}}
\newcommand{\grouprow}{\rowcolor{black!9}}
\usepackage{soul}
\colorlet{hlgray}{black!9}
\DeclareRobustCommand{\hlc}[2]{{\sethlcolor{#1}\hl{#2}}}
\DeclareRobustCommand{\csq}[1]{{\setlength{\fboxsep}{0pt}\colorbox{#1}{\rule{0pt}{1.3ex}\rule{1.3ex}{0pt}}}}

\newcommand{\bestv}[1]{\textbf{#1}}
\newcommand{\secondv}[1]{\underline{#1}}

\definecolor{jcpurple}{RGB}{125,40,200}
\usepackage[normalem]{ulem}

\colorlet{densec}{gray!70}
\colorlet{moec}{nvidiagreen}
\colorlet{repac}{blue!60}
\colorlet{norepac}{orange!80}

\title{Scaling Mixture-of-Experts Image Generators with Better Representations}

\author[1,2]{Junchen~Liu}
\author[1]{Wei~Xiong}
\author[1]{Ruofan~Liang}
\author[2]{Xuanchi~Ren}
\author[2]{Zian~Wang}
\author[2]{Yifan~Lu}
\author[2]{Sanja~Fidler}
\author[1]{Jialiang~Wang}
\author[1]{Shalini~De~Mello}
\author[1]{Haithem~Turki}
\affil[1]{NVIDIA}
\affil[2]{University of Toronto}
\renewcommand{\today}{2026-11-30}
\hypersetup{pdftitle={Scaling Mixture-of-Experts Image Generators with Better Representations},
  pdfauthor={Junchen Liu, Wei Xiong, Ruofan Liang, Xuanchi Ren, Zian Wang, Yifan Lu, Sanja Fidler, Jialiang Wang, Shalini De Mello, Haithem Turki}}

\titleformat{\paragraph}[runin]{\normalfont\normalsize\bfseries}{}{0pt}{#1}
\titlespacing*{\paragraph}{0pt}{1.2ex plus .5ex minus .2ex}{1em}

\begin{document}
\makeatletter
\newlength{\nvcolw}
\setlength{\nvcolw}{0.5\textwidth}
\newif\ifnvnarrow
\nvnarrowtrue
\let\nv@origxfloat\@xfloat
\def\nv@H{H}
\def\@xfloat#1[#2]{\def\nv@spec{#2}%
  \ifx\nv@spec\nv@H \nv@origxfloat{#1}[H]%
  \else\ifnvnarrow \edef\nv@tmp{\noexpand\nv@origxfloat{#1}[\nv@fspec]}\nv@tmp\else \nv@origxfloat{#1}[#2]\fi\fi}
\gdef\nv@fspec{!tb}
\let\nvorigfigure\figure
\let\endnvorigfigure\endfigure
\let\nvorigtable\table
\let\endnvorigtable\endtable
\ExplSyntaxOn
\cs_new_protected:Npn \nvrestoreeverypar { \tex_everypar:D { \g__para_standard_everypar_tl } }
\box_new:N \l_nv_indent_box
\cs_new_protected:Npn \nvsaveindent { \box_set_eq:NN \l_nv_indent_box \g_para_indent_box }
\cs_new_protected:Npn \nvrestoreindent { \box_gset_eq:NN \g_para_indent_box \l_nv_indent_box }
\ExplSyntaxOff
\newcount\nv@qhead
\newcount\nv@qtail
\newif\ifnv@heading
\newcounter{nvfl}
\newcounter{nvfigureseen}\setcounter{nvfigureseen}{1}
\newcounter{nvtableseen}
\newcommand{\nvfloatact}[2]{\expandafter\def\csname nv@act@#1\endcsname{#2}}
\def\nv@s@hold{hold}
\def\nv@s@pair{pair}
\def\nv@s@table{table}
\def\nv@s@afterpair{afterpair}
\def\nv@s@later{later}
\newcount\nv@hn
\newcount\nv@hi
\newdimen\nv@pairh
\newsavebox\nv@lbox
\newsavebox\nv@rbox
\newlength{\nvpairL}\setlength{\nvpairL}{0.42\textwidth}
\newlength{\nvpairR}\setlength{\nvpairR}{0.54\textwidth}
\setlength{\columnsep}{16pt}
\newif\ifnv@haslater
\def\nv@s@right{right}
\def\nv@s@pairlast{pairlast}
\newcommand{\nv@addheld}[1]{%
  \global\advance\nv@hn\@ne
  \global\expandafter\let\csname nv@h@body@\the\nv@hn\endcsname\BODY
  \expandafter\xdef\csname nv@h@type@\the\nv@hn\endcsname{#1}%
  \expandafter\xdef\csname nv@h@num@\the\nv@hn\endcsname{\the\numexpr\value{nv#1seen}-1\relax}}
\newcommand{\nv@setright}[1]{%
  \global\let\nv@rbody\BODY
  \xdef\nv@rtype{#1}%
  \xdef\nv@rnum{\the\numexpr\value{nv#1seen}-1\relax}}
\newcommand{\nv@emitblock}{%
  \nvorigfigure[!tb]%
    \sbox\nv@rbox{\begin{minipage}[t]{\nvpairR}\vspace{0pt}\setlength{\columnwidth}{\linewidth}%
      \AddToHookNext{env/tabular/begin}{\def\arraystretch{1.00}}%
      \setcounter{\nv@rtype}{\nv@rnum}\let\@captype\nv@rtype\nv@rbody\end{minipage}}%
    \nv@pairh=\dimexpr\ht\nv@rbox+\dp\nv@rbox\relax
    \sbox\nv@lbox{\begin{minipage}[t]{\nvpairL}\vspace{0pt}\setlength{\columnwidth}{\linewidth}%
      \nv@hi=\z@
      \loop\ifnum\nv@hi<\nv@hn
        \advance\nv@hi\@ne
        \ifnum\nv@hi>\@ne \par\vskip 6pt \fi
        \begingroup
        \edef\nv@t{\csname nv@h@type@\the\nv@hi\endcsname}%
        \setcounter{\nv@t}{\csname nv@h@num@\the\nv@hi\endcsname}%
        \let\@captype\nv@t
        \csname nv@h@body@\the\nv@hi\endcsname\par
        \endgroup
      \repeat
      \end{minipage}}%
    \ifdim\dimexpr\ht\nv@lbox+\dp\nv@lbox\relax<\nv@pairh
      \sbox\nv@lbox{\begin{minipage}[t][\nv@pairh][s]{\nvpairL}\vspace{0pt}\setlength{\columnwidth}{\linewidth}%
        \nv@hi=\z@
        \loop\ifnum\nv@hi<\nv@hn
          \advance\nv@hi\@ne
          \ifnum\nv@hi>\@ne \par\vfill \fi
          \begingroup
          \edef\nv@t{\csname nv@h@type@\the\nv@hi\endcsname}%
          \setcounter{\nv@t}{\csname nv@h@num@\the\nv@hi\endcsname}%
          \let\@captype\nv@t
          \csname nv@h@body@\the\nv@hi\endcsname\par
          \endgroup
        \repeat
        \end{minipage}}%
    \fi
    \typeout{NVQ block height \the\nv@pairh}%
    \noindent\box\nv@lbox\hfill\box\nv@rbox\label{nv:pairblock}%
  \endnvorigfigure
  \global\nv@hn=\z@}
\newcommand{\nv@dofloat}[1]{%
  \stepcounter{nvfl}\stepcounter{nv#1seen}%
  \@ifundefined{nv@act@\arabic{nvfl}}{\def\nv@a{wrap}}{\edef\nv@a{\csname nv@act@\arabic{nvfl}\endcsname}}%
  \ifx\nv@a\nv@s@hold
    \nv@addheld{#1}%
  \else\ifx\nv@a\nv@s@right
    \nv@setright{#1}%
  \else\ifx\nv@a\nv@s@pairlast
    \nv@addheld{#1}\nv@emitblock
  \else\ifx\nv@a\nv@s@pair
    \nv@setright{#1}\nv@emitblock
  \else\ifx\nv@a\nv@s@later
    \global\let\nv@lbody\BODY
    \xdef\nv@ltype{#1}%
    \xdef\nv@lnum{\the\numexpr\value{nv#1seen}-1\relax}%
    \global\nv@haslatertrue
  \else
    \global\advance\nv@qtail\@ne
    \global\expandafter\let\csname nv@q@body@\the\nv@qtail\endcsname\BODY
    \expandafter\gdef\csname nv@q@type@\the\nv@qtail\endcsname{#1}%
    \expandafter\xdef\csname nv@q@num@\the\nv@qtail\endcsname{\the\numexpr\value{nv#1seen}-1\relax}%
    \ifx\nv@a\nv@s@afterpair \expandafter\gdef\csname nv@q@ap@\the\nv@qtail\endcsname{1}\fi
    \ifnv@haslater
      \global\nv@haslaterfalse
      \global\advance\nv@qtail\@ne
      \global\expandafter\let\csname nv@q@body@\the\nv@qtail\endcsname\nv@lbody
      \expandafter\xdef\csname nv@q@type@\the\nv@qtail\endcsname{\nv@ltype}%
      \expandafter\xdef\csname nv@q@num@\the\nv@qtail\endcsname{\nv@lnum}%
    \fi
  \fi\fi\fi\fi\fi}
\AddToHook{env/figure*/begin}{\stepcounter{nvfigureseen}\setcounter{figure}{\numexpr\value{nvfigureseen}-1\relax}}
\AddToHook{env/table*/before}{\ifnvnarrow\suppressfloats[t]\gdef\nv@fspec{t}\fi}
\AddToHook{env/table*/after}{\gdef\nv@fspec{!tb}}
\AddToHook{env/table*/begin}{\stepcounter{nvtableseen}\setcounter{table}{\numexpr\value{nvtableseen}-1\relax}}
\newcommand{\nv@emitone}{%
  \global\advance\nv@qhead\@ne
  \typeout{NVQ emit \the\nv@qhead\space page \thepage\space pagetotal \the\pagetotal}%
  \nvsaveindent
  \begingroup
  \nvrestoreeverypar
  \edef\nv@t{\csname nv@q@type@\the\nv@qhead\endcsname}%
  \setcounter{\nv@t}{\csname nv@q@num@\the\nv@qhead\endcsname}%
  \csname wrap\nv@t\endcsname{R}{\nvcolw}\setlength{\columnwidth}{\nvcolw}%
  \csname nv@q@body@\the\nv@qhead\endcsname
  \ifx\nv@t\nv@s@table \par\vskip 4pt \fi
  \csname endwrap\nv@t\endcsname
  \typeout{NVQ size \the\WF@size}%
  \endgroup
  \nvrestoreindent}
\newcommand{\nv@tryemit}{%
  \ifnvnarrow\ifinner\else
    \ifnv@heading
    \else
      \ifnum\nv@qhead<\nv@qtail
        \@ifundefined{nv@q@ap@\the\numexpr\nv@qhead+1\relax}{\def\nv@minpage{0}}{\edef\nv@minpage{\getpagerefnumber{nv:pairblock}}}%
        \ifnum\c@page<\nv@minpage\relax\else
        \ifnum\c@WF@wrappedlines<\tw@ \ifvoid\WF@box \ifnum\@listdepth=\z@
          \nv@emitone
        \fi\fi\fi
      \fi\fi
    \fi
  \fi\fi}
\AddToHook{para/before}{\nv@tryemit}
\newcommand{\nvflushfloats}{%
  \par
  \ifvoid\WF@box\else
    \typeout{NVQ WARNING: pending wrap float set as a normal float}%
    \nvorigfigure[!tb]\centering\leavevmode\box\WF@box\endnvorigfigure
    \global\c@WF@wrappedlines\z@
    \ifx\everypar\WF@everypar \let\everypar\WF@@everypar \everypar\expandafter{\the\WF@everypar}\fi
  \fi
  \loop\ifnum\nv@qhead<\nv@qtail
    \global\advance\nv@qhead\@ne
    \typeout{NVQ flush \the\nv@qhead}%
    \edef\nv@t{\csname nv@q@type@\the\nv@qhead\endcsname}%
    \setcounter{\nv@t}{\csname nv@q@num@\the\nv@qhead\endcsname}%
    \nvorigfigure[!tb]\centering
      \begin{minipage}{\nvcolw}\setlength{\columnwidth}{\linewidth}\edef\@captype{\nv@t}\csname nv@q@body@\the\nv@qhead\endcsname\end{minipage}%
    \endnvorigfigure
  \repeat
  \setcounter{figure}{\value{nvfigureseen}}\setcounter{table}{\value{nvtableseen}}}
\RenewEnviron{figure}[1][]{\nv@dofloat{figure}}
\RenewEnviron{table}[1][]{\nv@dofloat{table}}
\nvfloatact{6}{hold}\nvfloatact{7}{hold}\nvfloatact{9}{pair}
\let\nv@origsection\section
\def\section{\global\nv@headingtrue\nv@origsection}
\let\nv@origsubsection\subsection
\def\subsection{\global\nv@headingtrue\nv@origsubsection}
\def\@afterheading{%
  \global\nv@headingfalse
  \ifnum\c@WF@wrappedlines>\@ne \global\advance\c@WF@wrappedlines-2 \fi
  \@nobreaktrue
  \everypar{%
    \if@nobreak
      \@nobreakfalse
      \clubpenalty \@M
      \if@afterindent \else
        {\setbox\z@\lastbox \ifvoid\z@\else\ifdim\wd\z@=\z@ \box\z@ \fi\fi}%
      \fi
    \else
      \clubpenalty \@clubpenalty
      \everypar{}%
    \fi}}
\let\nv@origfloathand\WF@floathand
\newif\ifnv@ok
\def\WF@floathand{%
  \ifnv@heading\else
    \nv@oktrue
    \if@nobreak \ifdim\dimexpr\pagegoal-\pagetotal\relax<\WF@size \nv@okfalse \fi\fi
    \ifnv@ok \nv@origfloathand \fi
  \fi}
\newdimen\nv@acc
\def\WF@mypar{\relax
 \WF@@par
 \ifnum\@@parshape=\z@ \let\WF@pspars\@empty \fi
 \global\advance\c@WF@wrappedlines-\prevgraf \prevgraf\z@
 \global\advance\nv@acc\parskip
 \ifdim\nv@acc>.75\baselineskip
   \global\advance\nv@acc-\baselineskip \global\advance\c@WF@wrappedlines\m@ne
 \fi
 \ifnum\c@WF@wrappedlines<\tw@ \global\nv@acc\z@ \WF@finale \fi}
\renewcommand{\topfraction}{0.9}
\renewcommand{\bottomfraction}{0.8}
\renewcommand{\textfraction}{0.08}
\renewcommand{\floatpagefraction}{0.85}
\makeatletter\g@addto@macro\@author{\par\vskip-3pt{\normalfont\fontsize{9}{13}\selectfont \href{https://research.nvidia.com/labs/amri/projects/repr-moe/}{https://research.nvidia.com/labs/amri/projects/repr-moe/}}}\makeatother
\let\nvmaketitle\maketitle
\renewcommand{\maketitle}{}
\RenewEnviron{strip}{\global\let\nvteaser\BODY}
\maketitle
\begin{strip}
  \centering
  \includegraphics[width=\linewidth]{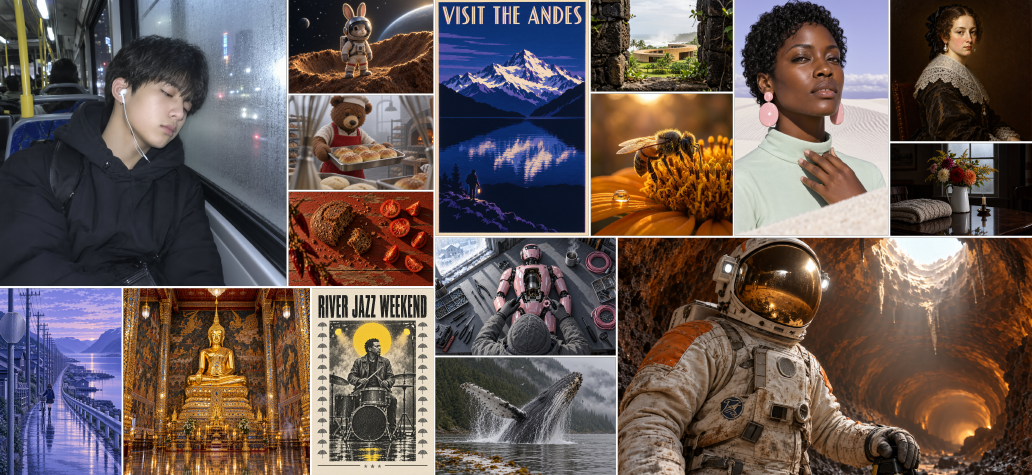}
  \captionsetup{hypcap=false}
  \captionof{figure}{\textbf{Visual results of Repr-MoE on text-to-image generation.} All images are generated in pixel space by a sparse MoE that activates 1.1B of its 11.2B parameters. Additional examples are provided in \supp{app:t2i-visual}.}
  \captionsetup{hypcap=true}
  \label{fig:headline}
\end{strip}

\begin{abstract}
Sparse mixture-of-experts (MoE) architectures let language models grow to trillions of parameters while activating only 3--10\% of them per token.
Visual generators have not followed: most state-of-the-art models remain dense, and when experts are added while each token activates the same number of them, the gains eventually plateau. We call this the \emph{Sparsity Ceiling}.
The experts remain utilized, so routing collapse does not explain this plateau; instead, we argue that it stems from the semantically weak representations learned from reconstruction or denoising alone, which leave the experts little to specialize on.
We introduce \emph{Repr-MoE}, which grounds the generator in pretrained visual knowledge, either by aligning the denoiser's intermediate features with a frozen vision encoder or by generating in a semantic latent space, while retaining standard top-$K$ routing.
With stronger representations, Repr-MoE breaks the Sparsity Ceiling: on ImageNet $256{\times}256$, increasing the expert count from 256 to 1024 reduces unguided FID by up to 23\%.
Paired with multi-head latent experts and the MuonH optimizer, Repr-MoEs match the guided FID of dense models with about half the per-image FLOPs, reach the final FIDs of PixelDiT and RAE in $4.1\times$ fewer epochs, and achieve state-of-the-art FIDs of 1.32 without guidance and 0.98 with guidance after 80 epochs.
In text-to-image generation, a pixel-space Repr-MoE that activates only 10\% of its 11.2B parameters outperforms strong pixel-space baselines on GenEval, DPG and Qwen-Image-Bench (QBench) at both 512 and 1024 resolution, and is comparable on GenEval and QBench to latent-space models that activate 2--18$\times$ as many parameters.

\end{abstract}

\nvmaketitle
\vskip-8pt
\noindent\begin{minipage}{\textwidth}
\nvteaser
\end{minipage}
\vskip10pt
\begingroup
\abscontent
\endgroup
\vskip8pt
\makeatother

\section{Introduction}
\label{sec:intro}

Diffusion transformers have become the backbone of modern visual generation, and their quality improves as model size increases~\cite{peebles2023dit,ma2024sit,yu2025pixeldit,liang2026ditscaling}. However, the training and inference costs of dense models grow with scale, as all parameters are activated at every denoising step. Language models faced the same trade-off and have largely resolved it with sparse mixture-of-experts (MoE) architectures~\cite{shazeer2017moe,fedus2022switch,jiang2024mixtral,dai2024deepseekmoe}, which route each token to a small subset of experts. As experts are added while each token activates the same number of them, these models become sparser and keep improving over a wide range of sparsity~\cite{kimi2025k2}, and frontier MoEs now have up to trillions of parameters while activating only 3--10\% of them per token~\cite{deepseek2024v2,deepseek2024v3,kimi2025k2,kimi2026k3}.

\begin{figure}[t]
  \centering
  \includegraphics[width=0.95\columnwidth]{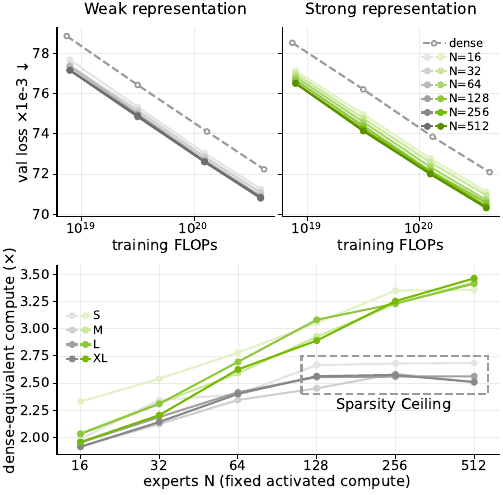}
  \caption{\textbf{A semantically strong representation breaks the Sparsity Ceiling.} MiniT2I at four scales, S to XL, 0.17B to 1.1B activated parameters, trained \hlc{hlgray}{without REPA (weak representation)} and \hlc{nvgreenlight}{with REPA (strong representation)}~\cite{yu2025repa}. At each scale, the number of activated experts is fixed, and $N$ grows from 16 to 512. Bottom: dense training compute to match each MoE's validation loss, relative to the MoE's (extrapolated beyond the largest dense model where needed).}
  \label{fig:saturation}
\end{figure}

One would expect visual generators to benefit in the same way, but so far they have not. Most state-of-the-art visual generators remain dense, and prior image and video MoEs report diminishing returns from additional experts~\cite{shi2025diffmoe,lingbot2026video} and smaller compute savings than in language models~\cite{ge2026chimera}. We observe the same pattern across class-conditional and text-to-image generation: with the number of activated experts held fixed, our MoEs first improve as experts are added but then plateau (Figs.~\ref{fig:saturation} and~\ref{fig:teacher-n}). We call this the \emph{Sparsity Ceiling}.

\begin{figure}[t]
  \centering
  \includegraphics[width=0.95\columnwidth]{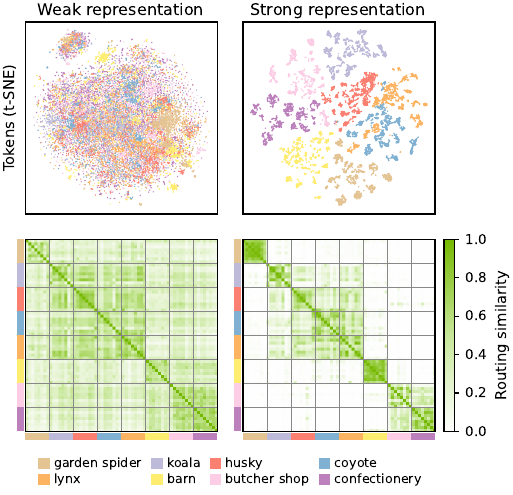}
  \caption{\textbf{A semantically strong representation lets experts specialize.} PixelDiT-B/16 MoEs with $N{=}256$ \hlc{hlgray}{without REPA (weak representation, left)} and \hlc{nvgreenlight}{with REPA (strong representation, right)}, on eight randomly drawn ImageNet classes with the first eight validation images of each, at noise level $t{=}0.5$. Top: t-SNE of the tokens entering the aligned MoE layer, colored by class. Bottom: routing similarity between images. With REPA, tokens cluster by class, and images of different classes share few experts.}
  \label{fig:cluster}
\end{figure}

The experts remain utilized (Sec.~\ref{sec:convergence}), so routing collapse does not explain the plateau. Instead, we find that experts specialize poorly: tokens from different classes stay intermixed at the MoE's input, and images of different classes share many experts (Fig.~\ref{fig:cluster}, left). Prior visual MoEs note this lack of specialization and add semantic router guidance~\cite{wei2026promoe,xu2026splitmoe}. However, a router can only partition the features it is given, and features from reconstruction or denoising alone are semantically weak compared with self-supervised vision encoders~\cite{yu2025repa,zheng2025rae}. We therefore argue that the Sparsity Ceiling stems from semantically weak representations.

\begin{figure*}[!t]
  \centering
  \resizebox{\textwidth}{!}{\providecolor{ovfill}{rgb}{1,1,1}\providecolor{ovdraw}{gray}{0.65}\providecolor{ovhi}{HTML}{76B900}\providecolor{ovhidraw}{HTML}{76B900}\providecolor{ovhitext}{rgb}{1,1,1}%
\begin{tikzpicture}[font=\footnotesize\sffamily,>={Stealth[length=4pt]},line width=0.5pt,
  box/.style={draw=ovdraw,rounded corners=1.5pt,fill=ovfill,minimum width=1.6cm,minimum height=0.48cm,inner sep=2pt,align=center},
  blk/.style={box},
  mat/.style={draw=ovdraw,rounded corners=1.5pt,fill=ovfill,minimum width=0.85cm,minimum height=1.3cm,inner sep=1pt,align=center},
  seg/.style={draw=ovdraw,rounded corners=1pt,fill=ovfill,minimum width=0.24cm,minimum height=0.7cm,inner sep=0pt},
  rt/.style={draw=ovdraw,rounded corners=1.5pt,fill=ovfill,minimum width=1.0cm,minimum height=0.48cm,inner sep=1pt},
  ex/.style={draw=ovdraw,rounded corners=0.6pt,fill=ovfill,minimum width=0.16cm,minimum height=0.38cm,inner sep=0pt},
  exon/.style={draw=ovhidraw,rounded corners=0.6pt,fill=ovhi,minimum width=0.16cm,minimum height=0.38cm,inner sep=0pt},
  dot/.style={draw=black!70,circle,inner sep=0.6pt,fill=white},
  note/.style={black!55},
  panel/.style={rounded corners=4pt,draw=black!25,fill=black!1.5}]

  \draw[panel] (0,-2.5) rectangle (6.25,2.3);
  \node (img) at (1.25,-1.9) {image};
  \node[box,dashed,minimum width=1.9cm] (ae) at (3.15,-1.9) {Autoencoder};
  \node[blk] (b1) at (5.25,-1.0) {blocks};
  \node[blk,draw=ovhidraw,fill=ovhi,text=ovhitext] (bl) at (5.25,0.2) {block $\ell$};
  \node[blk] (b3) at (5.25,1.4) {blocks};
  \node (vout) at (5.25,2.65) {$v_\theta$};
  \draw[->] (img) -- (ae);
  \draw[->] (ae.east) -| node[pos=0.8,right] {$x_t$} (b1.south);
  \draw[->] (b1) -- (bl);
  \draw[->] (bl) -- (b3);
  \draw[->] (b3) -- (vout);
  \node[box,minimum width=1.9cm] (enc) at (1.25,0.1) {Pretrained\\vision encoder};
  \draw[->] (img) -- (enc);
  \node[inner sep=2pt] (loss) at (1.25,1.55) {$\mathcal{L}_\mathrm{REPA}$};
  \draw[->] (enc.north) -- node[left] {$y$} (loss.south);
  \draw[->] (5.25,0.8) -- (3.9,0.8) |- node[pos=0.75,above] {$\phi(h_\ell)$} (loss.east);
  \node[anchor=south] at (1.25,1.8) {(a) alignment};
  \node[anchor=south] at (2.95,-1.62) {(b) semantic latents};
  \draw[black!40,dashed] (bl.north east) -- (7.0,2.3);
  \draw[black!40,dashed] (bl.south east) -- (7.0,-2.5);

  \begin{scope}[xshift=-0.5cm]
  \draw[panel] (7.5,-2.5) rectangle (10.3,2.3);
  \node (xin) at (9.05,-2.85) {$h_{\ell-1}$};
  \node[box] (n1) at (9.05,-1.47) {Norm};
  \node[box] (att) at (9.05,-0.79) {Attention};
  \coordinate (p1) at (9.05,-0.25);
  \node[box] (n2) at (9.05,0.59) {Norm};
  \node[box,draw=ovhidraw,fill=ovhi,text=ovhitext] (moe) at (9.05,1.27) {MoE};
  \coordinate (p2) at (9.05,1.81);
  \node (xout) at (9.05,2.65) {$h_\ell$};
  \draw[->] (xin) -- (n1);  \draw[->] (n1) -- (att);  \draw[->] (att) -- (n2);
  \draw[->] (n2) -- (moe);  \draw[->] (moe) -- (xout);
  \draw[->] (9.05,-2.01) -- (7.9,-2.01) |- (p1);
  \draw[->] (9.05,0.05) -- (7.9,0.05) |- (p2);
  \draw[black!40,dashed] (moe.north east) -- (11.05,2.3);
  \draw[black!40,dashed] (moe.south east) -- (11.05,-2.5);

  \draw[panel] (11.05,-2.5) rectangle (20.54,2.3);
  \node (x) at (10.72,0.05) {$h$};
  \node[mat] (down) at (12.45,0.05) {$W_\mathrm{in}$};
  \node[note,anchor=south] at (12.3,0.9) {latent projection};
  \draw[->] (x) -- (down);
  \foreach \y [count=\h] in {1.2, 0.5, -1.1} \node[seg] (a\h) at (13.85,\y) {};
  \draw[black!70,dashed] (13.73,-0.75) rectangle (13.97,0.15);
  \foreach \h in {1,2,3} \draw[->] (down.east) -- (a\h.west);
  \node[note,anchor=south] at (13.15,1.6) {head chunking};
  \draw[rounded corners=3pt,dashed,draw=black!35] (14.4,-1.6) rectangle (17.89,1.9);
  \node[note] at (16.15,1.67) {top-$K$ of $N$};
  \foreach \y/\ea/\eb [count=\h] in {1.2/1/5, 0.5/2/6, -1.1/0/4} {
    \node[rt] (r\h) at (15.1,\y) {router};
    \draw[->] (a\h.east) -- (r\h.west);
    \foreach \i in {0,...,7} {
      \pgfmathtruncatemacro{\on}{(\i==\ea) || (\i==\eb)}
      \ifnum\on=1 \node[exon] (e\h\i) at (16.0+\i*0.23,\y) {};
      \else       \node[ex]   (e\h\i) at (16.0+\i*0.23,\y) {};
      \fi
    }
    \draw[->] (r\h.east) -- (e\h0.west);
  }
  \node at (16.15,-0.25) {$\vdots$};
  \foreach \y [count=\h] in {1.2, 0.5, -1.1} {
    \node[seg] (c\h) at (18.44,\y) {};
    \draw[->] (e\h7.east) -- (c\h.west);
  }
  \draw[black!70,dashed] (18.32,-0.75) rectangle (18.56,0.15);
  \node[mat] (up) at (19.34,0.05) {$W_\mathrm{out}$};
  \foreach \h in {1,2,3} \draw[->] (c\h.east) -- (up.west);
  \node[box,draw=ovhidraw,fill=ovhi,text=ovhitext] (sh) at (16.15,-2.05) {shared};
  \draw[->] (11.55,0.05) |- (sh.west);
  \coordinate (plus) at (20.09,0.05);
  \draw[->] (up) -- (20.89,0.05);
  \draw[->] (sh.east) -| (plus);
  \end{scope}
\end{tikzpicture}}
  \caption{\textbf{Repr-MoE overview.} Left: we ground the model in a frozen pretrained encoder either by (a) aligning the hidden states of one block with its features or by (b) generating in a latent space shaped by it. Pixel-space models have no autoencoder (dashed) and use (a). Middle: a transformer block with the FFN replaced by MoE. Right: the multi-head latent MoE layer, which routes each of $H$ heads to its top-$K$ of $N$ experts in a latent space and adds a shared expert.}
  \label{fig:overview}
\end{figure*}
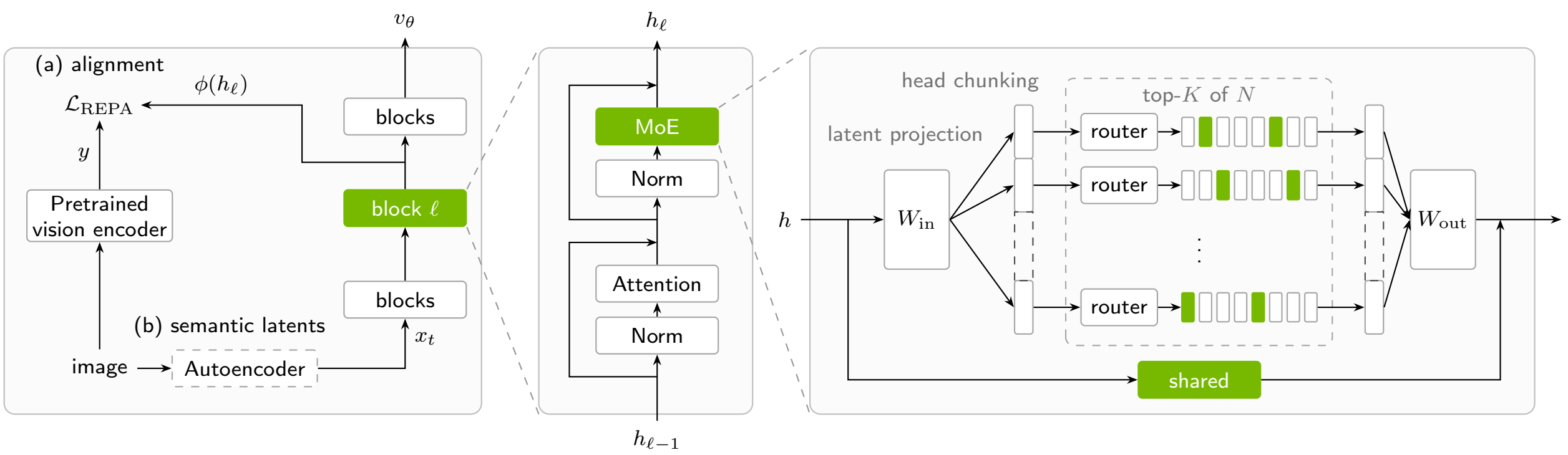

In this work, we introduce \emph{Repr-MoE}, which lets image generators benefit from more experts without activating more of them by strengthening their visual representations (Fig.~\ref{fig:overview}). We ground the generator in pretrained visual knowledge via two routes: aligning the denoiser's intermediate features with a frozen vision encoder, as in REPA~\cite{yu2025repa}, or generating directly in semantic latent spaces such as FLUX.2 VAE~\cite{bfl2025flux2} and RAE~\cite{zheng2025rae,singh2026raev2}. Both retain standard top-$K$ routing with load balancing, without prescribing semantic groups or expert assignments.

Semantically stronger representations give experts a clearer basis for specialization. With REPA, for example, tokens cluster by class, and images of different classes are routed to largely different experts (Fig.~\ref{fig:cluster}, right). Most importantly, this more differentiated expert usage accompanies continued improvements as experts are added, extending gains beyond the original plateau. On ImageNet, increasing the expert count from 256 to 1024 reduces unguided FID~\cite{heusel2017fid} by up to 23\% across seven alignment encoders, with consistent gains across four semantic latent spaces (Fig.~\ref{fig:teacher-n} and Table~\ref{tab:semvae}). These benefits extend to text-to-image pre-training: a 512-expert model matches the validation loss of a dense model trained with an estimated 3.4$\times$ its compute, compared with 2.6$\times$ without REPA (Fig.~\ref{fig:saturation}).

Turning this insight into effective large-scale models also requires a suitable training recipe. We therefore systematically study architectural and optimization choices for MoE diffusion transformers. Through controlled ablations in both pixel and latent space, we examine expert granularity, shared experts, multi-head latent designs~\cite{wu2024mhmoe,elango2026latentmoe,cui2026mhlatentmoe}, and Muon-family optimizers~\cite{jordan2024muon,liu2025moonlight,wen2026hyperball}. We identify which choices improve quality at matched activated compute and show that optimization substantially affects MoE's advantage over dense models (Table~\ref{tab:roadmap} and \supp{app:opt}). These findings establish a practical recipe combining representation grounding, multi-head latent experts, and hyperball Muon (MuonH).

The resulting models show strong performance on both ImageNet and text-to-image generation. On ImageNet, Repr-MoEs match the guided FID of dense models with about half the per-image FLOPs (Fig.~\ref{fig:c2i-pareto}). At scale, with about the same activated parameters as the dense models they build on, they reach the final FID of PixelDiT and RAE in $4.1\times$ fewer epochs (Fig.~\ref{fig:c2i-convergence}) and achieve state-of-the-art FIDs of 1.32 without guidance and 0.98 with guidance on RAEv2 latents after 80 epochs (Table~\ref{tab:c2i-headline}). In text-to-image generation, a pixel-space Repr-MoE that activates 10\% of its 11.2B parameters outperforms strong pixel-space baselines on GenEval, DPG and Qwen-Image-Bench~\cite{li2026qwenimagebench} (QBench) at both 512 and 1024 resolution, and is comparable on GenEval and QBench to latent-space models that activate 2--18$\times$ as many parameters (Table~\ref{tab:t2i-headline}).

Our contributions are fourfold: 1.~we confirm the Sparsity Ceiling across diverse image generation settings and link it to semantically weak representations; 2.~we show that grounding the generator in pretrained semantic representations breaks the Sparsity Ceiling; 3.~we systematically study scalable training recipes for MoE diffusion transformers via controlled ablations and identify multi-head latent experts and MuonH as effective; and 4.~we train Repr-MoEs reaching state-of-the-art FIDs of 1.32 without and 0.98 with guidance on ImageNet $256{\times}256$ and outperforming strong pixel-space text-to-image baselines.

\section{Related Work}

\paragraph{Mixture of experts in language models.} Sparse MoE layers~\cite{shazeer2017moe} route each token to a few experts, decoupling capacity from per-token compute in autoregressive~\cite{lepikhin2021gshard,fedus2022switch,jiang2024mixtral,dai2024deepseekmoe,deepseek2024v2,deepseek2024v3,muennighoff2025olmoe,kimi2025k2,kimi2026k3} and diffusion~\cite{zhu2025lladamoe,bie2025llada2,reda2026twotower} language models. Prior work studies load balancing, routing~\cite{zhou2022expertchoice,qi2025demons,wang2024auxfree,jin2026idbalancing,li2026mirrormoe}, and expert behavior~\cite{wang2026myth,wang2026dbes,chaudhari2026moelens}, while scaling studies find that more, finer experts improve efficiency up to a compute-dependent optimum~\cite{dai2024deepseekmoe,clark2022unified,krajewski2024finegrained,abnar2025sparsity,zhao2025comprehensive,nakamura2026reasoning,kimi2025k2}. These gains have not carried over to visual generation.

\paragraph{Mixture of experts in visual generation.} Following early use in recognition~\cite{riquelme2021vmoe}, visual generators have applied MoEs across noise levels~\cite{balaji2022ediffi,feng2023ernievilg2,park2024switchdit,wan22release}, inside U-Nets~\cite{xue2023raphael}, and in latent and pixel-space diffusion transformers~\cite{fei2024ditmoe,sun2025ecdit,liu2025recipe,jia2026mmoe,akiti2026nucleus,sandai2026magi2,cheng2025diffmoe,cai2025hidream,zheng2025dense2moe,dong2026pixelmoe}. Yet adding experts brings modest compute savings~\cite{ge2026chimera} and diminishing returns~\cite{shi2025diffmoe,lingbot2026video} while the models typically still activate 10--40\% of their parameters, against 3--10\% in frontier language MoEs~\cite{deepseek2024v2,deepseek2024v3,kimi2025k2,kimi2026k3}. Several prior works explore redesigning token routing~\cite{sun2025ecdit,akiti2026nucleus,shi2025diffmoe,yuan2025racedit,sharpmoe2026}. Closest to ours, ProMoE~\cite{wei2026promoe} and SplitMoE~\cite{xu2026splitmoe} guide the router semantically to encourage specialization, but derive this guidance from the denoiser's own features or from VAE features rather than a pretrained vision encoder, and study at most 100 experts. In contrast, we keep a standard top-$K$ router and pair a semantically stronger representation with a multi-head latent MoE layer and the MuonH optimizer, which sustains gains up to 1024 experts.

\paragraph{Representation learning for diffusion.} 
Stronger representations speed up convergence and improve sample quality~\cite{yu2025repa,zheng2025rae} in both latent~\cite{rombach2022ldm} and pixel-space~\cite{hoogeboom2024sid2,li2025jit,yu2025pixeldit,yu2026pixeldit2,chen2026dip,ma2026pixelgen,wang2026minit2i,zimagepixel2026} diffusion transformers~\cite{peebles2023dit,ma2024sit}. One line of work aligns the denoiser's features~\cite{yu2025repa,singh2025irepa,wang2026lap} with a frozen vision encoder such as DINOv2~\cite{oquab2024dinov2}, conditions on semantic features~\cite{razavi2026recon,ghosh2026dinoforcing,pan2026rg}, or derives the targets from the model itself~\cite{jiang2025sra,haghighi2026layersync,chefer2026selfflow,zhang2026jax}. Another denoises directly in a semantic latent space~\cite{yao2025vavae,leng2025repae,bfl2025flux2,zheng2025rae,singh2026raev2}. Almost all of this work targets dense models. Among MoEs, MEPA~\cite{zhou2026mepa} pairs alignment with a visual autoregressive MoE at one expert count, and Tong et~al.~\cite{beyond2026multimodal} find in multimodal pre-training that diffusion loss keeps improving with experts on RAE latents but saturates on VAE latents. We study both routes in pixel and latent diffusion and show each breaks the \emph{Sparsity Ceiling}.

\section{Method}

We present \emph{Repr-MoE}, a framework for scaling sparse diffusion transformers with a fixed number of activated experts (Fig.~\ref{fig:overview}). After reviewing flow matching and MoE routing in Sec.~\ref{sec:prelim}, we describe its two components. The first grounds the generator in a pretrained vision encoder so experts specialize, breaking the Sparsity Ceiling (Sec.~\ref{sec:grounding}). The second pairs a multi-head latent MoE layer with MuonH (Sec.~\ref{sec:recipe}).

\subsection{Preliminaries}
\label{sec:prelim}

\paragraph{Flow matching.}
Flow matching~\cite{lipman2023flow,liu2023flow,ma2024sit} interpolates between a clean sample $x$ (pixels or a latent) and Gaussian noise $\epsilon$ as $x_t=(1-t)\,x+t\,\epsilon$ for $t\in[0,1]$. The denoiser $v_\theta$ is trained to predict the velocity $v=\epsilon-x$ of this path by minimizing the loss $\mathcal{L}_{\text{FM}}=\mathbb{E}\,\|v_\theta(x_t,t,c)-v\|^2$, where $c$ is a class label or text prompt. Sampling integrates $v_\theta$ from $t{=}1$ to $t{=}0$, optionally with classifier-free guidance~\cite{ho2022cfg}.

\paragraph{Mixture of experts.}
An MoE layer~\cite{dai2024deepseekmoe} replaces the feed-forward network (FFN) of a transformer block with a set of experts $E_i$, each of which is itself an FFN. Every token uses $S$ shared experts and $K$ of $N$ routed experts:
\begin{equation}
  \operatorname{MoE}(h)=\sum_{i=1}^{S}E_i(h)+\sum_{i=S+1}^{S+N}g_i\,E_i(h),
\end{equation}
where $g_i$ is the router's score for expert $i$ if it is among the token's top $K$ and zero otherwise. The activated expert parameters thus stay fixed as $N$ grows; only the router, which scores all $N$ experts, grows with $N$ (see \supp{app:moe-design}). As in routing analyses of language MoEs~\cite{tang2025hiddencollab}, we define the \emph{routing similarity} of two images $x$ and $x'$ at a given MoE block and noise level as
\begin{equation}
  \operatorname{sim}(x,x')\!=\!\cos(\bar g(x),\bar g(x')),\;\bar g(x)\!=\!{\textstyle\sum_{m=1}^{M}}g(h^{m}),
\end{equation}
where $h^{m}$ is the $m$-th of the $M$ tokens of image $x$ and $g(h^{m})=(g_{S+1},\dots,g_{S+N})$ is its score vector, concatenated across heads when multi-head (Sec.~\ref{sec:recipe}).

\subsection{Representation grounding}
\label{sec:grounding}
To ground the generator in a pretrained vision encoder, we use one of two routes (Fig.~\ref{fig:overview}, left) and keep a standard top-$K$ router in both.

\paragraph{Representation alignment.}
Raw pixels and latents trained for reconstruction alone, such as those of a standard VAE, carry only a semantically weak representation, so the model must learn a stronger one for its experts to specialize. Similar to REPA~\cite{yu2025repa}, we align the hidden states $h_\ell$ of one block $\ell$, computed from the noisy input, with a frozen pretrained vision encoder's patch features $y$ of the clean image:
\begin{equation}
  \mathcal{L}_{\text{REPA}}=-\,\mathbb{E}\Big[\frac{1}{M}\sum_{m=1}^{M}\cos\big(y^{m},\phi(h_\ell^{m})\big)\Big],
\end{equation}
where $m$ indexes the $M$ patch tokens and $\phi$ is a small MLP projector. The training loss becomes $\mathcal{L}_{\text{FM}}+\lambda\,\mathcal{L}_{\text{REPA}}$, where $\lambda$ is a hyperparameter. Block $\ell$ is the aligned layer, always an MoE block. We use DINOv2~\cite{oquab2024dinov2} by default and compare other encoders in Sec.~\ref{sec:exp-align}.

\paragraph{Semantic latents.}
Alternatively, the latent space of a latent diffusion model can itself be shaped by a pretrained vision encoder. The generator is then grounded without an alignment loss. Two lines of work build such latents. The first keeps the VAE and trains it with a semantic objective, as in VA-VAE~\cite{yao2025vavae}, REPA-E~\cite{leng2025repae} and the FLUX.2 VAE~\cite{bfl2025flux2}. The second replaces the VAE with a representation autoencoder (RAE)~\cite{zheng2025rae,singh2026raev2}, which takes the features of a frozen pretrained encoder as latents and trains only a decoder that maps them back to pixels.

Both routes were proposed to speed up the training of dense models. We show that either one also breaks the Sparsity Ceiling of MoEs with a standard router (Sec.~\ref{sec:exp-align}).

\subsection{Architecture and optimization}
\label{sec:recipe}
We adopt our MoE layer and optimizer from recent language models: a multi-head latent MoE layer~\cite{cui2026mhlatentmoe} and the Muon family of optimizers~\cite{jordan2024muon,liu2025moonlight,kimi2025k2,wen2026hyperball}. We ablate each choice step by step from a dense baseline in Sec.~\ref{sec:exp-ablation}.

\paragraph{MoE designs.} In MoE blocks, we replace the dense FFN with a multi-head latent MoE layer (Fig.~\ref{fig:overview}, right). A projection $W_\mathrm{in}$ maps each token into a learned latent space~\cite{elango2026latentmoe,cui2026mhlatentmoe} and splits it into $H$ heads (\emph{head chunking})~\cite{wu2024mhmoe,sandai2026magi2}, each routed independently to $K$ of its own $N$ narrow experts~\cite{dai2024deepseekmoe,krajewski2024finegrained}, and $W_\mathrm{out}$ projects the concatenated heads back. A shared expert~\cite{dai2024deepseekmoe} acts on $h$ directly. Per-token parameters match about one dense FFN (see \supp{app:moe-design}). We balance load via a per-expert bias set by Quantile Balancing~\cite{kimi2026k3} without an auxiliary loss.

\paragraph{Optimizer.} Muon~\cite{jordan2024muon}, which orthogonalizes weight updates, is widely used for large language MoEs~\cite{liu2025moonlight,kimi2025k2,kimi2026k3,qwen2026next}, and Muon-family optimizers are now also used for diffusion transformers~\cite{chen2026cmuon,li2026muondit,akiti2026nucleus,dong2026pixelmoe,sandai2026magi2}. Muon helps MoE more than dense models (Sec.~\ref{sec:exp-ablation}). Without weight decay, however, it is unstable in our models: weights, activations and attention logits grow continuously, and pixel-space training without REPA diverges. Language models counter this with weight decay~\cite{liu2025moonlight}, which only partly stops it here. We instead use hyperball Muon (MuonH)~\cite{wen2026hyperball}, which fixes weight and update norms, similar in spirit to EDM2~\cite{karras2024edm2}. MuonH bounds all three quantities and outperforms Muon with weight decay in most settings (see \supp{app:opt}). We apply it to attention, FFN and expert matrices, and train most remaining parameters with AdamW, including the router~\cite{qwen2026next,xiaomi2026mimo}.

\section{Experiments}

\paragraph{Setup.} Unless stated otherwise, we compare each MoE with a matching dense model of roughly equal activated parameters, sharing width, depth, optimizer (Sec.~\ref{sec:recipe}) and learning rates. On ImageNet $256{\times}256$~\cite{deng2009imagenet}, we train PixelDiT-B/16~\cite{yu2025pixeldit} in pixel space and SiT-B/2~\cite{ma2024sit} on FLUX.1 VAE~\cite{bfl2024flux1} latents for 400K steps, with DINOv2~\cite{oquab2024dinov2} as the default REPA encoder, and report FID-50K~\cite{heusel2017fid}. Our main findings hold in both spaces. For text-to-image pre-training, we train pixel-space MiniT2I~\cite{wang2026minit2i} on CC12M~\cite{changpinyo2021cc12m} for 200K steps across scales S to XL (0.17B--1.1B activated parameters), reporting flow-matching loss on 1,024 held-out images as validation loss. Batch size scales with model size so every scale sees roughly the same tokens per parameter, making dense loss predictable across scales. Further details are in \supp{app:setup}.

\subsection{Representation breaks the Sparsity Ceiling}
\label{sec:exp-align}
We first test whether adding experts keeps improving visual generators, with and without representation grounding. We increase the number of experts $N$ in every MoE layer while keeping the number of experts that each token activates fixed, so that the total parameters grow with $N$ while the activated expert parameters do not, and report FID without guidance, which keeps the tuning of the guidance scale out of the comparison.

\begin{figure}[!tp]
  \centering
  \includegraphics[width=0.95\columnwidth]{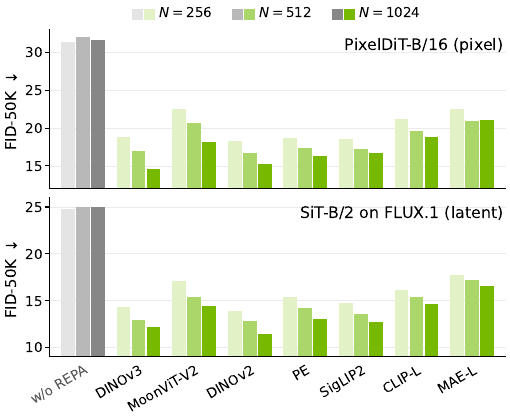}
  \caption{\textbf{REPA breaks the Sparsity Ceiling across encoders.} Unguided FID-50K at 400K steps on ImageNet $256{\times}256$ for $N{=}256$, 512 and 1024, \hlc{hlgray}{without REPA} and \hlc{nvgreenlight}{with seven encoders}.}
  \label{fig:teacher-n}
\end{figure}

\paragraph{Sparsity Ceiling.}\label{sec:convergence} Without representation grounding, increasing sparsity soon stops helping. On ImageNet, unguided FID changes by less than 1\% as the number of experts grows from 256 to 1024 (gray bars), in both pixel and latent spaces (Fig.~\ref{fig:teacher-n}). We observe a similar phenomenon in text-to-image pre-training (Fig.~\ref{fig:saturation}), where the validation loss stops improving beyond about 128 experts. This plateau sets in while our models still activate 13--31\% of their parameters, in line with the 10--40\% parameter activation at which earlier MoEs for image and video generation report diminishing gains~\cite{shi2025diffmoe,lingbot2026video}. Language models, in contrast, keep improving as experts are added~\cite{kimi2025k2}, and frontier models activate only 3--10\% of their parameters.

The Sparsity Ceiling is not due to routing collapse. The effective number of experts from expert loads (see \supp{app:effn}) is close to $N$ ($\ge 98.6\%$) in almost all MoE layers of PixelDiT, SiT and MiniT2I.

\begin{table}[!tp]
  \centering
  \small
  \setlength{\tabcolsep}{3pt}
  \caption{\textbf{No Sparsity Ceiling on semantic latents up to $N{=}1024$.} Unguided FID-50K at 400K steps; compare trends within a row. $\Delta$: relative change from $N{=}256$ to 1024. RAE and RAEv2 use DiT$^\text{DH}$-B~\cite{zheng2025rae}. \csq{black!9}~non-semantic, \csq{nvgreenlight}~semantic.}
  \label{tab:semvae}
  \resizebox{\columnwidth}{!}{%
  \setlength{\aboverulesep}{0pt}\setlength{\belowrulesep}{0pt}\renewcommand{\arraystretch}{1.05}
\begin{tabular}{|l|c|ccccc|c|}
    \hline
    Latent space & Dense & $N{=}64$ & 128 & 256 & 512 & 1024 & $\Delta$ \\
    \hline
    \rowcolor{black!9}FLUX.1 (reference) & 40.91 & 27.76 & 26.53 & 24.82 & 24.97 & 25.03 & +0.8\% \\
    \hline
    \rowcolor{nvgreenlight}FLUX.2             & 17.66 & 10.59 & 9.90 & 8.86 & 8.35 & 7.91 & $-$10.7\% \\
    \rowcolor{nvgreenlight}FLUX.1 (REPA-E)    & 18.99 & 11.75 & 10.78 & 10.60 & 9.81 & 9.35 & $-$11.8\% \\
    \hline
    \rowcolor{nvgreenlight}RAE              & 3.14 & 2.78 & 2.59 & 2.53 & 2.46 & 2.32 & $-$8.1\% \\
    \rowcolor{nvgreenlight}RAEv2            & 3.98 & 2.88 & 2.75 & 2.51 & 2.37 & 2.33 & $-$7.3\% \\ %
    \hline
  \end{tabular}}
\end{table}

\paragraph{Representation alignment.} Aligning the denoiser with a pretrained vision encoder lowers FID at every $N$ and removes this plateau, in pixel and latent space alike (Fig.~\ref{fig:teacher-n}). With our default encoder, DINOv2, FID falls with every doubling from 256 to 1024 experts, by 17\% in total, in both spaces. We repeat the sweep with six other encoders: the self-supervised DINOv3~\cite{simeoni2025dinov3}, the language-supervised CLIP-L~\cite{radford2021clip}, SigLIP~2~\cite{tschannen2025siglip2} and Perception Encoder (PE)~\cite{bolya2025pe}, MoonViT-V2~\cite{kimi2026k3} and MAE-L~\cite{he2022mae}. From 256 to 1024 experts, they lower FID by 7--23\%, and at 1024 experts the models activate only 4--7\% of their parameters, well below the 13--31\% at which the plateau sets in without alignment. MoonViT-V2 is trained from scratch with next-token prediction together with a language model~\cite{kimi2026k3}, and its gain of 16--20\% is among the largest. The smallest gain is with MAE-L, whose FID also stalls beyond 512 experts in pixel space. We conjecture that its features, learned by reconstructing pixels, are less semantic than those of the other encoders and give the experts less to specialize on.

\paragraph{Semantic latents.} Generating in a semantic latent space removes the plateau as well, without any alignment loss (Table~\ref{tab:semvae}). We replace the FLUX.1 latents with four semantic latent spaces, those of FLUX.2, REPA-E, RAE~\cite{zheng2025rae} and RAEv2~\cite{singh2026raev2}, and use the DiT$^\text{DH}$-B denoiser of RAE for the last two. In all four cases, FID keeps falling as we add experts, by 7--12\% from 256 to 1024. A semantic representation therefore breaks the Sparsity Ceiling whether it is learned by the denoiser or built into the latent space.%

\begin{figure}[!t]
  \centering
  \includegraphics[width=0.95\columnwidth]{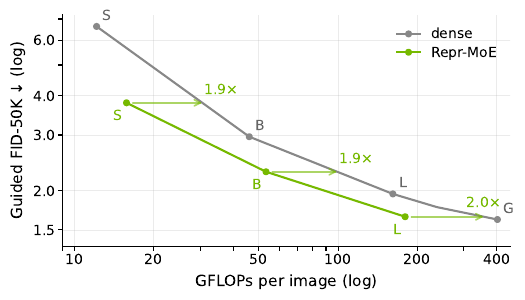}
  \caption{\textbf{Repr-MoE needs less compute than dense for the same FID.} SiT on FLUX.1 latents with REPA, ImageNet $256{\times}256$. Arrows mark the dense GFLOPs at equal FID.}
  \label{fig:c2i-pareto}
\end{figure}

\paragraph{Text-to-image pre-training.} REPA also breaks it in text-to-image pre-training (Fig.~\ref{fig:saturation}). With REPA, validation loss keeps improving up to 512 experts at scales M to XL, where the models activate only 8--10\% of their parameters. Averaged across S--XL, a 512-expert MoE matches the validation loss of a dense model trained with an estimated $3.4\times$ its compute when both use REPA, up from $2.6\times$ when neither does.

\paragraph{Expert specialization.} We find that REPA makes experts specialize (Fig.~\ref{fig:cluster}). Without REPA, different classes share many experts: in PixelDiT's aligned block, routing similarity (Sec.~\ref{sec:prelim}) between two images is 0.53 within a class and 0.31 across classes. With REPA, it is 0.58 and 0.08, so different classes share far fewer experts. Experts also specialize by class on RAE and RAEv2 latents without an alignment loss (see \supp{app:mech}).

\subsection{Compute efficiency}
\label{sec:exp-scale}
Sec.~\ref{sec:exp-align} compares models at matched activated parameters. We now ask how large a dense model Repr-MoE can match. Repr-MoE, here SiT with REPA on FLUX.1 latents, matches a dense model's guided FID with about half the per-image FLOPs (Fig.~\ref{fig:c2i-pareto}). Following DiT~\cite{peebles2023dit} and SiT~\cite{ma2024sit}, we measure single-pass FLOPs. We train dense SiTs and Repr-MoEs of increasing size on FLUX.1 latents with REPA for 400K steps, evaluate each at its best guidance scale, and interpolate between dense models to find equal-FID FLOPs. Repr-MoE matches a dense model with about 1.9$\times$ its FLOPs at SiT-S and SiT-B, and 2.0$\times$ at SiT-L (see \supp{app:pareto}).

\begin{figure}[!t]
  \centering
  \includegraphics[width=0.95\columnwidth]{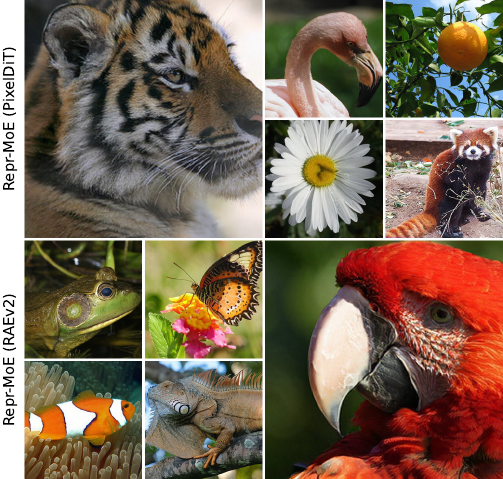}
  \caption{\textbf{Visual results of Repr-MoE on class-conditional generation.} ImageNet $256{\times}256$ samples after 320 epochs (PixelDiT) and 80 epochs (RAEv2). Additional examples are provided in \supp{app:c2i-visual}.}
  \label{fig:c2i-visual}
\end{figure}

\begin{figure}[!t]
  \centering
  \includegraphics[width=0.95\columnwidth]{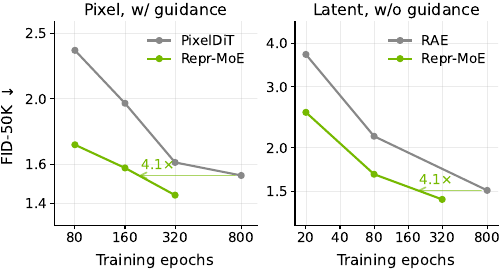}
  \caption{\textbf{Repr-MoE needs fewer training epochs.} FID-50K on ImageNet $256{\times}256$. Arrows mark where Repr-MoE reaches the final FID of PixelDiT and RAE, found by log-log interpolation between our evaluated checkpoints.}
  \label{fig:c2i-convergence}
\end{figure}

\begin{table*}[!tp]
  \centering
  \footnotesize
  \setlength{\tabcolsep}{5.4pt}
  \caption{\textbf{System-level comparison on ImageNet $256{\times}256$.} Repr-MoE on RAEv2 latents has the lowest FID both without and with guidance after only 80 epochs, and in pixel space it reaches a guided FID of 1.44 in about half the epochs of the next best listed model. Baselines are from each paper, except their FDr$^6$~\cite{yang2026fdloss,singh2026raev2} and the parameters of RAEv2 (from its released checkpoint; 839M in its paper). IS~\cite{salimans2016improved}; Pre.\ and Rec.: precision and recall~\cite{kynkaanniemi2019improved}. Act.\ and Total are activated and total parameters. SiD2~\cite{hoogeboom2024sid2} reports a guided FID of 1.38 in pixel space but not its model size, and is not listed. The best and second best in each block are in bold and underlined.}
  \label{tab:c2i-headline}
  \resizebox{\textwidth}{!}{%
  \setlength{\aboverulesep}{0pt}\setlength{\belowrulesep}{0pt}\renewcommand{\arraystretch}{1.05}
\begin{tabular}{|l|ccc|ccccc|ccccc|}
    \hline
     & & & & \multicolumn{5}{c|}{w/o guidance} & \multicolumn{5}{c|}{w/ guidance} \\
    \cmidrule{5-9}\cmidrule{10-14}
    Method & Act. & Total & Epochs & FID$\downarrow$ & IS$\uparrow$ & Pre.$\uparrow$ & Rec.$\uparrow$ & FDr$^6\!\downarrow$ & FID$\downarrow$ & IS$\uparrow$ & Pre.$\uparrow$ & Rec.$\uparrow$ & FDr$^6\!\downarrow$ \\
    \hline
    \grouprow \multicolumn{14}{|l|}{\textbf{\textit{Pixel space}}} \\
    ADM~\cite{dhariwal2021adm} & 554M & 554M & 396 & 10.94 & 100.98 & \bestv{0.69} & \secondv{0.63} & -- & 4.59 & 186.70 & \bestv{0.82} & 0.52 & -- \\
    PixelFlow-XL~\cite{chen2025pixelflow} & -- & -- & -- & -- & -- & -- & -- & -- & 1.98 & 282.1 & \secondv{0.81} & 0.60 & -- \\
    PixNerd-XL~\cite{wang2026pixnerd} & 700M & 700M & 160 & -- & -- & -- & -- & -- & 2.15 & 297 & 0.79 & 0.59 & -- \\
    JiT-H/16~\cite{li2025jit} & 953M & 953M & 600 & -- & -- & -- & -- & -- & 1.86 & \secondv{303.4} & 0.78 & 0.62 & -- \\
    PixelDiT-XL~\cite{yu2025pixeldit} & 797M & 797M & 320 & -- & -- & -- & -- & -- & 1.61 & 292.7 & 0.78 & \secondv{0.64} & -- \\
    PixelDiT-XL~\cite{yu2025pixeldit} & 797M & 797M & 800 & -- & -- & -- & -- & -- & 1.54 & 297.0 & 0.78 & \bestv{0.65} & -- \\
    PixelDiT2-H/16~\cite{yu2026pixeldit2} & 1008M & 1008M & 600 & -- & -- & -- & -- & -- & \secondv{1.46} & 301.6 & -- & -- & -- \\
    DeCo-XL~\cite{ma2026deco} & 682M & 682M & 800 & -- & -- & -- & -- & -- & 1.62 & 301 & 0.80 & 0.62 & -- \\
    \oursrow Repr-MoE (PixelDiT) & 854M & 6.80B & 160 & \secondv{7.85} & \secondv{135.8} & 0.67 & \bestv{0.68} & \secondv{10.38} & 1.58 & 302.0 & 0.78 & \bestv{0.65} & \secondv{3.60} \\
    \oursrow Repr-MoE (PixelDiT) & 854M & 6.80B & 320 & \bestv{6.87} & \bestv{149.6} & \secondv{0.68} & \bestv{0.68} & \bestv{9.34} & \bestv{1.44} & \bestv{308.5} & 0.79 & \bestv{0.65} & \bestv{3.18} \\
    \hline
    \grouprow \multicolumn{14}{|l|}{\textbf{\textit{Latent space}}} \\
    DiT-XL/2~\cite{peebles2023dit} & 675M & 675M & 1400 & 9.62 & 121.50 & 0.67 & \secondv{0.67} & -- & 2.27 & 278.24 & \bestv{0.83} & 0.57 & -- \\
    SiT-XL/2~\cite{ma2024sit} & 675M & 675M & 1400 & 8.26 & 131.65 & 0.68 & \secondv{0.67} & -- & 2.06 & 277.50 & \bestv{0.83} & 0.59 & 8.44 \\
    DDT-XL~\cite{wang2026ddt} & 675M & 675M & 400 & 6.27 & 154.7 & 0.68 & \bestv{0.69} & -- & 1.26 & 310.6 & 0.79 & 0.65 & 5.70 \\
    REPA (SiT-XL/2)~\cite{yu2025repa} & 675M & 675M & 800 & 5.9 & 157.8 & 0.70 & \bestv{0.69} & -- & 1.42 & 305.7 & 0.80 & 0.65 & 5.45 \\
    LightningDiT-XL~\cite{yao2025vavae} & 675M & 675M & 800 & 2.17 & 205.6 & 0.77 & 0.65 & -- & 1.35 & 295.3 & 0.79 & 0.65 & 4.57 \\
    REG (SiT-XL/2)~\cite{wu2025reg} & 677M & 677M & 800 & 1.8 & 230.8 & 0.77 & 0.66 & -- & 1.36 & 299.4 & 0.77 & \secondv{0.66} & 4.64 \\
    REPA-E (SiT-XL/2)~\cite{leng2025repae} & 675M & 675M & 800 & 1.69 & 219.3 & 0.77 & \secondv{0.67} & -- & 1.12 & 302.9 & 0.79 & \secondv{0.66} & 3.04 \\
    RAE (DiT$^\text{DH}$-XL)~\cite{zheng2025rae} & 839M & 839M & 800 & 1.51 & 242.9 & \secondv{0.79} & 0.63 & -- & 1.13 & 262.6 & 0.78 & \bestv{0.67} & 3.26 \\
    RAEv2~\cite{singh2026raev2} & 875M & 875M & 80 & 1.65 & 228.0 & -- & -- & -- & 1.06 & 255.3 & -- & -- & \secondv{2.17} \\
    DiT-MoE-XL~\cite{fei2024ditmoe} & 1.5B & 4.1B & 5600 & -- & -- & -- & -- & -- & 1.72 & \secondv{315.73} & \bestv{0.83} & 0.64 & -- \\
    DiffMoE-L~\cite{shi2025diffmoe} & 458M & 1.18B & 1400 & 9.60 & 131.46 & 0.67 & \secondv{0.67} & -- & 2.13 & 274.39 & \secondv{0.81} & 0.60 & -- \\
    Race-DiT~\cite{yuan2025racedit} & 707M & 2.79B & 1400 & -- & -- & -- & -- & -- & 2.06 & \bestv{318.64} & \bestv{0.83} & 0.60 & -- \\
    ProMoE~\cite{wei2026promoe} & 675M & 1.57B & 400 & 8.16 & 139.66 & -- & -- & -- & 2.08 & 304.26 & -- & -- & -- \\
    \oursrow Repr-MoE (RAE) & 846M & 7.54B & 320 & \secondv{1.42} & \secondv{253.8} & \secondv{0.79} & 0.64 & \secondv{4.33} & \secondv{1.01} & 274.8 & 0.78 & \secondv{0.66} & 3.15 \\
    \oursrow Repr-MoE (RAEv2) & 816M & 9.18B & 80 & \bestv{1.32} & \bestv{256.9} & \bestv{0.81} & 0.63 & \bestv{2.86} & \bestv{0.98} & 269.6 & 0.79 & \secondv{0.66} & \bestv{1.93} \\
    \hline
  \end{tabular}}
\end{table*}

\subsection{System-level comparison}
\label{sec:exp-headline}

The experiments above are controlled comparisons at modest scale. We now scale Repr-MoE up on ImageNet and on text-to-image generation. These models use $N{=}256$ or 512 experts and activate only 9--13\% of their 6.8B--11.2B parameters, with about the same activated parameters as the dense models that they build on. Baseline numbers are taken from prior papers where reported, while QBench scores and unreported baseline metrics are measured by us (see \supp{app:setup-t2i}); the training setups differ.

\paragraph{Class-conditional ImageNet $256{\times}256$.} Table~\ref{tab:c2i-headline} compares Repr-MoE with published models. PixelDiT reports FID with guidance only, so we compare with guidance in pixel space. There, Repr-MoE reaches a guided FID of 1.58 after 160 epochs, below the 1.61 that PixelDiT-XL reaches after 320, and 1.44 after 320 epochs. On RAEv2 latents, it reaches an unguided FID of 1.32 after 80 epochs, against 1.65 for the dense RAEv2 model at the same training length. This is the lowest in Table~\ref{tab:c2i-headline}, and below the 1.51 that RAE reaches after 800 epochs. With guidance, it reaches 0.98, against 1.06 for the dense RAEv2 model. On RAE latents, it reaches 1.42 after 320 epochs, and 1.01 with guidance. Compared with the 800-epoch results of PixelDiT and RAE, Repr-MoE needs $4.1\times$ fewer epochs to reach the same FID in both cases (Fig.~\ref{fig:c2i-convergence}). We show visual results in Fig.~\ref{fig:c2i-visual}.

\begin{table}[!tp]
  \centering
  \small
  \setlength{\tabcolsep}{5.5pt}
  \setlength{\aboverulesep}{0pt}
  \setlength{\belowrulesep}{0pt}
  \renewcommand{\arraystretch}{1.05}
  \caption{\textbf{From a dense model to Repr-MoE.} The full recipe more than halves the unguided FID of the dense model in both spaces. FID-50K$\downarrow$ on ImageNet 256 at 400K steps with REPA, without and with guidance. Each row adds one change at matched activated parameters. The vanilla MoE has $N{=}8$, later rows $N{=}64$.}
  \label{tab:roadmap}
  \resizebox{\columnwidth}{!}{%
  \begin{tabular}{|l|ccc|ccc|}
    \hline
    \multicolumn{1}{|l|}{} & \multicolumn{3}{c|}{PixelDiT (pixel)} & \multicolumn{3}{c|}{SiT on FLUX.1 (latent)} \\
    \cmidrule{2-4}\cmidrule{5-7}
    Variant & Total & unguided & guided & Total & unguided & guided \\
    \hline
    Dense, AdamW & 185M & 46.38 & 4.33 & 131M & 36.62 & 4.19 \\
    \hline \grouprow \multicolumn{7}{|l|}{\textbf{\textit{Optimizer}}} \\
    + Muon & 185M & 35.85 & 3.51 & 131M & 27.89 & 3.36 \\
    + MuonH & 185M & 30.65 & 3.32 & 131M & 23.52 & 2.92 \\
    \hline \grouprow \multicolumn{7}{|l|}{\textbf{\textit{Architecture}}} \\
    + vanilla MoE & 515M & 27.14 & 2.94 & 428M & 22.24 & 2.85 \\
    + fine-grained & 516M & 24.12 & 2.79 & 428M & 18.99 & 2.58 \\
    + shared & 521M & 24.86 & 2.83 & 434M & 18.30 & 2.51 \\
    + head chunking & 525M & 22.51 & 2.72 & 437M & 17.41 & 2.50 \\
    + latent projection & 441M & 22.14 & 2.80 & 361M & 15.39 & 2.39 \\
    \hline
  \end{tabular}}
\end{table}

\begin{table}[!tp]
  \centering
  \small
  \setlength{\tabcolsep}{1.6pt}
  \caption{\textbf{System-level comparison on text-to-image generation.} Repr-MoE scores highest among pixel-space models on all three metrics at both resolutions, and at $1024{\times}1024$ is second on QBench only to Qwen-Image-2.1. Act.\ and Total: activated and total parameters, in billions. $^\dagger$With the official prompt rewriter. The best and second best in each block are in bold and underlined. Evaluation details are in \supp{app:setup-t2i}.}
  \label{tab:t2i-headline}
  \resizebox{\columnwidth}{!}{%
  \setlength{\aboverulesep}{0pt}\setlength{\belowrulesep}{0pt}\renewcommand{\arraystretch}{1.24}
\begin{tabular}{|l|c|cc|ccc|}
    \hline
    Method & Space & Act. & Total & GenEval$\uparrow$ & DPG$\uparrow$ & QBench$\uparrow$ \\
    \hline
    \grouprow \multicolumn{7}{|l|}{\textbf{\textit{512$\boldsymbol{\times}$512}}} \\
    PixArt-$\alpha$~\cite{chen2024pixartalpha} & Latent & 0.6 & 0.6 & 0.48 & 71.6 & 38.1 \\
    PixArt-$\Sigma$~\cite{chen2024pixartsigma} & Latent & 0.6 & 0.6 & 0.52 & 79.5 & 40.2 \\
    PixelFlow~\cite{chen2025pixelflow} & Pixel & 0.9 & 0.9 & 0.60 & 77.9 & 39.9 \\
    PixNerd~\cite{wang2026pixnerd} & Pixel & 1.2 & 1.2 & 0.73 & 80.9 & 33.9 \\
    DeCo~\cite{ma2026deco} & Pixel & 1.1 & 1.1 & 0.86 & 81.4 & 36.8 \\
    PixelDiT-T2I~\cite{yu2025pixeldit} & Pixel & 1.3 & 1.3 & 0.78 & 83.7 & 37.5 \\
    MiniT2I-B/16~\cite{wang2026minit2i} & Pixel & 0.25 & 0.25 & 0.87 & 84.2 & 40.1 \\
    MiniT2I-L/16~\cite{wang2026minit2i} & Pixel & 0.91 & 0.91 & \secondv{0.88} & \secondv{85.9} & \secondv{41.1} \\
    \oursrow Repr-MoE (ours) & Pixel & 1.1 & 11.2 & \bestv{0.91} & \bestv{88.15} & \bestv{49.1} \\
    \hline
    \grouprow \multicolumn{7}{|l|}{\textbf{\textit{1024$\boldsymbol{\times}$1024}}} \\
    GPT Image 1 [High]~\cite{openai2025gptimage1} & -- & -- & -- & 0.84 & 85.15 & -- \\
    Seedream 3.0~\cite{gao2025seedream3} & -- & -- & -- & 0.84 & \secondv{88.27} & -- \\
    FLUX.1 [Dev]~\cite{bfl2024flux1} & Latent & 12 & 12 & 0.66 & 83.84 & 43.8 \\
    SD3 Medium~\cite{esser2024sd3} & Latent & 2 & 2 & 0.62 & 84.08 & 41.3 \\
    BAGEL~\cite{deng2025bagel} & Latent & 7 & 14 & 0.88$^{\dagger}$ & 85.44 & 42.8 \\
    HiDream-I1-Full~\cite{cai2025hidream} & Latent & -- & 17 & 0.83 & 85.89 & 46.8 \\
    Lumina-Image 2.0~\cite{qin2025lumina2} & Latent & 3 & 3 & 0.73 & 87.20 & 41.2 \\
    Z-Image~\cite{zimage2025} & Latent & 6 & 6 & 0.84 & 88.14 & 48.4 \\
    Qwen-Image~\cite{wu2025qwenimage} & Latent & 20 & 20 & 0.87 & \bestv{88.32} & 49.1 \\
    Qwen-Image-2.1$^{\dagger}$~\cite{qwen2026qwenimage21} & Latent & 7 & 7 & 0.81 & 84.81 & \bestv{59.2} \\
    BLIP3o-NEXT-G-G~\cite{chen2025blip3onext} & Latent & 3 & 3 & \secondv{0.91} & 81.93 & 41.3 \\
    i1$^{\dagger}$~\cite{zeng2026i1} & Latent & 3 & 3 & 0.84 & 86.73 & 47.1 \\
    L2P~\cite{l2p2026} & Pixel & 6 & 6 & 0.76 & 86.00 & 46.6 \\
    AsymFLUX.2 klein~\cite{asymflow2026} & Pixel & 9 & 9 & 0.82 & 86.8 & 46.2 \\
    Iris-3B~\cite{iris2026} & Pixel & 3 & 3 & 0.80 & 86.52 & 46.7 \\
    Z-Image-Pixel~\cite{zimagepixel2026} & Pixel & 6 & 6 & 0.76 & 87.60 & -- \\
    \oursrow Repr-MoE (ours) & Pixel & 1.1 & 11.2 & \bestv{0.93} & 87.87 & \secondv{56.1} \\
    \hline
  \end{tabular}}
\end{table}

\paragraph{Text-to-image.} We build our text-to-image model on MiniT2I~\cite{wang2026minit2i} as an XL MoE with 512 experts, which activates 1.1B of its 11.2B parameters. We pre-train it on an internal dataset at progressively higher resolutions and, following MiniT2I, fine-tune it on internal high-quality data (see \supp{app:setup-t2i}). At $512{\times}512$, it scores highest in Table~\ref{tab:t2i-headline} on GenEval~\cite{ghosh2023geneval}, DPG~\cite{hu2024ella} and QBench. At $1024{\times}1024$, its GenEval of 0.93 is the highest in the table, and only Qwen-Image-2.1 exceeds its QBench of 56.1. Its DPG of 87.87 is above that of every pixel-space model in the table and competitive with latent models that activate 2--18$\times$ more parameters. At the same resolution, it scores 76.0 on PRISM-Bench~\cite{fang2026prism}, which is also competitive with most open-source baselines. We show visual results in Fig.~\ref{fig:headline}, with more samples in Appendix~\ref{app:t2i-visual} and detailed benchmarks in Appendix~\ref{app:prism}.

\subsection{Ablations}
\label{sec:exp-ablation}
Starting from a dense model trained with REPA, we build Repr-MoE up one step at a time in both pixel and latent space, at matched activated parameters (Table~\ref{tab:roadmap}). Together, the steps lower unguided FID from 46.38 to 22.14 in pixel space and from 36.62 to 15.39 in latent space, and guided FID from 4.33 to 2.80 and from 4.19 to 2.39, respectively.

\paragraph{Optimizer.} Replacing AdamW with Muon lowers unguided FID by 23--24\%, and MuonH lowers it by another 15--16\%. Muon-family optimizers help MoE more than dense: MoE's unguided FID is 12--20\% below the dense model's with AdamW versus 24--40\% with Muon-family optimizers (see \supp{app:opt}). Prior reports on Muon for diffusion transformers are mixed~\cite{li2026muondit,chen2026cmuon,ge2026chimera}. Muon's gains come with instability (Sec.~\ref{sec:recipe}); MuonH stays stable while beating AdamW in all four settings and Muon with weight decay in three, staying close in the fourth (latent without REPA; see \supp{app:opt}).

\paragraph{MoE designs.} With MuonH, the MoE steps lower unguided FID by a further 28\% in pixel space and 35\% on latents. Splitting the experts into finer ones lowers it by 11--15\% and head chunking by 5--9\%, while the shared expert helps only on latents. The last step, the latent projection, cuts total parameters by 16--17\% while still lowering unguided FID in both spaces. With guidance, FID improves on latents and rises slightly in pixel space, from 2.72 to 2.80.

\section{Conclusion}
We confirm the \emph{Sparsity Ceiling} in class-conditional and text-to-image diffusion MoEs: when each token activates a fixed number of experts, adding experts soon stops helping. We argue that it stems from semantically weak representations, which give the experts little to specialize on.
Grounding the model in a pretrained vision encoder, either by aligning the denoiser's features or by generating in a semantic latent space, breaks the Sparsity Ceiling with a standard top-$K$ router and no routing loss.
We pair this grounding with multi-head latent experts and MuonH, and find that Muon-family optimizers widen MoE's advantage over dense models. The resulting \emph{Repr-MoEs} match dense ImageNet models with about half the per-image FLOPs, achieve state-of-the-art FIDs of 1.32 without guidance and 0.98 with guidance, and outperform strong pixel-space text-to-image baselines at both 512 and 1024 resolution while activating 10\% of their 11.2B parameters.
Our findings point to the importance of the representation itself. However, a separate frozen encoder is not the only way to obtain representations, and Self-Flow~\cite{chefer2026selfflow} argues that such an encoder can become a bottleneck as generators scale.
Our largest MoE models activate about 1.1B parameters, whereas recent work finds that larger image and video generators are themselves general vision learners~\cite{gabeur2026image,wang2026video}. Whether the Sparsity Ceiling and the benefit of grounding persist at larger scales remains to be verified.
This raises two natural questions. The first is how to build strong representations for video and unified multimodal MoEs~\cite{lingbot2026video,beyond2026multimodal}. The second is whether they can emerge without a separate encoder, through self-alignment~\cite{jiang2025sra,haghighi2026layersync,chefer2026selfflow} or directly from language supervision in unified models. Language supervision looks promising, as MoonViT-V2, trained end to end with a language model~\cite{kimi2026k3}, is already among our strongest encoders. Overall, our results suggest that the sparse scaling seen in language MoEs can carry over to image generation once the representation is semantically strong.

\nvflushfloats
\nvnarrowfalse
\section*{Acknowledgements}
We thank Zhengyang Liang, Hang Gao, Jun Gao, and Wei Yu for helpful discussions and feedback.

\section*{LLM Usage Statement}
We used LLM-based assistants (Claude, Anthropic; GPT via Codex, OpenAI; Gemini, Google) in this project. Under the authors' direction, they wrote and edited code for training, evaluation and figures, helped plan and run experiments and summarize logs and results, helped search and check the related literature and citations, and assisted with drafting and editing the text. Every experimental design was approved by the authors. The results of our models come from our own runs and were checked against the raw result files; baseline numbers are taken from the cited papers or evaluated by us where noted. The authors reviewed and take full responsibility for all text, code and results.

{
    \small
    \bibliographystyle{ieeenat_fullname}
    \bibliography{references}
}

\clearpage
\let\figure\nvorigfigure \let\endfigure\endnvorigfigure
\let\table\nvorigtable \let\endtable\endnvorigtable
\makeatletter
\let\nv@origcaption\caption
\let\nv@origlabel\label
\renewcommand{\enlargethispage}[1]{}
\let\nvorignewpage\newpage
\let\nvorigclearpage\clearpage
\def\clearpage{\let\newpage\nvorignewpage \nvorigclearpage \let\newpage\relax}
\let\newpage\relax
\@clubpenalty=10000
\clubpenalty=10000
\widowpenalty=10000
\let\nv@apporigsec\section
\renewcommand{\section}{\suppressfloats[t]\nv@apporigsec}
\makeatother
\appendix
\section*{Appendix}
\noindent This appendix first covers experimental configurations (Sec.~\ref{app:setup}) and MoE design and optimization (Sec.~\ref{app:arch-opt}).
It then presents additional quantitative results on expert scaling, compute efficiency, and text-to-image benchmarks (Sec.~\ref{app:scaling}), followed by routing and specialization analyses (Sec.~\ref{app:routing}).
We conclude with extended visual results (Sec.~\ref{app:visuals}).

\section{Experimental setup}
\label{app:setup}
This section details the model architectures, training recipes, and evaluation protocols for our text-to-image and class-conditional ImageNet experiments.
Across all controlled comparisons, each MoE matches the width, depth, and approximate activated parameters of its dense counterpart (Sec.~\ref{app:moe-design}).
When varying the expert count $N$, we hold $K$, $S$, and the expert width fixed, so that the activated expert parameters stay constant while the total parameters grow.

\subsection{Text-to-image}
\label{app:setup-t2i}
Our text-to-image experiments comprise a CC12M~\cite{changpinyo2021cc12m} scaling study (Fig.~\ref{fig:saturation}) and a scaled-up model (Table~\ref{tab:t2i-headline}), summarized in Table~\ref{tab:setting-t2i}.
Following MiniT2I~\cite{wang2026minit2i}, all models use a frozen Flan-T5-Large text encoder~\cite{chung2022flant5}.
With REPA~\cite{yu2025repa}, we align features to DINOv2-B~\cite{oquab2024dinov2} with $\lambda{=}0.1$ via a three-layer MLP projector.

\begin{table}[!b]
  \centering
  \footnotesize
  \caption{\textbf{Text-to-image configurations in pixel space.} Parameter counts exclude the text encoder, with total ranges over $N$.}
  \label{tab:setting-t2i}
  \arrayrulecolor{black}\renewcommand{\arraystretch}{0.97}
\setlength{\tabcolsep}{5pt}
\resizebox{\textwidth}{!}{%
\begin{tabular}{|l|*{4}{w{c}{1.5cm}}|w{c}{4.7cm}|}
    \hline
    \multirow{2}{*}{Setting} & \multicolumn{4}{c|}{Scaling study (Fig.~\ref{fig:saturation})} & System-level (Table~\ref{tab:t2i-headline}) \\
    \cline{2-6}
     & Scale S & Scale M & Scale L & Scale XL & Repr-MoE (Scale XL) \\
    \hline \grouprow \multicolumn{6}{|l|}{\textbf{\textit{Architecture}}} \\
    Patch size & \multicolumn{4}{c|}{32} & 16 \\
    Width & 768 & 1024 & 1280 & 1536 & 1536 \\
    Depth & 10 & 12 & 16 & 20 & 20 \\
    Dense parameters & 0.15B & 0.32B & 0.66B & 1.18B & -- \\
    \hline \grouprow \multicolumn{6}{|l|}{\textbf{\textit{MoE layer}}} \\
    Heads $H$ & 6 & 8 & 10 & 12 & 12 \\
    Experts per head $N$ & \multicolumn{4}{c|}{16, 32, 64, 128, 256, 512} & 512 \\
    Routed experts $K$ & \multicolumn{4}{c|}{7} & 7 \\
    Shared experts $S$ & \multicolumn{4}{c|}{1} & 1 \\
    Expert width & \multicolumn{4}{c|}{256} & 256 \\
    Activated parameters & 0.17B & 0.32B & 0.64B & 1.1B & 1.1B \\
    Total parameters & 0.20--2.25B & 0.39--3.90B & 0.75--7.10B & 1.28--11.2B & 11.2B \\
    \hline \grouprow \multicolumn{6}{|l|}{\textbf{\textit{Training}}} \\
    Data & \multicolumn{4}{c|}{CC12M} & internal \\
    Resolution & \multicolumn{4}{c|}{$512{\times}512$} & $256^2 \rightarrow 512^2 \rightarrow 1024^2$, multi-aspect \\
    Batch size & 160 & 320 & 640 & 1152 & 1024 $\rightarrow$ 512 \\
    Steps & \multicolumn{4}{c|}{200K} & 540K \\
    Learning rate, MuonH / AdamW & \multicolumn{4}{c|}{$2.6{\times}10^{-3}$ / $2{\times}10^{-3}$, linear to 5\%} & $2.6{\times}10^{-3}$ / $2{\times}10^{-3}$, decayed to 0 \\
    REPA block & 5 & 6 & 8 & 11 & 11 \\
    \hline
  \end{tabular}}
\end{table}

\paragraph{Scaling study.} We define four MiniT2I scales from S to XL with our own widths and depths and a patch size of 32 (Table~\ref{tab:setting-t2i}); they are not the released MiniT2I-B/16 and MiniT2I-L/16 configurations~\cite{wang2026minit2i}.
At each scale, we train a dense baseline and MoEs across six values of $N$, both without and with REPA, using 1K warmup steps.
Following the scaling ladder of ABRA~\cite{chickering2026abra}, we keep training steps fixed and scale batch size with model size.
Every scale thus trains at a fixed ratio of roughly 50 image tokens per parameter of the dense model (TPP), excluding the text encoder; each MoE activates about as many parameters as its dense counterpart.
We sweep the learning rate on dense models and find that the optimal value stays fixed across scales under MuonH~\cite{wen2026hyperball}, analogous to~\cite{chickering2026abra}, so we use this single rate for all scales (Sec.~\ref{app:predictable}).
Models are compared by validation flow-matching loss on 1,024 held-out CC12M images from 16 excluded chunks, averaged over five fixed noise levels with fixed per-image noise.

\paragraph{System-level model.} For Table~\ref{tab:t2i-headline}, we train the XL MoE with $N{=}512$ and patch size 16 on an internal dataset using aspect-ratio bucketing.
Pre-training has three stages at $256^2$, $512^2$, and $1024^2$ pixel-area budgets, for 200K, 150K, and 100K steps at batch sizes 1024, 1024, and 512, on about 102M, 19M, and 9M image--text pairs.
Each stage starts with 1K--2K warmup steps, and the learning rate decreases within and across stages.
REPA is applied during the $256^2$ stage and turned off at higher resolutions.
As in MiniT2I~\cite{wang2026minit2i}, we then fine-tune the model on internal high-quality data for 90K steps at batch size 512.

\paragraph{Evaluation.} Table~\ref{tab:t2i-headline} reports QBench~\cite{li2026qwenimagebench}, GenEval~\cite{ghosh2023geneval}, and DPG~\cite{hu2024ella}.
Following MiniT2I~\cite{wang2026minit2i}, our model is sampled with 100 Euler steps and the original prompts on all three benchmarks, with a negative prompt for QBench.
The text-to-image visual results in Fig.~\ref{fig:headline} and Figs.~\ref{fig:t2i-gallery-places}--\ref{fig:t2i-gallery-text} are generated with the same 100 Euler steps, guidance scales between 2 and 6, and nine aspect ratios at about one megapixel.
QBench scores images with its judge model, Q-Judger. We score every open-source model ourselves on its 1,000 English prompts with the official aggregation. Q-Judger sees the $1024{\times}1024$ rows at native resolution, so their scores are not comparable with the $512{\times}512$ rows.

\paragraph{Baselines.} The open-source baselines use their official default sampling settings and the original prompts, with three exceptions.
PixelFlow~\cite{chen2025pixelflow} and PixelDiT-T2I~\cite{yu2025pixeldit} are sampled at $1024{\times}1024$ and downsampled to $512{\times}512$ before scoring.
Qwen-Image~\cite{wu2025qwenimage} and Qwen-Image-2.1~\cite{qwen2026qwenimage21} are sampled at $1024{\times}1024$, below their recommended $1328{\times}1328$ and $2048{\times}2048$, and Qwen-Image-2.1 uses its default 40 steps without guidance.
Models marked $^\dagger$ are sampled from prompts rewritten by their official rewriter and judged against the original prompts; for BAGEL~\cite{deng2025bagel}, this applies to GenEval only.
GenEval and DPG of the baselines are from each paper where reported and otherwise from later comparisons~\cite{yu2025pixeldit,zeng2026i1}.
Those of the pixel-space baselines at $1024{\times}1024$, L2P~\cite{l2p2026}, AsymFLUX.2 klein~\cite{asymflow2026}, Iris-3B~\cite{iris2026}, and Z-Image-Pixel~\cite{zimagepixel2026}, are taken from each paper or model card.
Those of Qwen-Image-2.1 are not reported, so we measure them with the same sampling settings as for QBench.
``--'' means not disclosed, or not scored for a closed-source model.

\subsection{Class-conditional}
\label{app:setup-c2i}
Table~\ref{tab:setting-c2i} lists the three ImageNet~\cite{deng2009imagenet} $256{\times}256$ Repr-MoE models in Table~\ref{tab:c2i-headline}, built on PixelDiT-XL in pixel space and DiT$^\text{DH}$-XL on RAE~\cite{zheng2025rae} and RAEv2~\cite{singh2026raev2} latents.
All three use 1K warmup steps and decay the learning rate to zero.

\begin{table}[!b]
  \centering
  \footnotesize
  \caption{\textbf{Class-conditional Repr-MoE configurations on ImageNet $256{\times}256$} (Table~\ref{tab:c2i-headline}). Parameter counts exclude frozen encoders and decoders. Depth lists patch and pixel blocks (26+4) or encoder blocks and the 2048-wide DH head (28+2). The dense parameters of RAEv2 are measured on its released checkpoint (875M); its paper reports 839M.}
  \label{tab:setting-c2i}
  \arrayrulecolor{black}\renewcommand{\arraystretch}{1.22}
\setlength{\tabcolsep}{5pt}
\resizebox{\textwidth}{!}{%
\begin{tabular}{|l|*{3}{w{c}{4.1cm}}|}
    \hline
    \rule{0pt}{2.4ex}Setting & Repr-MoE (PixelDiT) & Repr-MoE (RAE) & Repr-MoE (RAEv2) \\
    \hline \grouprow \multicolumn{4}{|l|}{\textbf{\textit{Architecture}}} \\
    Backbone & PixelDiT-XL & DiT$^\text{DH}$-XL & DiT$^\text{DH}$-XL \\
    Space & pixel & DINOv2-B latent & DINOv3-L latent \\
    Patch size & 16 & 1 & 1 \\
    Width & 1152 & 1152 & 1440 \\
    Depth & 26 + 4 & 28 + 2 & 28 + 2 \\
    Dense parameters & 0.80B & 0.84B & 0.88B \\
    \hline \grouprow \multicolumn{4}{|l|}{\textbf{\textit{MoE layer}}} \\
    Heads $H$ & 8 & 4 & 4 \\
    Experts per head $N$ & \multicolumn{3}{c|}{256} \\
    Routed experts $K$ & \multicolumn{3}{c|}{7} \\
    Shared experts $S$ & \multicolumn{3}{c|}{1} \\
    Expert width & \multicolumn{3}{c|}{288} \\
    Activated parameters & 0.85B & 0.85B & 0.82B \\
    Total parameters & 6.80B & 7.54B & 9.18B \\
    \hline \grouprow \multicolumn{4}{|l|}{\textbf{\textit{Training}}} \\
    Batch size & \multicolumn{3}{c|}{1024} \\
    Epochs & 320 & 320 & 80 \\
    Learning rate, MuonH / AdamW & \multicolumn{3}{c|}{$4{\times}10^{-3}$ / $2{\times}10^{-4}$, decayed to 0} \\
    REPA encoder, $\lambda$, block & DINOv2-B, 0.1, 8 & none (semantic latent) & none (semantic latent) \\
    \hline
  \end{tabular}}
\end{table}

For evaluation on ImageNet, Table~\ref{tab:c2i-headline} reports FID-50K~\cite{heusel2017fid} and FDr$^6$~\cite{yang2026fdloss} without and with guidance using EMA weights.
FDr$^6$ is the Fr\'echet distance to the training set divided by that of the validation set, averaged over six encoders.
We use EMA decay 0.9996 for Repr-MoE (PixelDiT) and 0.9995 for Repr-MoE (RAE) and Repr-MoE (RAEv2).
Repr-MoE (PixelDiT) is sampled with 100 FlowDPMSolver steps and CFG.
Repr-MoE (RAE) and Repr-MoE (RAEv2) use 50 and 100 Euler steps, respectively, with the guidance methods of their baselines, AutoGuidance~\cite{zheng2025rae} and Internal Guidance~\cite{singh2026raev2}. AutoGuidance uses the guide model released with RAE, DiT$^{\mathrm{DH}}$-S after 14 epochs.
For each of our rows, we report the guidance scale with the lowest FID: 3.0, 1.42, and 2.0, respectively.
Unless noted otherwise, 400K controlled comparisons report unguided FID-50K, while Fig.~\ref{fig:c2i-pareto} uses each model's best guidance scale.
For Fig.~\ref{fig:c2i-convergence}, we find by log-log interpolation between adjacent checkpoints when Repr-MoE matches the 800-epoch FID of PixelDiT or RAE, and report 800 over those epochs.

\section{MoE design and optimization}
\label{app:arch-opt}

\subsection{MoE layer parameter accounting}
\label{app:moe-design}
In Table~\ref{tab:roadmap}, every design step from a vanilla top-1 MoE ($N{=}8$) to fine-grained ($N{=}64$, top-8), shared ($K{+}S{=}7{+}1$), head chunking ($H{=}8$), and latent projection keeps the activated parameters of the MoE layer per token, excluding the router, equal to one dense FFN of size $F{=}8d^2$.
With $H$ independently routed heads, each head maintains $N$ experts and activates $K$ of them on a $d/H$-dimensional input, so activated expert parameters remain unchanged.
The gates of a head are the router scores of its selected experts, normalized to sum to one and scaled by $\sqrt{K}$.
Introducing the two $d{\times}d$ latent projections $W_\mathrm{in}$ and $W_\mathrm{out}$ adds $2d^2 = F/4$ parameters per layer.
To hold activated parameters constant, we narrow every expert to $3/4$ width, reducing total parameters by 16--17\%. The reduction is smaller than a quarter because the experts hold only 73--79\% of the total parameters, the rest of the network and the routers are unchanged, and the projections add parameters back.
The router scores $NH$ experts per layer and adds under 3\% to activated parameters at $N{=}64$ in Table~\ref{tab:roadmap}, where both pixel and latent models use $H{=}8$.
The router grows with $N$: at $N{=}256$ in Table~\ref{tab:c2i-pareto}, it accounts for the 8--21\% by which the Repr-MoEs exceed the activated parameters of the dense models.
The router design differs between settings and is fixed within each. It is the only part of the activated parameters that grows with $N$: relative to the activated parameters without it, it adds up to about 22\% at $N{=}512$ and 43\% at $N{=}1024$ in our B-scale ImageNet models, and under 2\% in MiniT2I up to $N{=}512$. The models without REPA carry the same router yet stop improving beyond $N{=}256$.

\subsection{Optimizer and training stability}
\label{app:opt}
We apply MuonH~\cite{wen2026hyperball} to the attention, feed-forward, and expert weight matrices of all transformer blocks.
The router, biases, normalization gains, and all parameters outside the transformer blocks use AdamW.
Keeping the router on AdamW follows recent language MoEs~\cite{qwen2026next,xiaomi2026mimo}, as Muon can exacerbate early-training fluctuations and destabilize routing~\cite{qwen2026next}.
Where a model uses adaLN modulation, its weights in patch-level blocks are trained with plain Muon~\cite{jordan2024muon}.
Dense and MoE models share the same learning rates, both without and with REPA.
On the fused QKV projection, per-head orthogonalization~\cite{kimi2026k3} achieves lower training loss than separate Q, K, and V updates~\cite{chen2026cmuon}, and both beat the merged update.

\begin{figure}[!t]
  \centering
  \includegraphics[width=\linewidth]{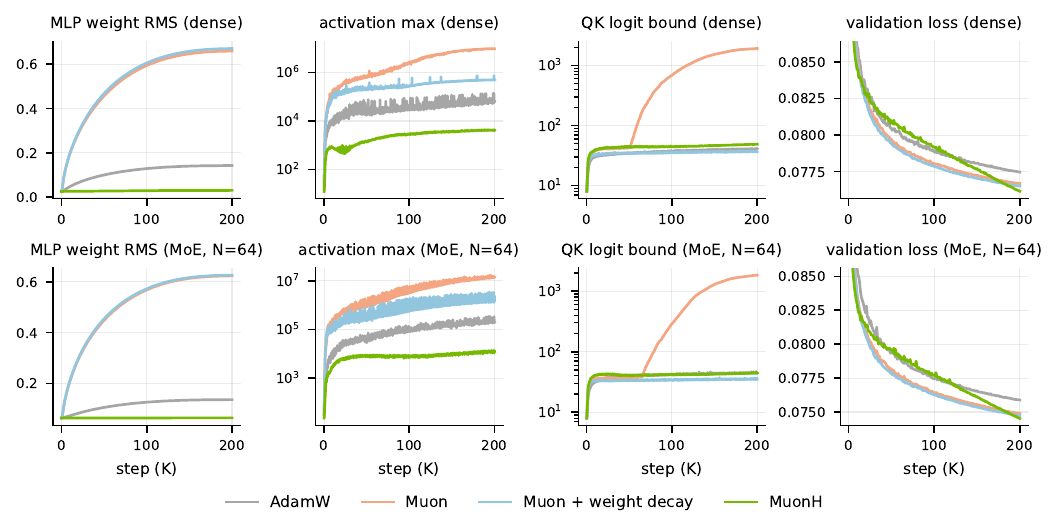}
  \caption{\textbf{Training stability under different optimizers} (MiniT2I, scale M, dense and MoE $N{=}64$). Plain Muon allows weights, activations, and attention logits to grow continuously. Weight decay bounds only attention logits, while MuonH keeps all three bounded and reaches the lowest final validation loss (rightmost column).}
  \label{fig:abl-opt}
\end{figure}

Fig.~\ref{fig:abl-opt} compares optimizer stability and validation loss on MiniT2I at scale M with REPA.
Under plain Muon, weight RMS, maximum activations, and attention logits grow throughout training.
Adding weight decay of 0.01 to the normalization gains, which Moonlight~\cite{liu2025moonlight} finds crucial for stability, bounds the attention logits, but weights and activations still drift upward.
MuonH keeps all three quantities bounded.
In the rightmost column, MuonH trails slightly during mid-training but overtakes the other optimizers after learning-rate decay to reach the lowest final validation loss.
Comparing the two rows, MoE achieves lower validation loss than dense under every optimizer.

To test whether these optimizer effects hold generally, Table~\ref{tab:abl-opt} compares AdamW and Muon variants across both pixel and latent spaces, without and with REPA.
Across all four settings, Muon-family optimizers consistently widen MoE's gain over dense models from 12--20\% under AdamW to 24--40\%, and MuonH outperforms Muon with weight decay in three of the four settings.
Each optimizer uses a single learning rate for both dense and MoE models.
All Muon variants use $2{\times}10^{-3}$, tuned for MuonH on the dense models.
For AdamW, we use $4{\times}10^{-4}$ on SiT-B/2 and $10^{-4}$ on PixelDiT-B/16, the best tested rates that train stably both without and with REPA.

\begin{table}[H]
  \centering
  \small
  \caption{\textbf{Muon helps MoE more than dense.} Unguided FID-50K at 400K steps on ImageNet $256{\times}256$ for PixelDiT-B/16 in pixel space and SiT-B/2 on FLUX.1 latents ($N{=}128$). WD: weight decay; $\Delta$: (MoE$-$dense)/dense.}
  \label{tab:abl-opt}
  \arrayrulecolor{black}\setlength{\tabcolsep}{5pt}
  \setlength{\aboverulesep}{0pt}\setlength{\belowrulesep}{0pt}\renewcommand{\arraystretch}{1.0}
\begin{tabular}{|l|*{3}{w{c}{1.45cm}}|*{3}{w{c}{1.45cm}}|}
    \hline
    \multicolumn{1}{|l|}{} & \multicolumn{3}{c|}{Without REPA} & \multicolumn{3}{c|}{With REPA} \\
    \cline{2-7}
    Optimizer & Dense & MoE & $\Delta$ & Dense & MoE & $\Delta$ \\
    \hline \grouprow \multicolumn{7}{|l|}{\textbf{\textit{Pixel space}}} \\
    AdamW & 69.30 & 60.98 & $-$12\% & 46.38 & 40.27 & $-$13\% \\
    Muon & diverged & diverged & -- & 35.85 & 24.22 & $-$32\% \\
    Muon+WD & 51.83 & 34.85 & $-$33\% & 34.08 & 20.61 & $-$40\% \\
    MuonH & 43.69 & 33.35 & $-$24\% & 30.65 & 20.36 & $-$34\% \\
    \hline \grouprow \multicolumn{7}{|l|}{\textbf{\textit{Latent space}}} \\
    AdamW & 55.42 & 44.14 & $-$20\% & 36.62 & 30.31 & $-$17\% \\
    Muon & 44.69 & 32.55 & $-$27\% & 27.89 & 21.33 & $-$24\% \\
    Muon+WD & 40.41 & 25.68 & $-$36\% & 24.35 & 15.74 & $-$35\% \\
    MuonH & 40.91 & 26.53 & $-$35\% & 23.52 & 14.79 & $-$37\% \\
    \hline
  \end{tabular}
\end{table}

\section{Additional quantitative results}
\label{app:scaling}

\subsection{Representation breaks the Sparsity Ceiling}
\label{app:teachers}
\label{app:guided-n}
Table~\ref{tab:teacher-n} lists the values behind Fig.~\ref{fig:teacher-n}: the unguided FID for each encoder and expert count, in pixel and latent space. Without REPA, FID changes by less than 1\% from 256 to 1024 experts, and with each of the seven encoders it falls by 6.6--22.7\%, although with MAE-L in pixel space it stops improving beyond 512 experts.

\begin{table}[!t]
  \centering
  \small
  \setlength{\tabcolsep}{5pt}
  \caption{\textbf{REPA breaks the Sparsity Ceiling across encoders.} Unguided FID-50K at 400K steps on ImageNet $256{\times}256$ for $N{=}256$, 512 and 1024. $\Delta$: relative change from $N{=}256$ to 1024. \csq{black!9}~without REPA, \csq{nvgreenlight}~with seven encoders.}
  \label{tab:teacher-n}
  \arrayrulecolor{black}\setlength{\aboverulesep}{0pt}\setlength{\belowrulesep}{0pt}\renewcommand{\arraystretch}{1.12}
\begin{tabular}{|w{l}{2.9cm}|*{3}{w{c}{1.05cm}}|w{c}{1.05cm}|*{3}{w{c}{1.05cm}}|w{c}{1.05cm}|}
    \hline
     & \multicolumn{4}{c|}{PixelDiT-B/16 (pixel)} & \multicolumn{4}{c|}{SiT-B/2 on FLUX.1 (latent)} \\
    \cline{2-9}
    Encoder & $N{=}256$ & 512 & 1024 & $\Delta$ & $N{=}256$ & 512 & 1024 & $\Delta$ \\
    \hline
    \rowcolor{black!9}Without REPA & 31.37 & 31.97 & 31.61 & +0.8\% & 24.82 & 24.97 & 25.03 & +0.8\% \\
    \hline
    \rowcolor{nvgreenlight}DINOv2 & 18.33 & 16.70 & 15.24 & $-$16.9\% & 13.83 & 12.82 & 11.43 & $-$17.4\% \\
    \rowcolor{nvgreenlight}DINOv3 & 18.79 & 16.94 & 14.53 & $-$22.7\% & 14.33 & 12.96 & 12.18 & $-$15.0\% \\
    \rowcolor{nvgreenlight}CLIP-L & 21.16 & 19.67 & 18.88 & $-$10.8\% & 16.12 & 15.32 & 14.61 & $-$9.4\% \\
    \rowcolor{nvgreenlight}SigLIP~2 & 18.51 & 17.24 & 16.70 & $-$9.8\% & 14.74 & 13.52 & 12.73 & $-$13.6\% \\
    \rowcolor{nvgreenlight}PE & 18.74 & 17.31 & 16.25 & $-$13.3\% & 15.37 & 14.21 & 13.00 & $-$15.4\% \\
    \rowcolor{nvgreenlight}MoonViT-V2 & 22.57 & 20.70 & 18.16 & $-$19.5\% & 17.04 & 15.34 & 14.38 & $-$15.6\% \\
    \rowcolor{nvgreenlight}MAE-L & 22.58 & 20.98 & 21.09 & $-$6.6\% & 17.70 & 17.23 & 16.54 & $-$6.6\% \\
    \hline
  \end{tabular}
\end{table}

The results above and Table~\ref{tab:semvae} are without guidance.
To verify that our findings hold under guidance, Table~\ref{tab:guided-n} evaluates the latent-space models at their best guidance scales, using CFG for FLUX.1 and FLUX.2 and AutoGuidance or Internal Guidance for RAE and RAEv2.
We reach the same conclusion without and with guidance: on non-semantic FLUX.1 latents without REPA, FID stalls beyond $N{=}256$ (+2.3\% from 256 to 1024), whereas with DINOv2 REPA or semantic latents (FLUX.2, RAE, RAEv2) FID keeps falling with every increase in $N$, by 6.7--9.8\% from 256 to 1024.

\begin{table}[H]
  \centering
  \small
  \setlength{\tabcolsep}{5pt}
  \caption{\textbf{No Sparsity Ceiling with semantic grounding under guidance up to $N{=}1024$.} Guided FID-50K at 400K steps on ImageNet $256{\times}256$; compare trends within a row. Each model uses its best guidance scale, selected on FID-10K with a separate seed. $\Delta$: relative change from $N{=}256$ to 1024. RAE and RAEv2 use DiT$^\text{DH}$-B~\cite{zheng2025rae}. \csq{black!9}~non-semantic, \csq{nvgreenlight}~semantically grounded.}
  \label{tab:guided-n}
  \arrayrulecolor{black}\setlength{\aboverulesep}{0pt}\setlength{\belowrulesep}{0pt}\renewcommand{\arraystretch}{1.12}
\begin{tabular}{|w{l}{2.9cm}|w{c}{1.25cm}|*{5}{w{c}{1.25cm}}|w{c}{1.25cm}|}
    \hline
    Latent space & Dense & $N{=}64$ & 128 & 256 & 512 & 1024 & $\Delta$ \\
    \hline
    \rowcolor{black!9}FLUX.1 (reference) & 7.86 & 5.25 & 4.94 & 4.35 & 4.37 & 4.45 & +2.3\% \\
    \hline
    \rowcolor{nvgreenlight}FLUX.1 + REPA & 2.92 & 2.40 & 2.36 & 2.29 & 2.23 & 2.14 & $-$6.7\% \\
    \rowcolor{nvgreenlight}FLUX.2 & 4.71 & 3.24 & 3.10 & 2.83 & 2.70 & 2.63 & $-$6.9\% \\
    \hline
    \rowcolor{nvgreenlight}RAE & 2.15 & 1.81 & 1.66 & 1.58 & 1.51 & 1.43 & $-$9.8\% \\
    \rowcolor{nvgreenlight}RAEv2 & 2.89 & 2.25 & 2.19 & 2.02 & 1.92 & 1.88 & $-$7.0\% \\
    \hline
  \end{tabular}
\end{table}

\subsection{Dense scaling and equivalent compute}
\label{app:predictable}
\begin{wrapfigure}[17]{r}{0.5\textwidth}
  \centering
  \vspace{-8pt}
  \includegraphics[width=0.5\textwidth]{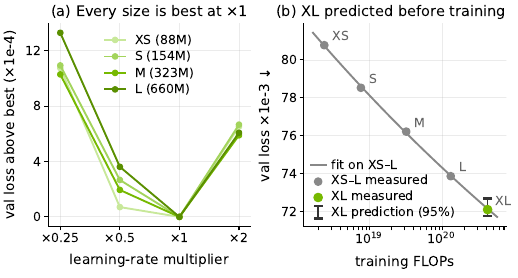}
  \caption{\textbf{Learning rate and loss across dense models of increasing size.} All models are trained with REPA. (a) Validation loss above each size's best, against the learning-rate multiplier ($\times1 = 2.6{\times}10^{-3}$). (b) Validation loss against training compute, with the XL prediction made before XL trained.}
  \label{fig:lr-ladder}
  \vspace{-6pt}
\end{wrapfigure}
This section gives the learning-rate sweep and the power-law fit behind the dense-equivalent compute in Fig.~\ref{fig:saturation}.
For the MiniT2I scaling study, we sweep the learning rate at $\times0.25$, $\times0.5$, $\times1$, and $\times2$ of a base MuonH rate of $2.6{\times}10^{-3}$ on dense models with REPA across scales XS, S, M, and L.
Here XS is an additional smaller dense scale not included in Fig.~\ref{fig:saturation}, with width 640, depth 8, and 88M parameters.
Every size is best at $\times1$ (Fig.~\ref{fig:lr-ladder}a), so we use this learning rate for all sizes, dense and MoE.

With this learning rate, validation loss follows a power law in training compute, $L = aC^{-b} + c$, where $C$, the training FLOPs, is six times the activated parameters times the image tokens processed.
We add XS so that this three-parameter fit has four scales before training XL.
We fitted it on XS to L and fixed the prediction for XL before training it: a validation loss of 0.07215, with a 95\% interval of 0.07178--0.07268 from a parametric bootstrap.
XL reached 0.07209 (Fig.~\ref{fig:lr-ladder}b).

The bottom panel of Fig.~\ref{fig:saturation} reads each MoE's dense-equivalent compute from this power law, fitted separately without and with REPA to the five dense models.
Every XL MoE, and the L MoEs with REPA at $N{=}256$ and $512$, reach a lower loss than the largest dense model, so their values extrapolate the fit beyond it, by up to about $3.5\times$ in compute.
We keep TPP fixed across scales, and in this setting the power law predicted the XL loss, at $3.2\times$ the compute of L, before training (Fig.~\ref{fig:lr-ladder}b), supporting extrapolation over this range.
Scaling studies of dense text-to-image diffusion transformers also find that the loss follows a power law in compute over several orders of magnitude~\cite{liang2026ditscaling,chickering2026abra}.

\enlargethispage{1.5\baselineskip}
\subsection{Compute efficiency on ImageNet}
\label{app:pareto}
Table~\ref{tab:c2i-pareto} lists the models behind Fig.~\ref{fig:c2i-pareto}, all trained on FLUX.1 latents with REPA for 400K steps, with their sizes, per-image GFLOPs, and FID-50K without and with guidance.
These dense models are trained independently of the MuonH dense SiT-B/2 in Tables~\ref{tab:roadmap}, \ref{tab:abl-opt} and~\ref{tab:guided-n}, with the same recipe; Dense-B reaches 23.75 and 2.96 here against 23.52 and 2.92 there, a run-to-run difference of 1--1.5\%.
Equivalent dense compute is obtained by log-log interpolation between the two nearest dense models, yielding 1.74--2.00$\times$ without guidance and 1.86--1.99$\times$ with guidance.

\begin{table}[H]
  \centering
  \small
  \setlength{\tabcolsep}{5pt}
  \caption{\textbf{Dense models and Repr-MoE} (Fig.~\ref{fig:c2i-pareto}). Repr-MoE uses $N{=}256$, $H{=}8$, and $K{+}S{=}8$, with the same activated FFN as dense. Equiv.\ dense is the dense GFLOPs at equal FID divided by the MoE GFLOPs.}
  \label{tab:c2i-pareto}
  \adjustbox{max width=\textwidth}{%
  \arrayrulecolor{black}\setlength{\aboverulesep}{0pt}\setlength{\belowrulesep}{0pt}\renewcommand{\arraystretch}{1.12}
\begin{tabular}{|l|ccc|cc|cc|}
  \hline
   & & & & \multicolumn{2}{c|}{w/o guidance} & \multicolumn{2}{c|}{w/ guidance} \\
  \cline{5-8}
  Model & Act. & Total & GFLOPs & FID$\downarrow$ & Equiv.\ dense & FID$\downarrow$ & Equiv.\ dense \\
  \hline
  Dense-S  & 33M  & 33M  & 12.1  & 51.83 & -- & 6.63 & -- \\
  Dense-B  & 131M & 131M & 46.0  & 23.75 & -- & 2.96 & -- \\
  Dense-L  & 458M & 458M & 161.4 & 9.98  & -- & 1.95 & -- \\
  Dense-XL & 675M & 675M & 237.3 & 8.34  & -- & 1.77 & -- \\
  Dense-G  & 1.15B & 1.15B & 402.4 & 6.96 & -- & 1.62 & -- \\
  \hline
  Repr-MoE-S & 40M  & 289M  & 15.8  & 32.14 & 1.74$\times$ & 3.80 & 1.94$\times$ \\
  Repr-MoE-B & 145M & 1.14B & 53.3  & 13.83 & 1.89$\times$ & 2.29 & 1.86$\times$ \\
  Repr-MoE-L & 494M & 3.84B & 179.7 & 7.24  & 2.00$\times$ & 1.65 & 1.99$\times$ \\
  \hline
\end{tabular}
}
\end{table}

\subsection{Detailed text-to-image benchmarks}
\label{app:prism}
Table~\ref{tab:t2i-prism} reports the five-dimension QBench breakdown at $1024{\times}1024$ alongside GenEval and DPG from Table~\ref{tab:t2i-headline}, as well as PRISM-Bench~\cite{fang2026prism} and LongText-Bench~\cite{geng2025xomni}.
Baseline scores for PRISM-Bench and LongText-Bench are taken from i1~\cite{zeng2026i1}, except for Qwen-Image-2.1, L2P, AsymFLUX.2 klein, and Iris-3B, which we evaluate ourselves with the official judges at their official default sampling settings.
On PRISM-Bench and LongText-Bench, Repr-MoE is sampled at $1024{\times}1024$ with the original prompts and no negative prompt.
The last row of Table~\ref{tab:t2i-prism} uses a single conservative setting on all benchmarks, guidance scale 4 and no negative prompt, as in Fig.~\ref{fig:t2i-compare}; its scores drop slightly but stay above those of all pixel-space baselines except on LongText-Bench.
On QBench, Repr-MoE ranks first among pixel-space models and second overall in every dimension, behind only Qwen-Image-2.1, which uses an LLM prompt rewriter.
On PRISM-Bench, Repr-MoE scores 76.0, the highest among pixel-space models and second overall, close to Qwen-Image-2.1 at 76.5.
On LongText-Bench, it reaches 0.870, again the highest among pixel-space models, while trailing the top latent models.

\begin{table}[tbp]
  \centering
  \footnotesize
  \setlength{\tabcolsep}{3.6pt}
  \caption{\textbf{Detailed text-to-image benchmarks at $1024{\times}1024$.} Act.\ and Total: activated and total parameters (B). $^\dagger$Evaluated with the model's official prompt rewriter (GenEval only for BAGEL). Best and second best in each column are bold and underlined. The last row (not ranked) uses guidance scale 4 and no negative prompt.}
  \label{tab:t2i-prism}
  \arrayrulecolor{black}\renewcommand{\arraystretch}{1.12}
  \setlength{\aboverulesep}{0pt}\setlength{\belowrulesep}{0pt}
  \resizebox{\textwidth}{!}{%
\begin{tabular}{|l|c|cc|cccccc|cc|cc|}
    \hline
     & & & & \multicolumn{6}{c|}{QBench$\uparrow$} & & & & \\
    \cmidrule{5-10}
    Method & Space & Act. & Total & Overall & Aesth. & Align. & Creat. & Qual. & Real. & GenEval$\uparrow$ & DPG$\uparrow$ & PRISM$\uparrow$ & LongText$\uparrow$ \\
    \hline
    GPT Image 1 [High]~\cite{openai2025gptimage1} & -- & -- & -- & -- & -- & -- & -- & -- & -- & 0.84 & 85.15 & -- & \secondv{0.956} \\
    Seedream 3.0~\cite{gao2025seedream3} & -- & -- & -- & -- & -- & -- & -- & -- & -- & 0.84 & \secondv{88.27} & -- & 0.896 \\
    FLUX.1 [Dev]~\cite{bfl2024flux1} & Latent & 12 & 12 & 43.8 & 48.9 & 42.1 & 35.1 & 47.1 & 40.4 & 0.66 & 83.84 & 65.1 & 0.607 \\
    SD3 Medium~\cite{esser2024sd3} & Latent & 2 & 2 & 41.3 & 42.9 & 40.6 & 31.0 & 46.6 & 40.1 & 0.62 & 84.08 & 61.9 & 0.322 \\
    BAGEL$^{\dagger}$~\cite{deng2025bagel} & Latent & 7 & 14 & 42.8 & 46.5 & 42.4 & 34.8 & 45.2 & 40.0 & 0.88 & 85.44 & 61.8 & 0.373 \\
    HiDream-I1-Full~\cite{cai2025hidream} & Latent & -- & 17 & 46.8 & 51.3 & 46.1 & 38.8 & 49.2 & 42.6 & 0.83 & 85.89 & 66.1 & 0.543 \\
    Lumina-Image 2.0~\cite{qin2025lumina2} & Latent & 3 & 3 & 41.2 & 44.7 & 39.6 & 30.8 & 45.2 & 39.8 & 0.73 & 87.20 & 63.5 & 0.088 \\
    Z-Image~\cite{zimage2025} & Latent & 6 & 6 & 48.4 & 49.3 & 49.6 & 43.9 & 49.5 & 46.5 & 0.84 & 88.14 & 74.2 & 0.935 \\
    Qwen-Image~\cite{wu2025qwenimage} & Latent & 20 & 20 & 49.1 & 51.1 & 51.0 & 46.8 & 48.9 & 44.1 & 0.87 & \bestv{88.32} & 73.9 & 0.943 \\
    Qwen-Image-2.1$^{\dagger}$~\cite{qwen2026qwenimage21} & Latent & 7 & 7 & \bestv{59.2} & \bestv{62.2} & \bestv{59.7} & \bestv{63.2} & \bestv{56.4} & \bestv{53.9} & 0.81 & 84.81 & \bestv{76.5} & \bestv{0.959} \\
    BLIP3o-NEXT-G-G~\cite{chen2025blip3onext} & Latent & 3 & 3 & 41.3 & 42.8 & 41.0 & 31.1 & 45.8 & 39.7 & \secondv{0.91} & 81.93 & 57.5 & 0.114 \\
    i1$^{\dagger}$~\cite{zeng2026i1} & Latent & 3 & 3 & 47.1 & 49.0 & 48.0 & 43.0 & 48.5 & 43.1 & 0.84 & 86.73 & 70.1 & 0.922 \\
    L2P~\cite{l2p2026} & Pixel & 6 & 6 & 46.6 & 48.4 & 46.9 & 40.0 & 49.2 & 44.5 & 0.76 & 86.00 & 65.0 & 0.774 \\
    AsymFLUX.2 klein~\cite{asymflow2026} & Pixel & 9 & 9 & 46.2 & 47.2 & 46.5 & 40.8 & 48.4 & 45.8 & 0.82 & 86.8 & 72.5 & 0.770 \\
    Iris-3B~\cite{iris2026} & Pixel & 3 & 3 & 46.7 & 48.6 & 48.2 & 39.6 & 48.6 & 43.4 & 0.80 & 86.52 & 69.5 & 0.853 \\
    Z-Image-Pixel~\cite{zimagepixel2026} & Pixel & 6 & 6 & -- & -- & -- & -- & -- & -- & 0.76 & 87.60 & -- & -- \\
    \oursrow Repr-MoE (ours) & Pixel & 1.1 & 11.2 & \secondv{56.1} & \secondv{59.2} & \secondv{57.0} & \secondv{55.3} & \secondv{54.3} & \secondv{51.1} & \bestv{0.93} & 87.87 & \secondv{76.0} & 0.870 \\
    \oursrow \quad CFG 4, no negative prompt & Pixel & 1.1 & 11.2 & 54.2 & 56.3 & 56.9 & 52.7 & 52.5 & 48.9 & 0.92 & 87.69 & 75.0 & 0.843 \\
    \hline
  \end{tabular}}
\end{table}

\section{Routing and specialization analysis}
\label{app:routing}

\subsection{Effective number of experts}
\label{app:effn}
For each MoE layer, let $p_i$ be the fraction of top-$K$ selections routed to expert $i$ at the end of training over 2,048--4,096 images across training noise levels.
We measure utilization by the normalized effective number of experts, $N_{\mathrm{eff}}/N = \exp\!\big({-}\sum_i p_i \log p_i\big)/N \in [1/N, 1]$, which equals $1$ under uniform load and $m/N$ if routing collapses uniformly onto $m$ experts.
Values are averaged over routing heads for MiniT2I and PixelDiT and pooled across heads for SiT.

Without REPA, $N_{\mathrm{eff}}/N \ge 0.99$ across all 9 MoE layers of MiniT2I (scale M, $N{=}16\text{--}512$) and $\ge 0.986$ across all 10 of PixelDiT-B/16 ($N{=}32\text{--}1024$), whose layer average drifts only from 0.999 to 0.991. In SiT-B/2 on FLUX.1 latents ($N{=}256\text{--}1024$), it is $\ge 0.99$ in 8 of 9 MoE layers; the sole exception is its first MoE layer (block 4) at 0.48--0.56, which shows no downward trend with $N$: its effective expert count still grows roughly fourfold from $N{=}256$ to $1024$, while unguided FID changes by less than 1\%.
Thus, routing collapse does not explain the Sparsity Ceiling.

\pagebreak\subsection{Expert distinctness and class specialization}
\label{app:mech}
\begin{wrapfigure}{r}{0.49\textwidth}
  \centering
  \vspace{-8pt}
  \includegraphics[width=0.49\textwidth]{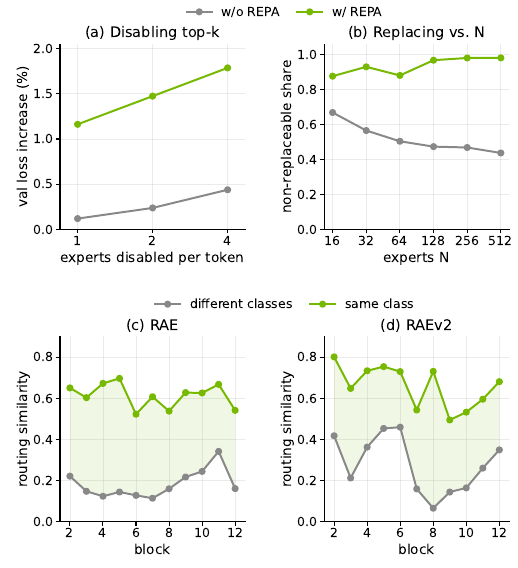}
  \caption{\textbf{Expert distinctness and specialization.} (a), (b): MiniT2I at scale M after 200K steps, aligned block 6, without and with REPA. (a) Validation loss increase when disabling each token's top-$k$ routed experts, $k \in \{1, 2, 4\}$, at $N{=}512$. (b) Non-replaceable share, the fraction of experts remaining at a 0.1\% loss increase, vs.\ $N$. (c), (d): routing similarity between images of the same class and of different classes in every MoE layer of RAE and RAEv2, with the images and $t$ of Fig.~\ref{fig:cluster}.}
  \label{fig:mech33}
  \vspace{-6pt}
\end{wrapfigure}
\paragraph{Expert distinctness.} To test whether experts become functionally distinct rather than redundant, we probe block 6, the aligned block of MiniT2I at scale M, with two retraining-free validation-loss tests (Fig.~\ref{fig:mech33}).
Following DeepSeekMoE~\cite{dai2024deepseekmoe}, \emph{disabling} removes each token's top-$k$ routed experts per head, routes to the next-ranked ones, and renormalizes the gates to measure reliance on chosen experts.
\emph{Replacing} tests whether one expert can cover another's tokens: we greedily take the pair of experts in a head with the smallest output difference on calibration images, remove one, and reroute its tokens to the other, with all weights and other routes fixed, until validation loss rises by 0.1\%.
The remaining fraction is the \emph{non-replaceable share}~\cite{engmann2026causal,li2024mcsmoe,chen2025hcsmoe}; a higher share means fewer redundant experts.

At $N{=}512$, disabling a token's top expert raises validation loss by 1.16\% with REPA and by 0.12\% without (Fig.~\ref{fig:mech33}a).
The top expert carries more router weight with REPA, 33.5\% against 20.8\%, but these peakier gates do not explain the gap: without REPA, disabling the top two experts removes 37.4\% of the weight and raises the loss by only 0.24\%.
Disabling all seven routed experts of a token in this block raises the loss by 2.19\% with REPA and 0.74\% without; disabling the top expert alone gives about half of this with REPA and less than one sixth without, at every $N$ from 64 to 512.
In the replacing test, about half of the experts can be replaced without REPA at every $N{\ge}64$, and the non-replaceable share is 0.44 at $N{=}512$. With REPA, almost every expert is non-replaceable, with a share of 0.98 at $N{=}512$ (Fig.~\ref{fig:mech33}b).

\paragraph{Specialization without an alignment loss.} Experts also specialize by class on RAE and RAEv2 latents without an alignment loss.
With the routing similarity of Fig.~\ref{fig:cluster} on the same 64 images at $t{=}0.5$ with heads concatenated, same-class images route more alike than different-class ones in every MoE layer (Fig.~\ref{fig:mech33}c,\,d).
Within-class similarity exceeds cross-class similarity by 0.43 on RAE and 0.38 on RAEv2, and at least 0.32 and 0.27 per layer, versus 0.21 on average for PixelDiT without REPA, which receives the same class labels.

\subsection{Concept-specialized experts in text-to-image generation}
\label{app:routing-t2i}
To test whether experts in our 11.2B text-to-image model (Table~\ref{tab:t2i-headline}) specialize in concrete visual concepts, we run a causal masking test on text rendering and portraits.
Using a localization prompt set, we rank experts in each MoE block (blocks 4--20) by their relative selection frequency on target-concept versus general prompts, and mask the router logits of the top $p\%$ per block before top-$K$ selection.
We evaluate on prompts not used for ranking with the same initial noise, against a default control that masks frequency-matched random experts in the same blocks.

\begin{figure}[H]
  \centering
  \includegraphics[width=\textwidth]{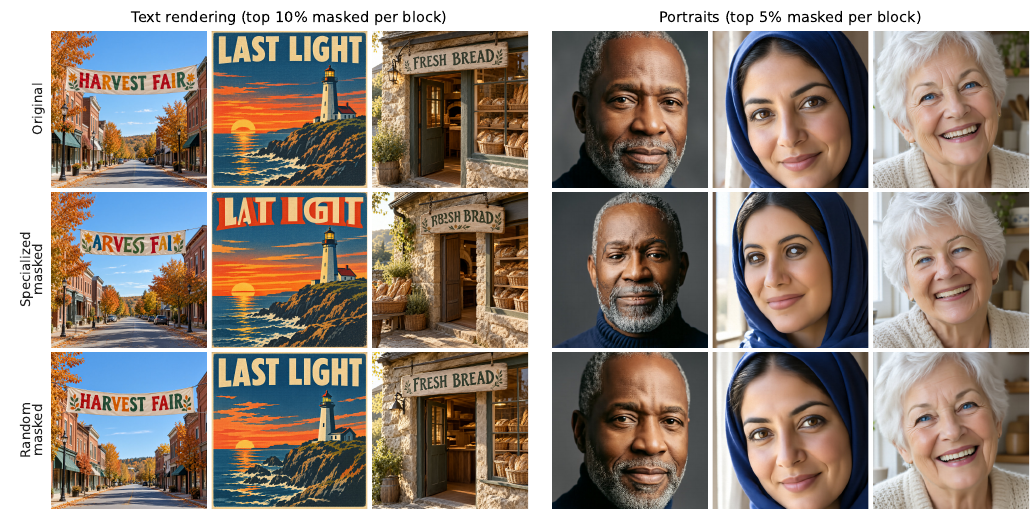}
  \caption{\textbf{Masking concept-specialized experts degrades text rendering and faces.} $1024{\times}1024$ generations from our 11.2B model with the same prompt and initial noise per column (portraits cropped to the face). Rows show the original model, top 10\% (text) or 5\% (portraits) specialized experts masked, and random experts masked. Masking specialized experts corrupts text (left) and distorts facial features, especially the eyes (right), while largely preserving the scene, whereas random masking has little effect.}
  \label{fig:knockout}
\end{figure}

\looseness=-1Masking the specialized experts corrupts rendered text and distorts facial features, especially the eyes, while largely preserving the surrounding scene, whereas random masking has little effect (Fig.~\ref{fig:knockout}).
Across 27 held-out text prompts at $512{\times}512$, OCR word accuracy drops from 0.81 to 0.62 at $p{=}1\%$ and to 0.08 at $p{=}10\%$, while the random control stays at 0.76--0.79.
Random experts that remove about the same total share of the expert output lower it only to 0.74 at $p{=}10\%$.
On unrelated prompts, masking the text-specialized experts changes the images (LPIPS) less than masking random experts, and leaves their CLIP score 0.1--0.3 points lower than random masking does.
Across 21 held-out portrait prompts, faces are still detected after masking, while CLIP score drops by 1.1, 1.9 and 3.4 points from 27.4 at $p{=}1$, 2 and 5\%, against at most 0.3 for random experts. Random experts removing at least the same total share of expert output across all blocks lower it by at most 0.1.
For both concepts, we cannot rule out that drops come from expert output removed in particular blocks.

\clearpage
\section{Additional visual results}
\label{app:visuals}

\subsection{Class-conditional generation}
\label{app:c2i-visual}
\begin{figure}[H]
  \centering
  \includegraphics[width=\textwidth,height=\dimexpr\textheight-122pt\relax,keepaspectratio,trim=23.5 1.7 23.5 0.5,clip]{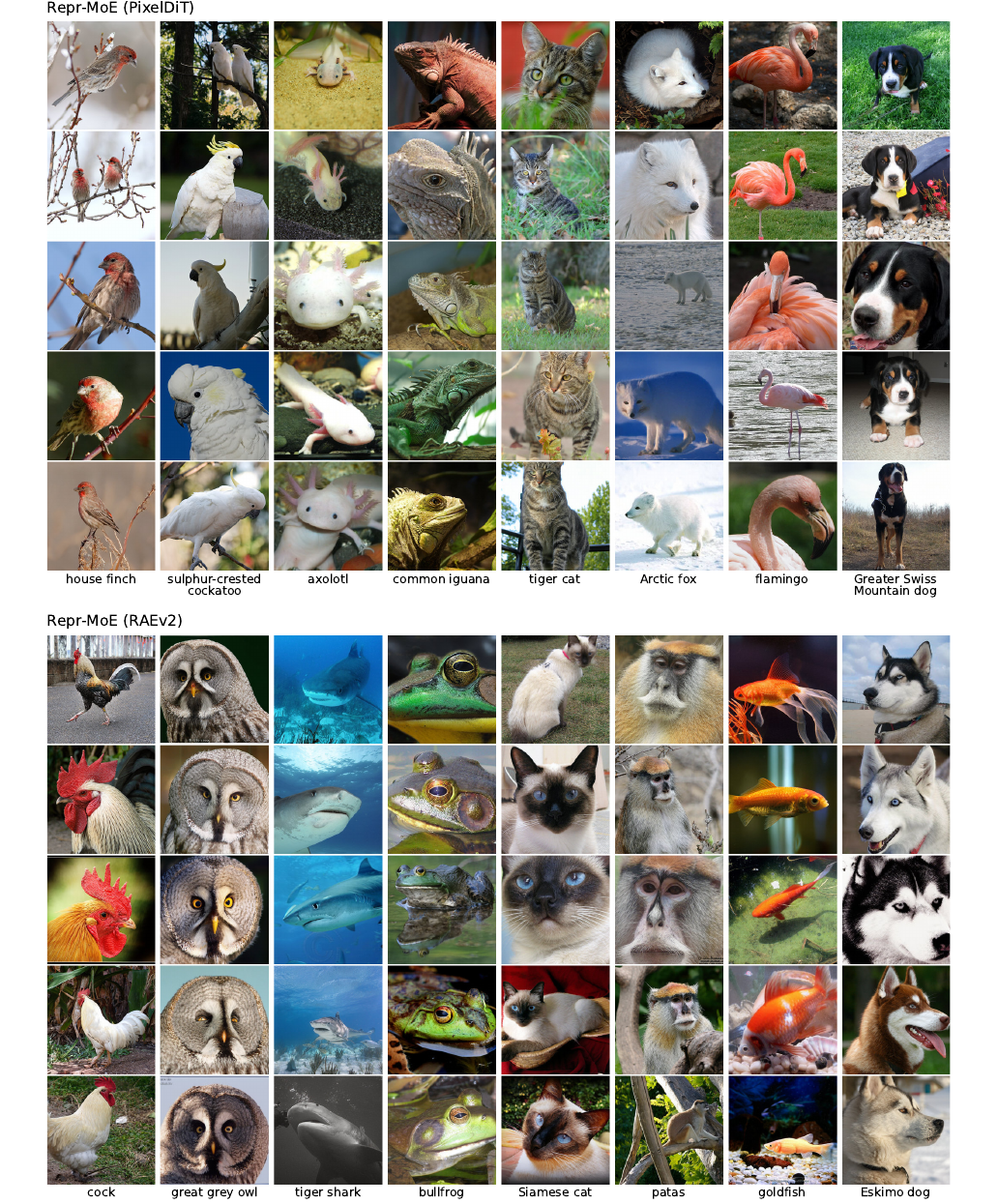}
  \caption{\textbf{Visual results of Repr-MoE on class-conditional generation.} ImageNet $256{\times}256$ results of the Repr-MoE models in Table~\ref{tab:c2i-headline}, with the guidance used there: PixelDiT after 320 epochs with CFG at scale 3.0 (top) and RAEv2 after 80 epochs with Internal Guidance at scale 2.0 (bottom). Each column shows five results of one class.}
  \label{fig:c2i-visual-app}
\end{figure}

Fig.~\ref{fig:c2i-visual-app} shows additional visual results of Repr-MoE on ImageNet $256{\times}256$.

\clearpage
\subsection{Text-to-image generation}
\label{app:t2i-visual}
\begin{figure}[H]
  \centering
  \scalebox{0.91}{\begin{minipage}{\textwidth}{\setlength{\tabcolsep}{0pt}\renewcommand{\arraystretch}{1}%
\def\cmpblk{\dimexpr0.5\textwidth-0.035in-0.2pt\relax}%
\def\cmpside{\dimexpr(\cmpblk-0.024in)/3\relax}%
\def\cmpsans{\fontencoding{T1}\fontfamily{DejaVuSans-TLF}\fontsize{7.5}{9}\selectfont}%
\def\cmpfont{\fontsize{7.40}{8.66}\selectfont}%
\def\cmpimg#1{\includegraphics[width=\cmpside]{figures/t2i_compare/#1}}%
\def\cmprow#1#2#3#4{\cmpimg{#1}\hfill\cmpimg{#2}\hfill\cmpimg{#3}\par\vspace{1.5pt}{\cmpfont\raggedright #4\par}}%
\def\cmphead{\makebox[\cmpside]{\cmpsans Repr-MoE (ours)}\hfill\makebox[\cmpside]{\cmpsans Qwen-Image-2.1$^{\dagger}$}\hfill\makebox[\cmpside]{\cmpsans AsymFLUX.2 klein}}%
\begin{tabular}{@{}p{\cmpblk}@{\hspace{0.035in}}!{\color{black!35}\vrule width 0.4pt}@{\hspace{0.035in}}p{\cmpblk}@{}}
\cmphead & \cmphead \\[1.5pt]
\cmprow{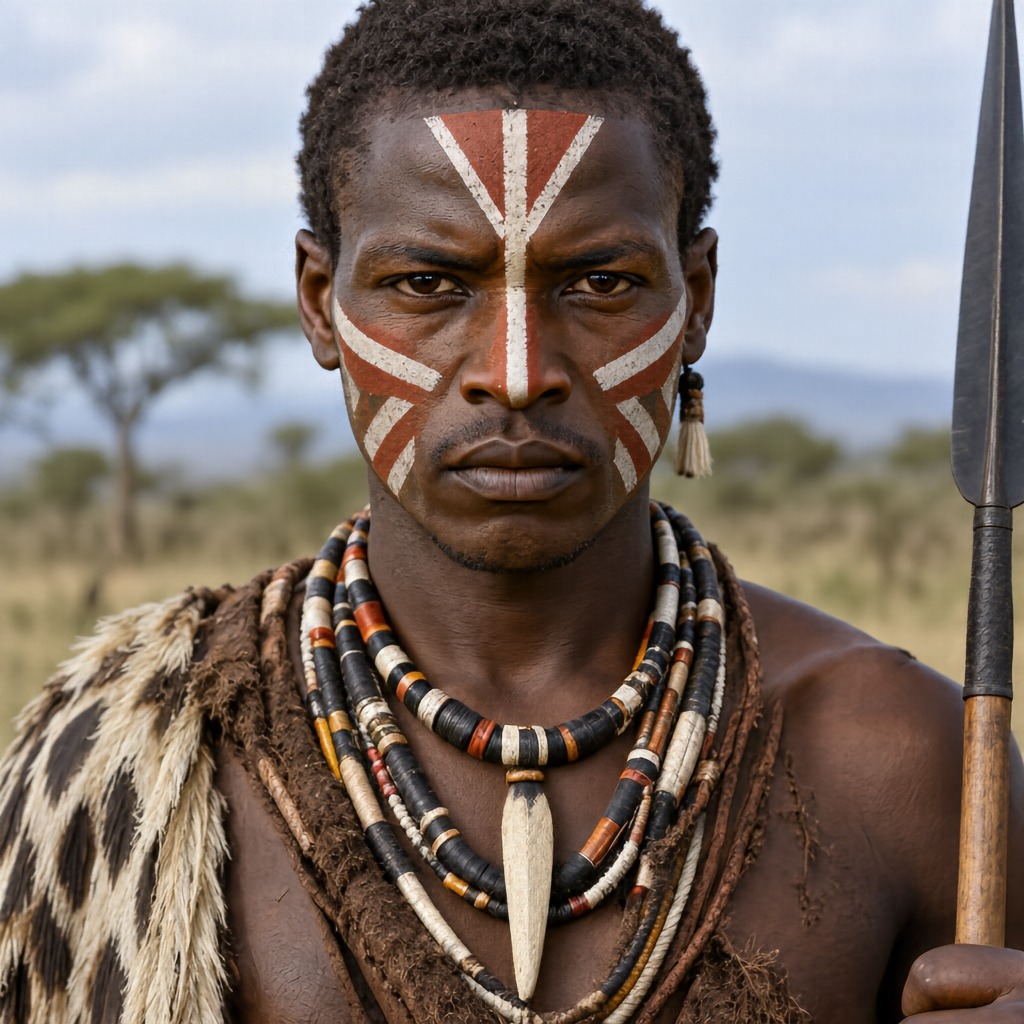}{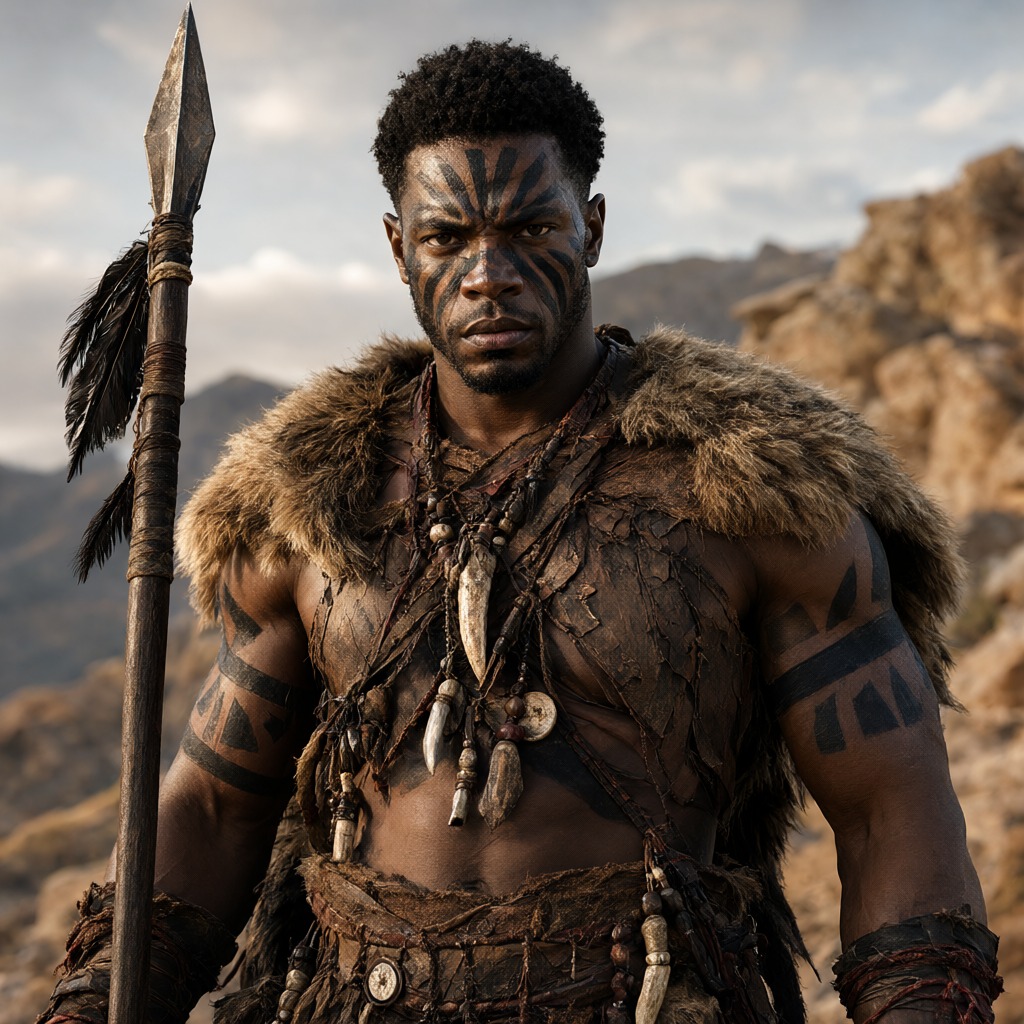}{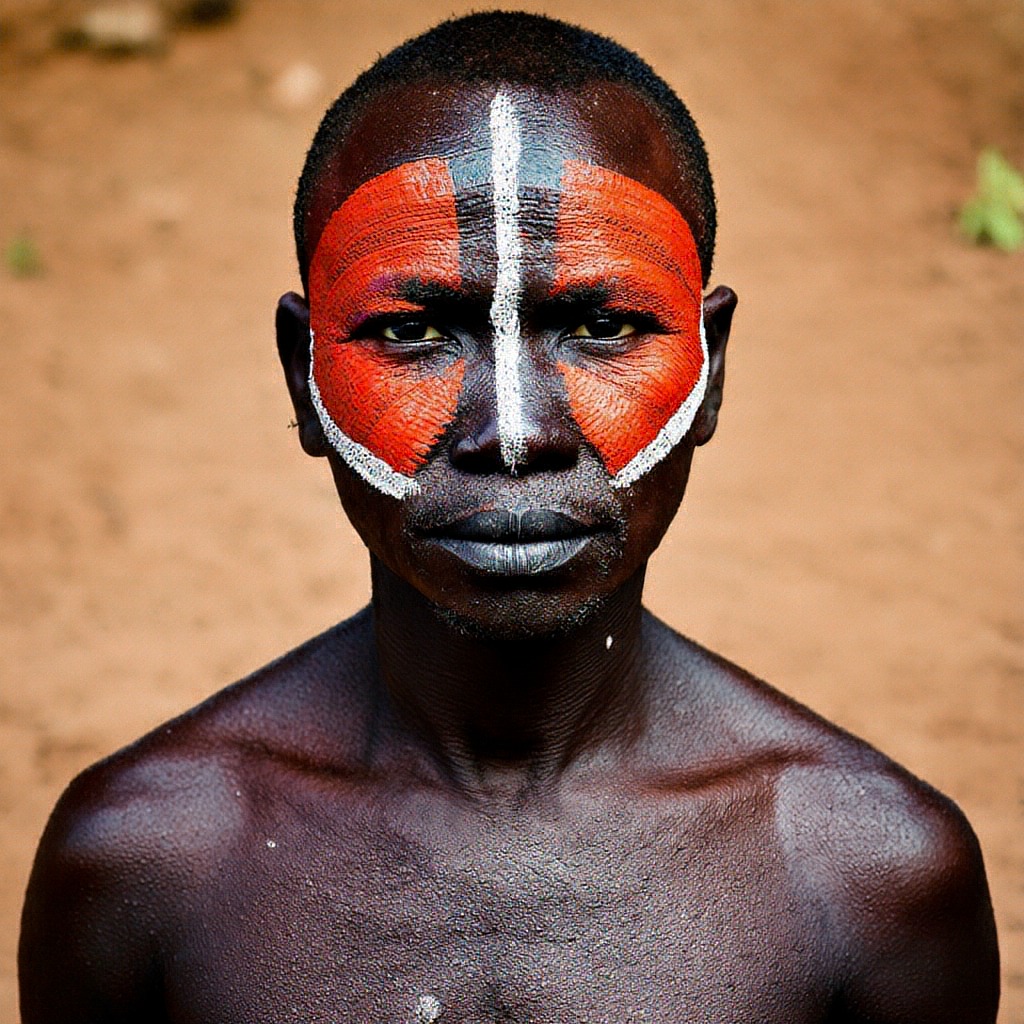}{An African tribal warrior standing facing the camera, face painted with symmetrical traditional patterns across the forehead, cheeks and nose bridge, paint edges crisp without bleeding. Underlying skin tone visible through the pigment, rough yet realistic skin texture, strong and determined gaze} &
\cmprow{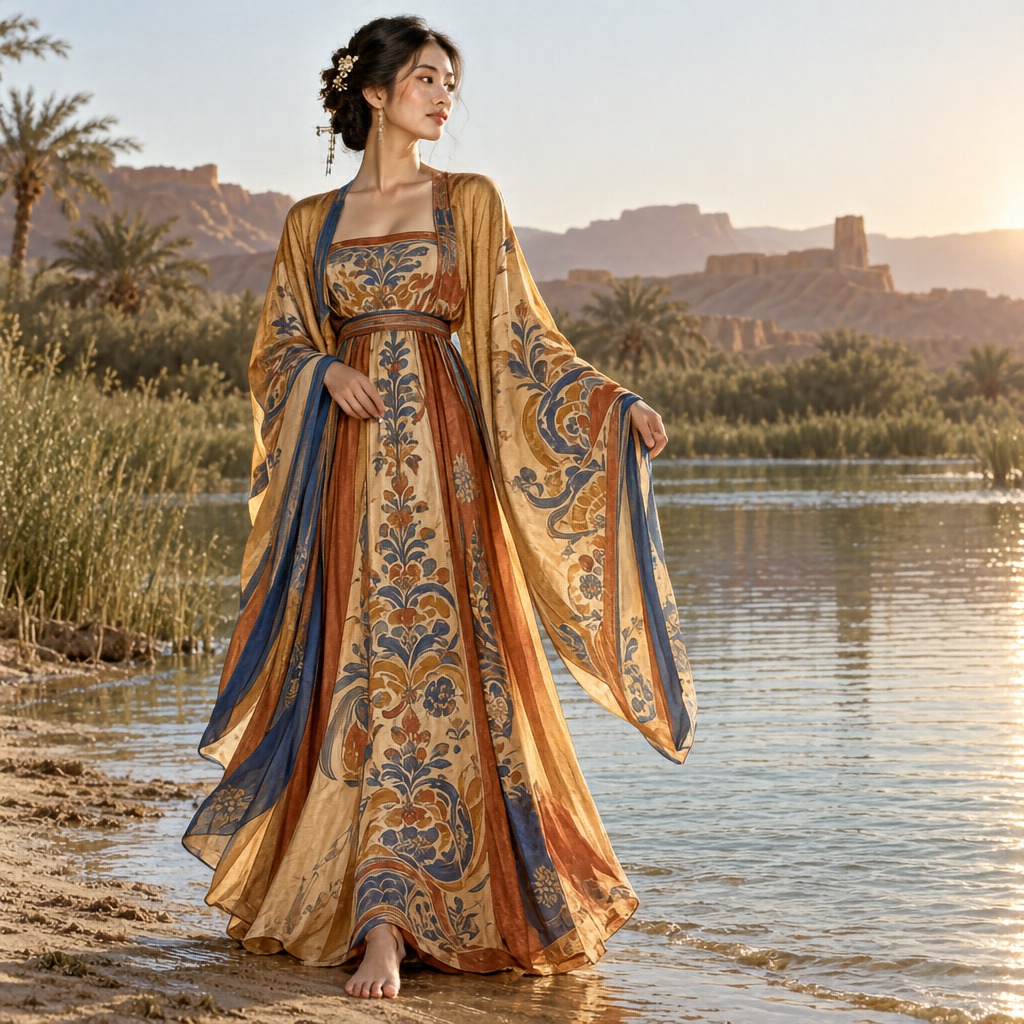}{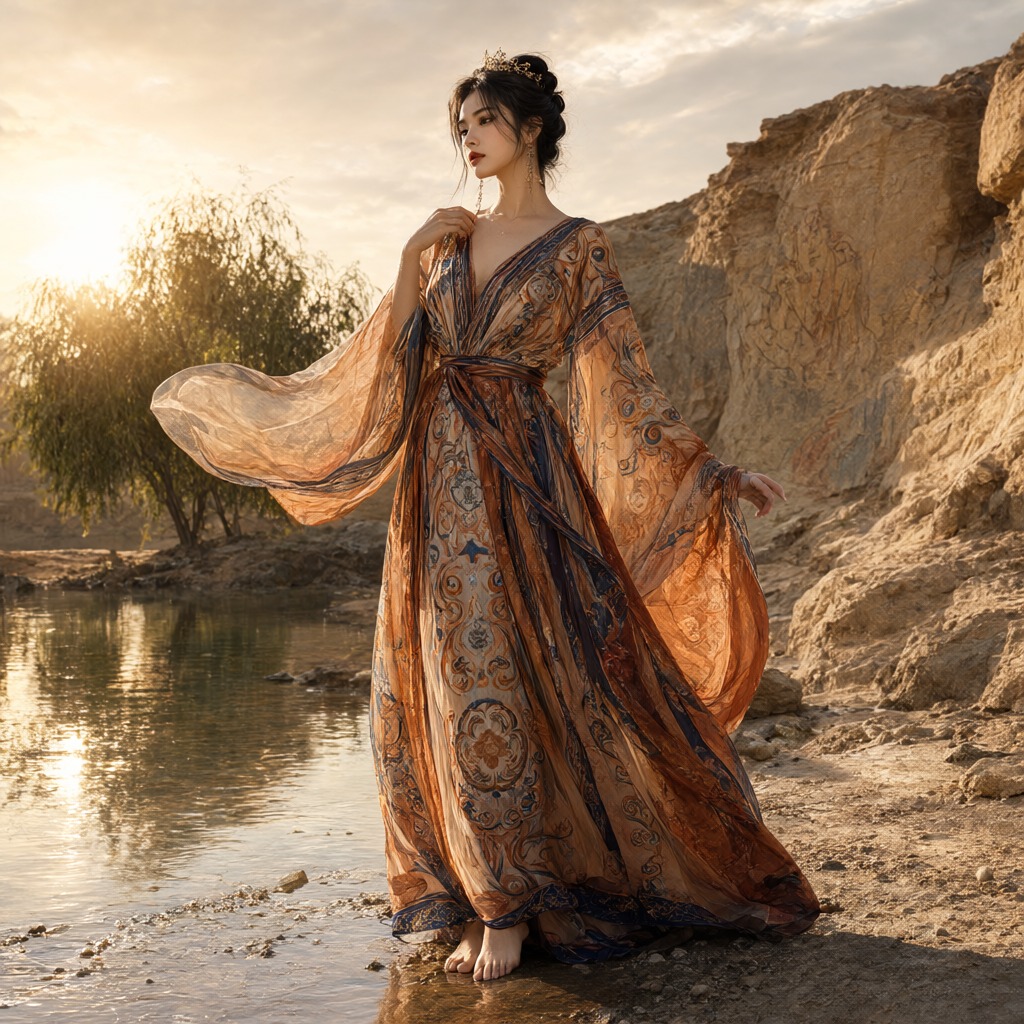}{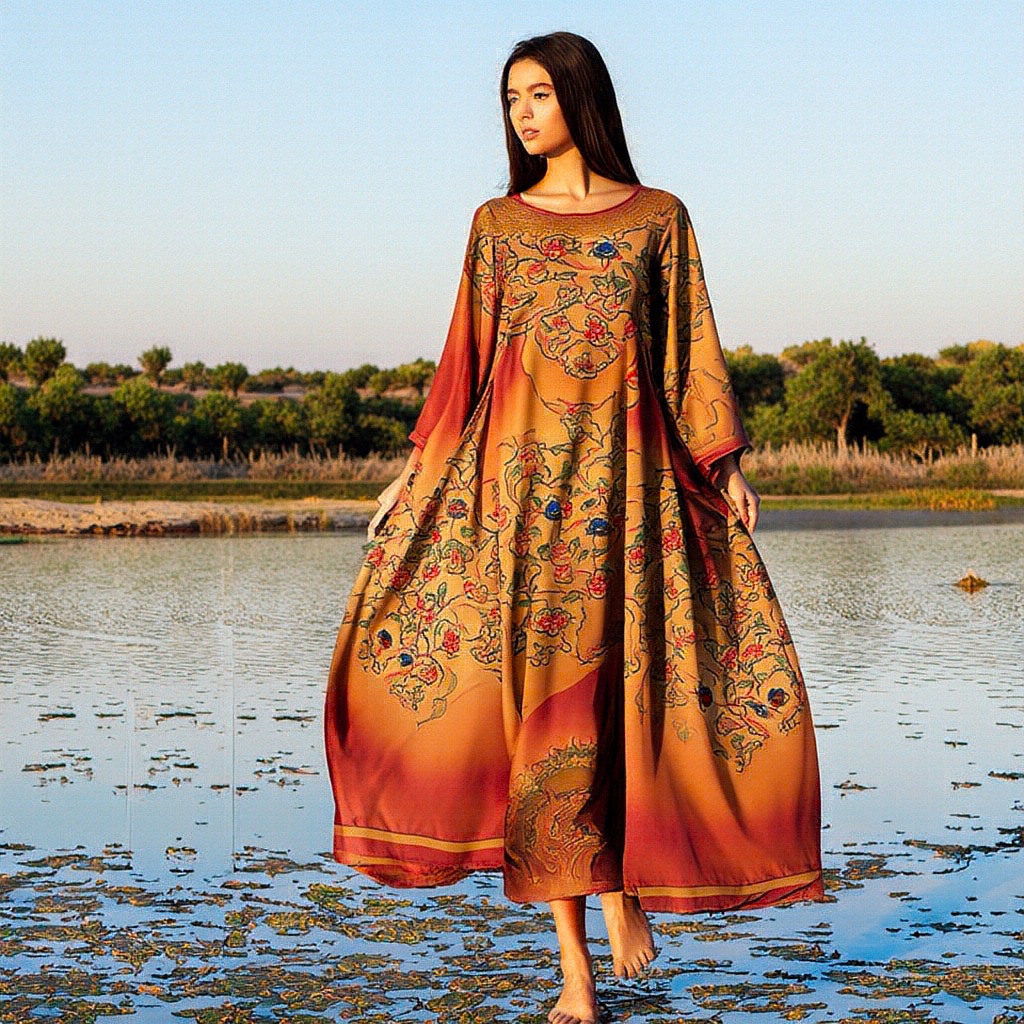}{Design a resort silk gown in Dunhuang mural palette: colors derived from mineral pigments---ochre, earth red, and lapis lazuli blue. Wide sleeves drape fluidly. Print faithfully reproduces Mogao Cave caisson motifs: honeysuckle scrolls and pearl-roundels, symmetrically arranged with fine linework. Model barefoot at edge of oasis shallow water, soft low-angle morning light, gentle glow highlighting silk sheen against matte skin---evoking Eastern mystique and serene ease.} \\[5pt]
\cmprow{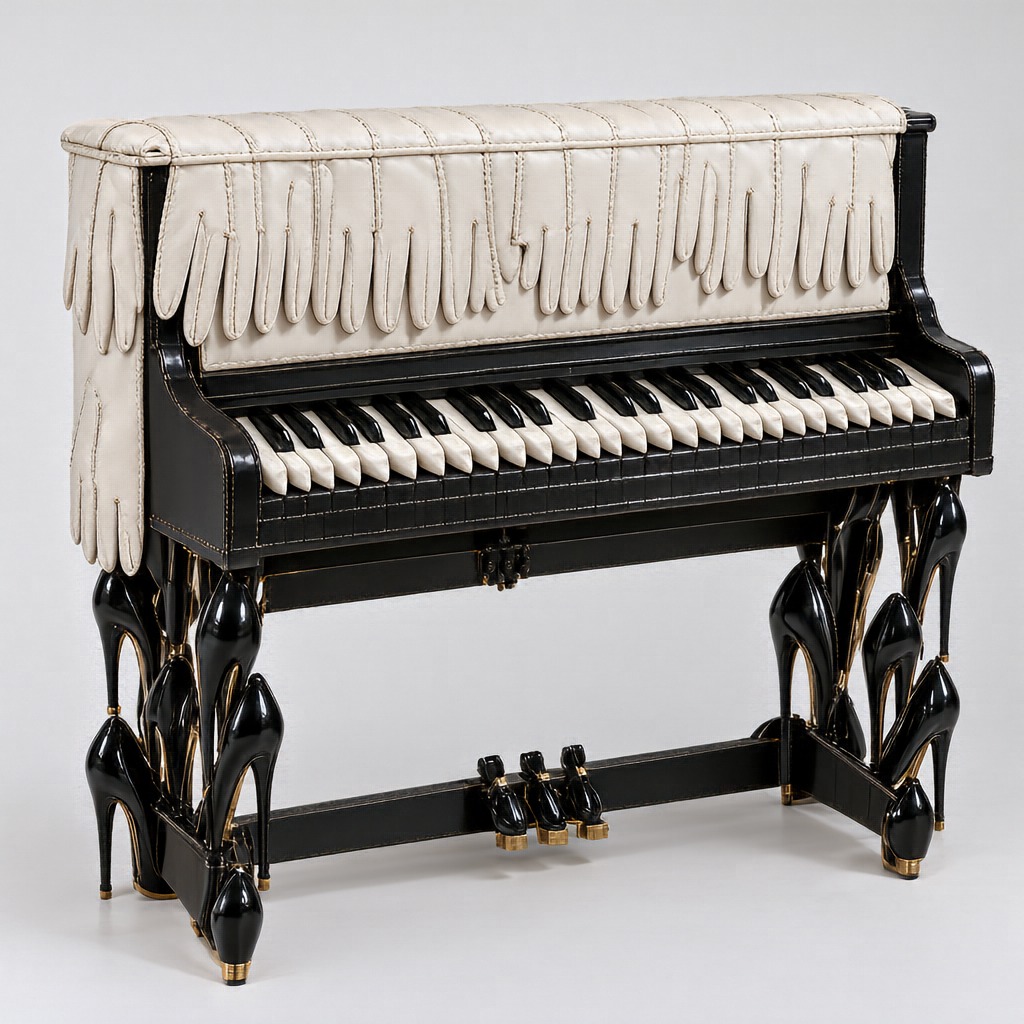}{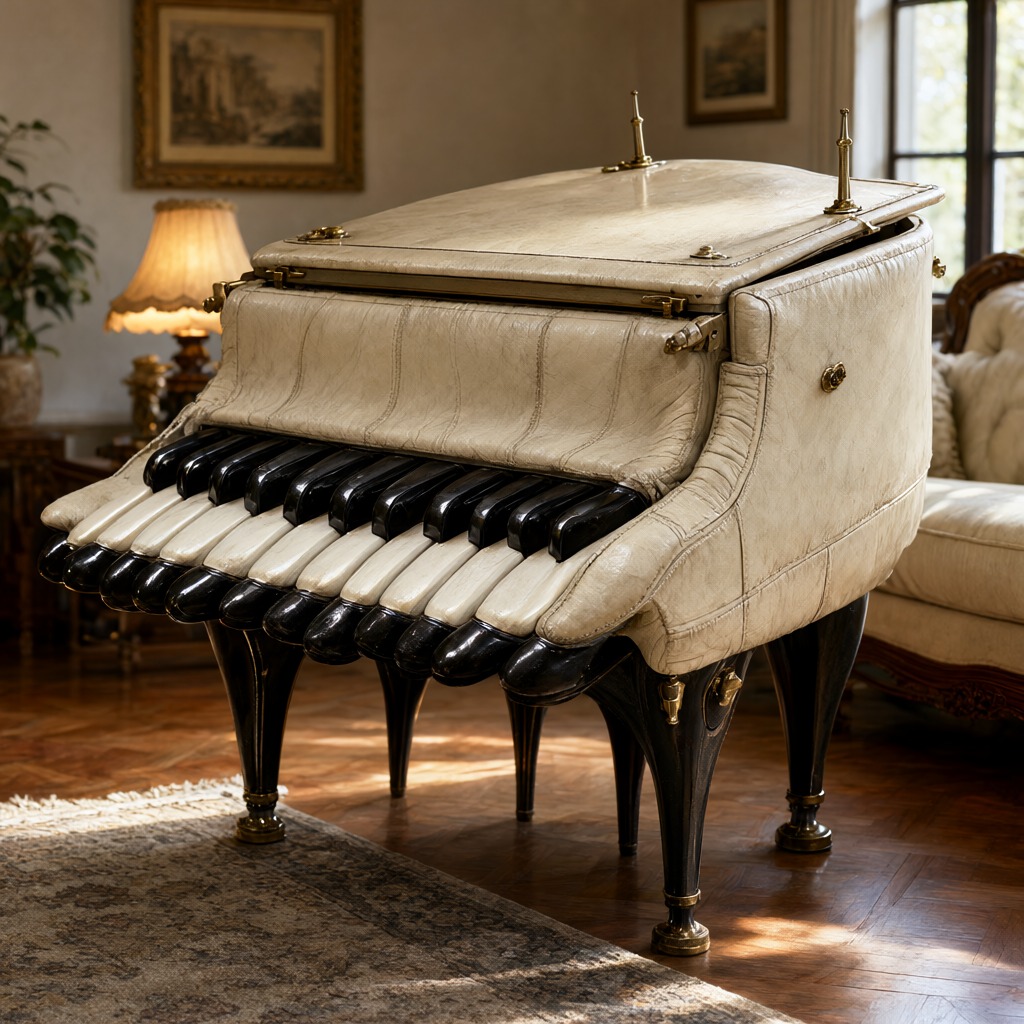}{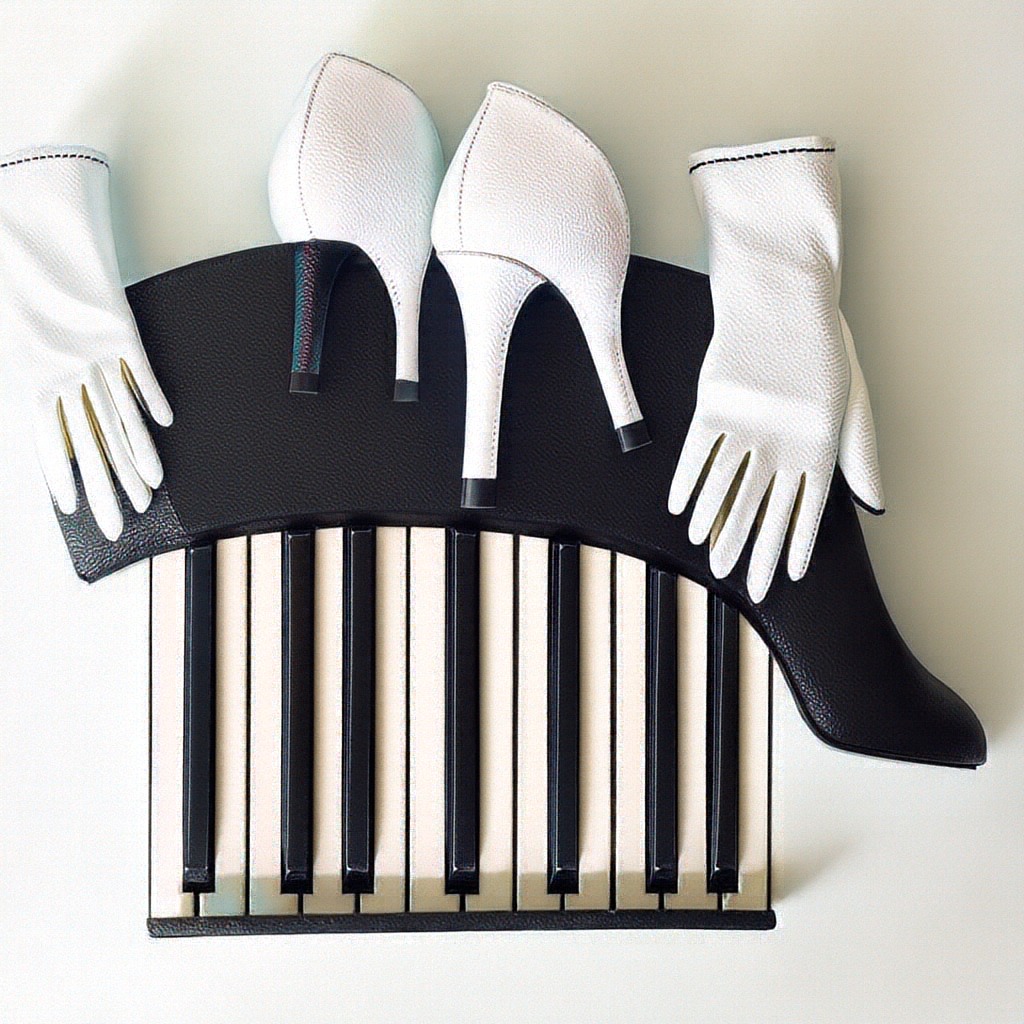}{The piano is cleverly composed of black high heels and white gloves. The toes of the shoes are closely arranged to form black and white keys, and the heels form the supporting structure. The white gloves are stitched to cover the piano body, and the leather texture and seams are clearly visible. The overall outline retains the elegant proportions of a traditional upright piano.} &
\cmprow{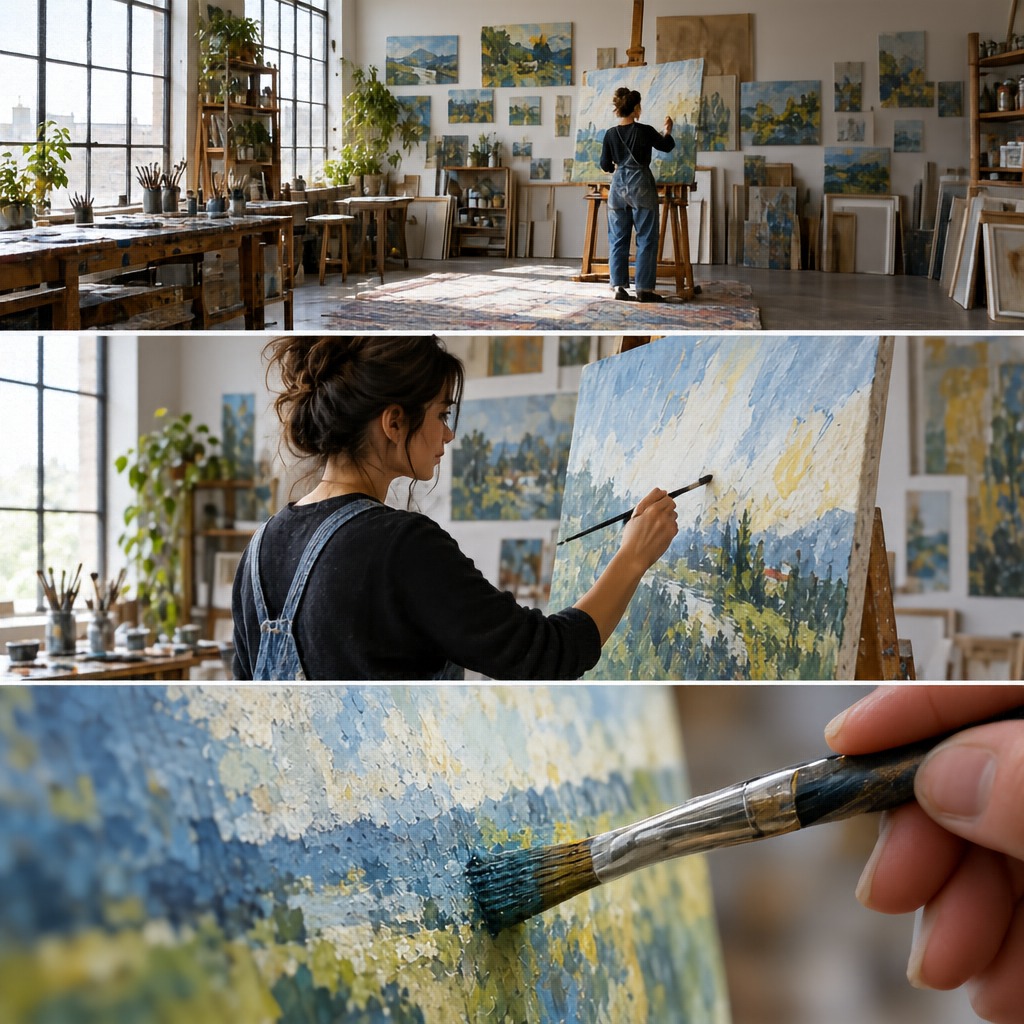}{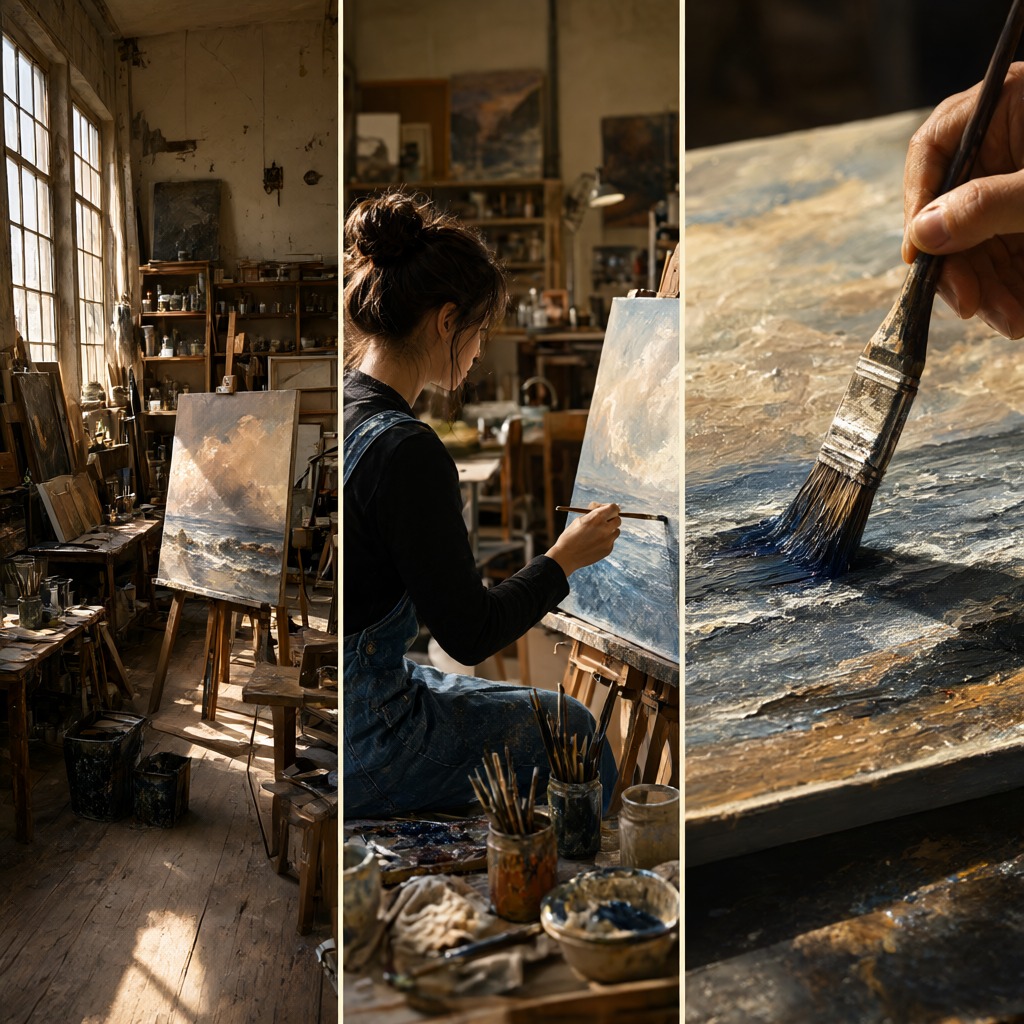}{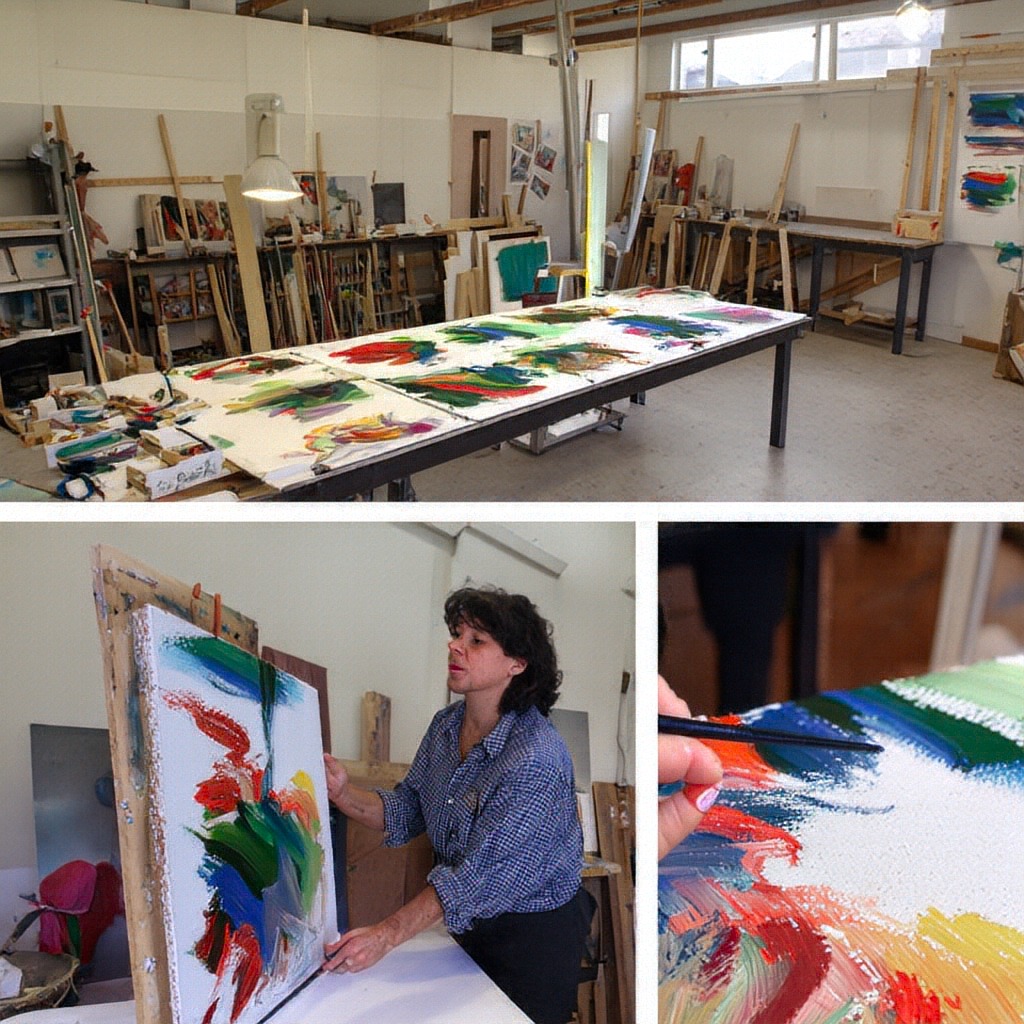}{Showing an artist's creative process, starting with a wide shot of the studio, followed by a medium shot of the artist working, ending with a close-up of brushstrokes touching the canvas.} \\[5pt]
\cmprow{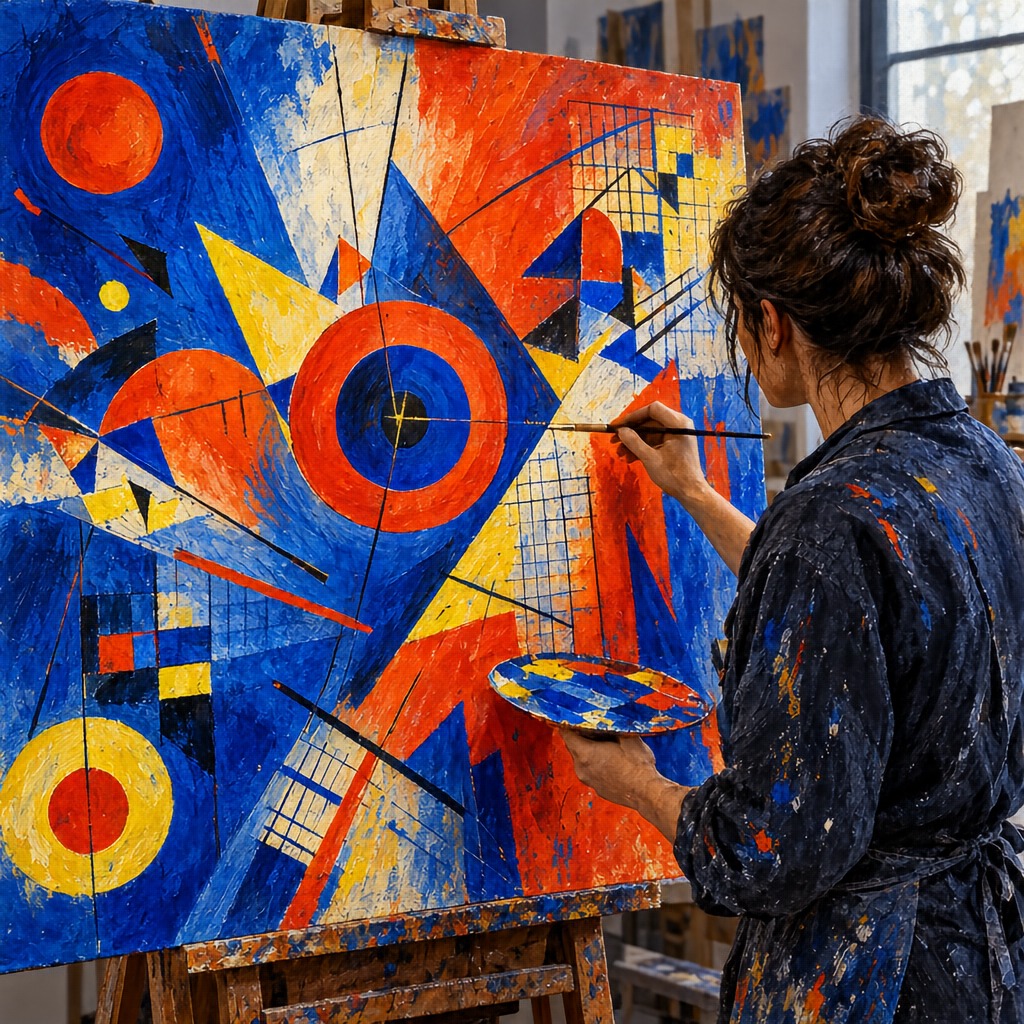}{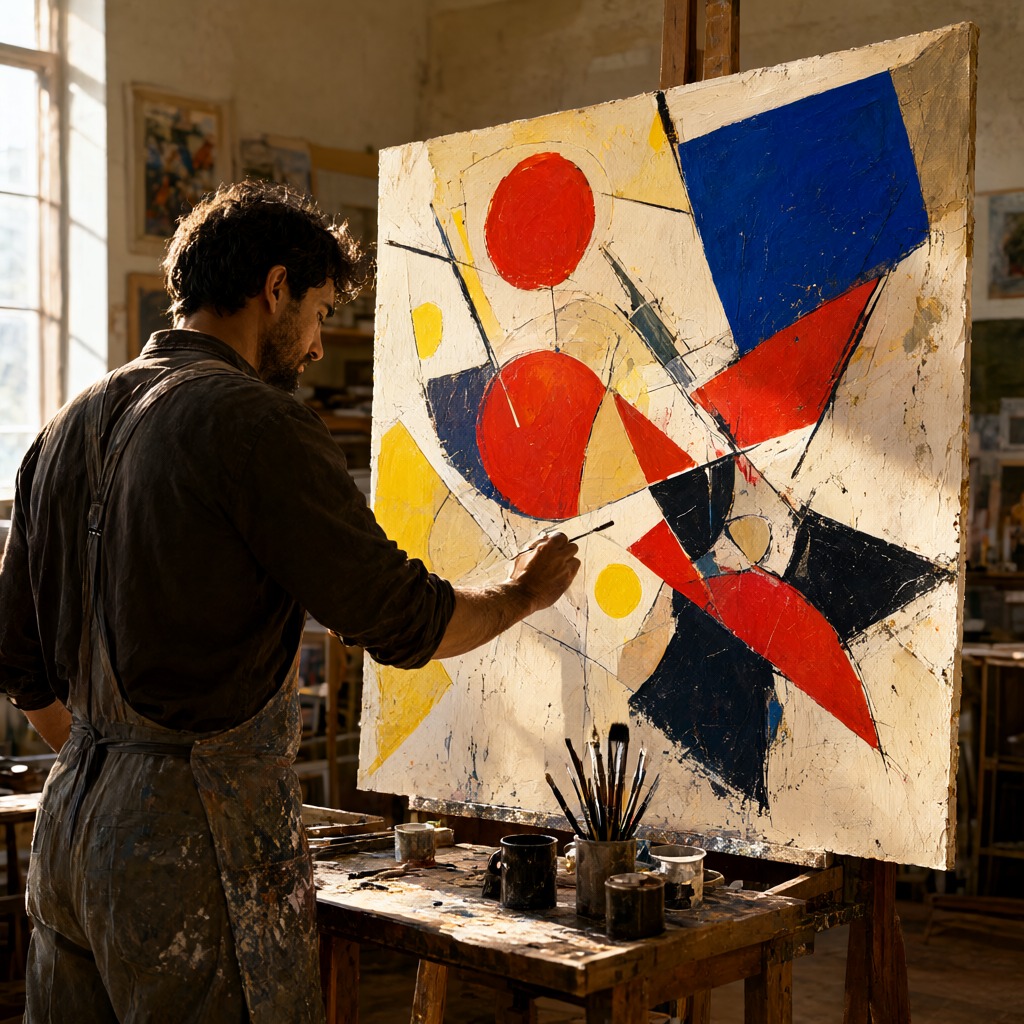}{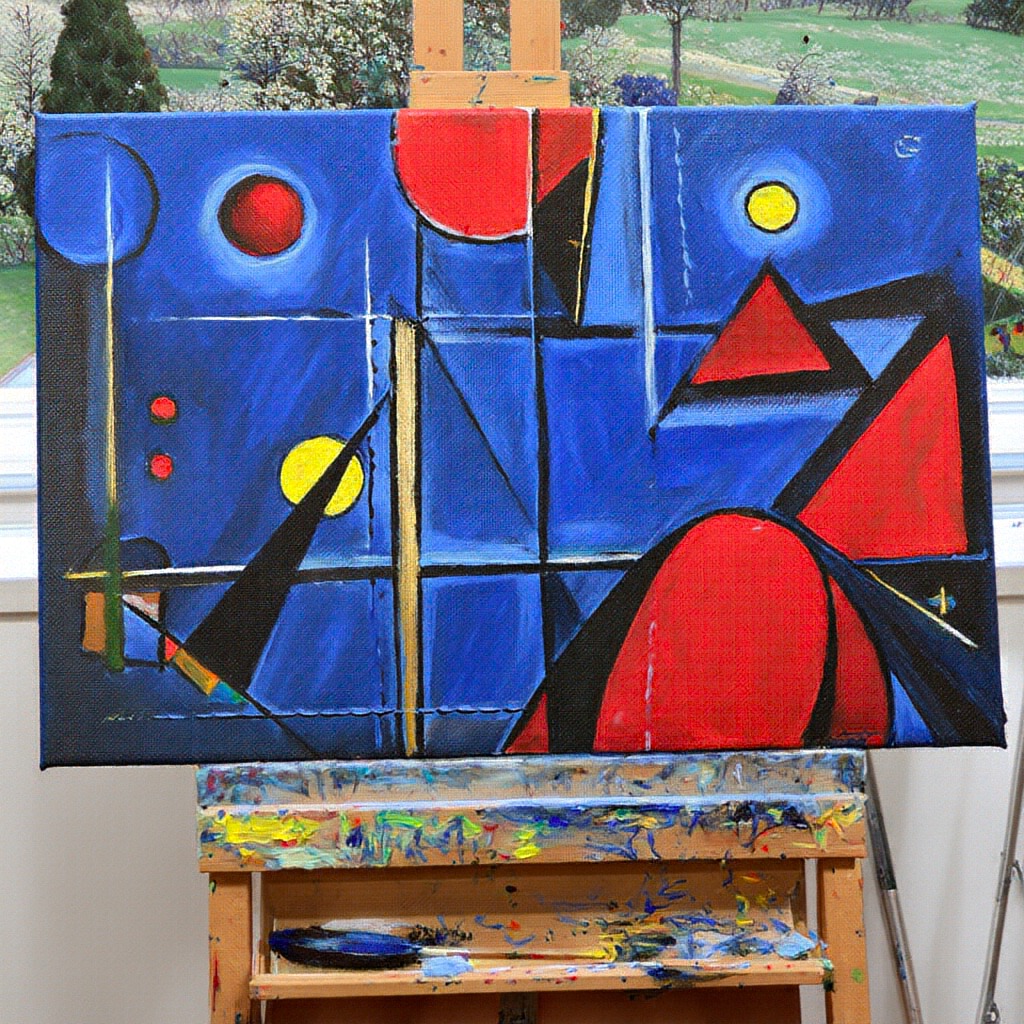}{Using the rule of thirds, an artist is creating an abstract masterpiece on a canvas, with the artwork occupying two-thirds of the image. The painting employs Kandinsky's concept of ``geometric shapes + color clashes'': incorporating circles, triangles, jagged lines, and fine grids; the main colors are cobalt blue and vermilion, supplemented by a small amount of lemon yellow; the center of the image is required to be close to the golden ratio point to form a visual anchor point, and the contrast between warm and cool colors presents a ``musical vibration'' feeling, with the overall brushstrokes and textures unified.} &
\cmprow{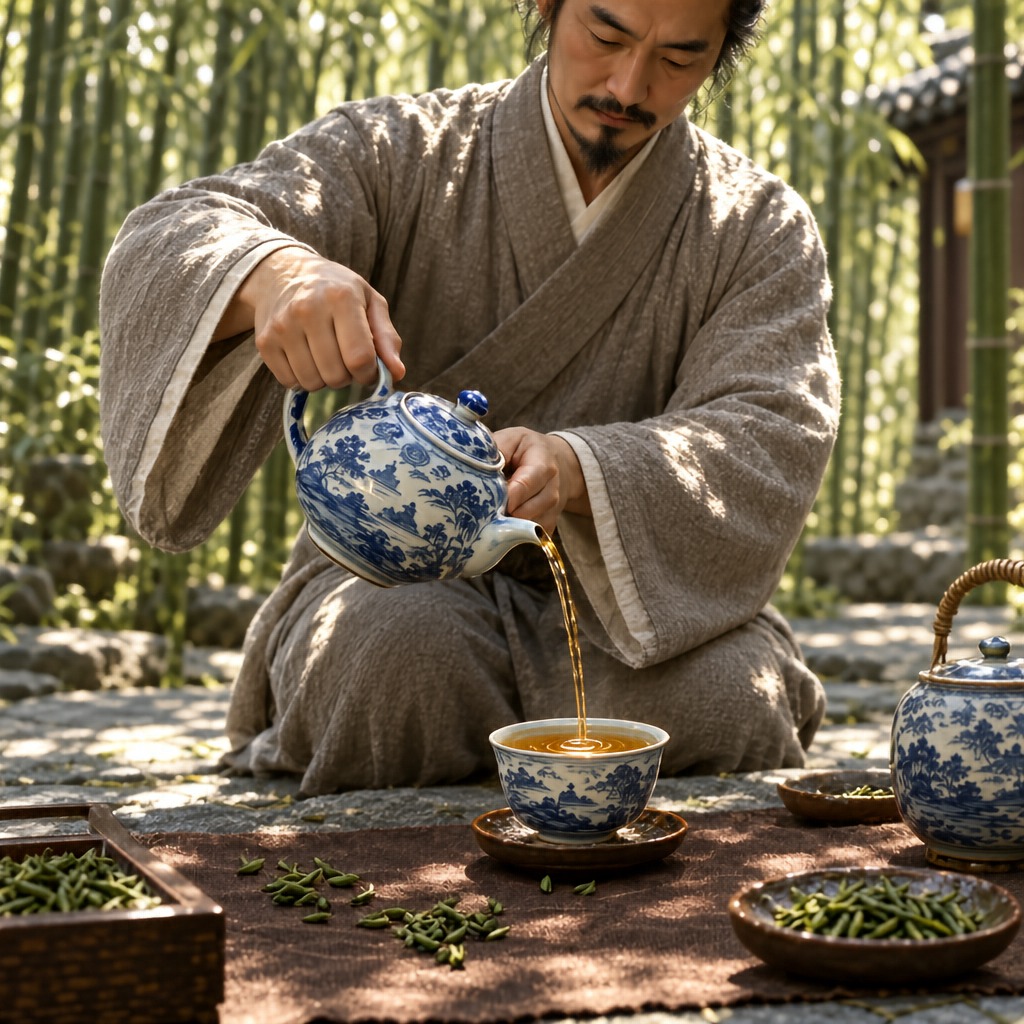}{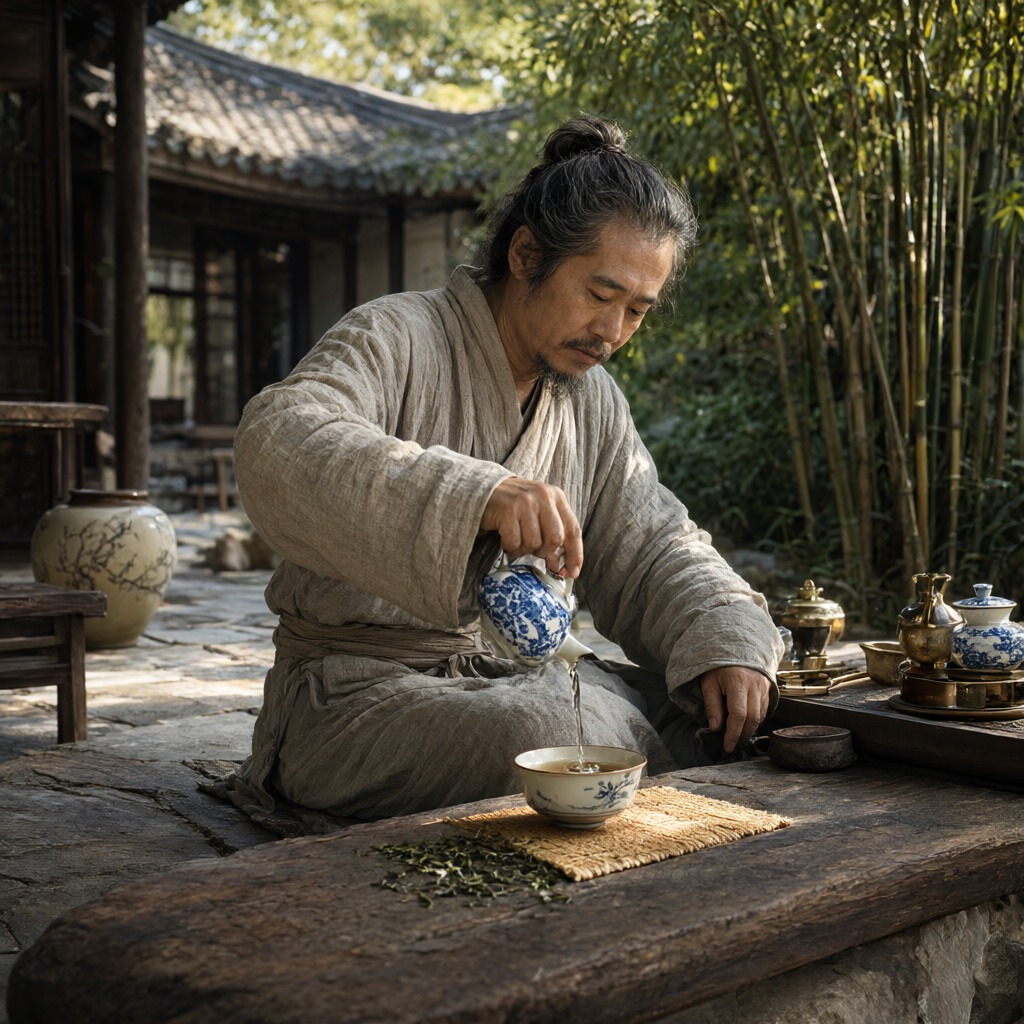}{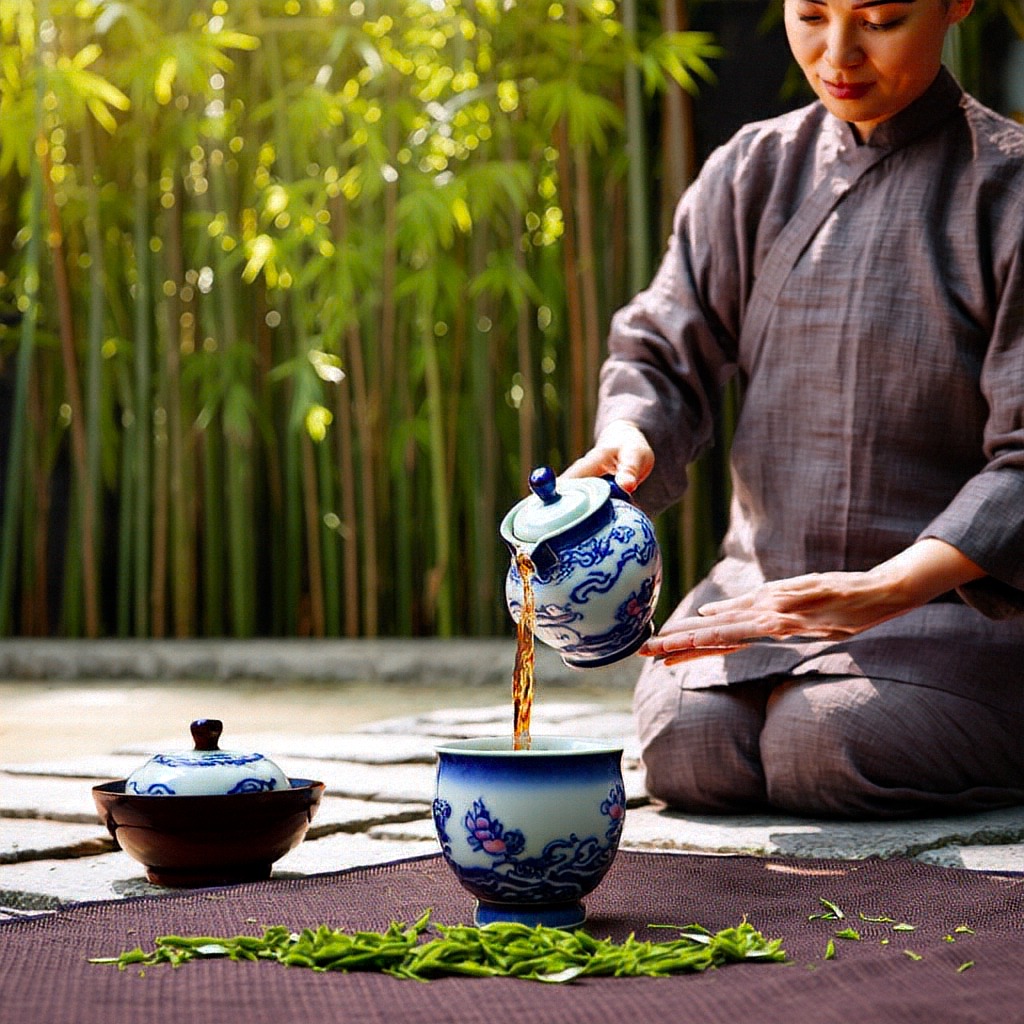}{A tea ceremony scene in a traditional Chinese courtyard: a tea master in a muted linen cross-collar robe kneels on bluestone slabs, both hands steadily pouring amber-hued tea from a cobalt-blue-and-white porcelain teapot into a matching cup; at the moment of pour, concentric ripples radiate clearly across the liquid surface, catching soft highlights; the porcelain---decorated with intricate scrolling lotus and landscape motifs---shows nuanced cobalt shading and a lustrous, warm glaze; behind, a dense bamboo grove filters sunlight into dappled specks that dance across the tea mat, vessels, and sleeves; in the foreground, a dark brown linen tea mat is scattered with fresh green tea buds; rendered in shallow-depth photographic style, focus locked on the flowing stream and cup rim, with the bamboo background softly blurred yet texturally rich; dominated by celadon, white, umber, and golden tones, evoking tranquility, presence, and the contemplative spirit of Chinese tea culture} \\[5pt]
\cmprow{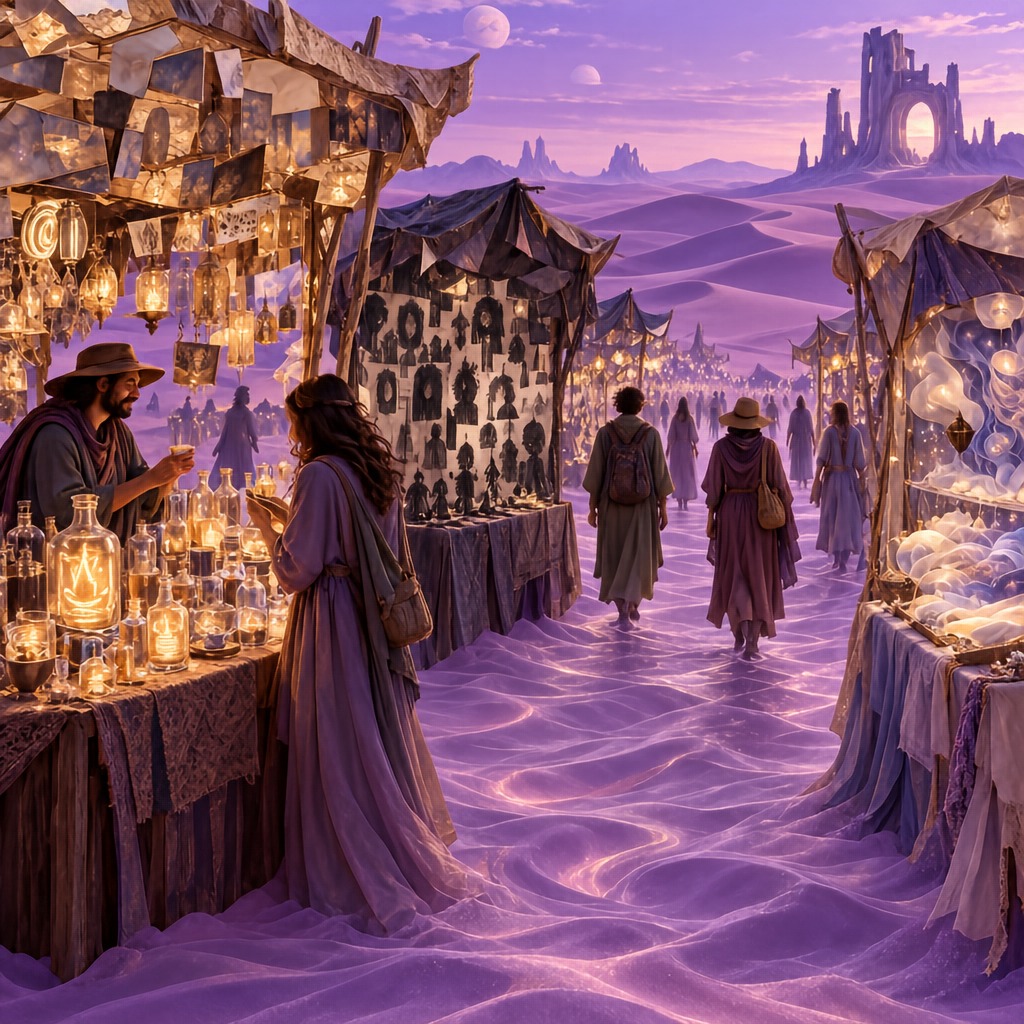}{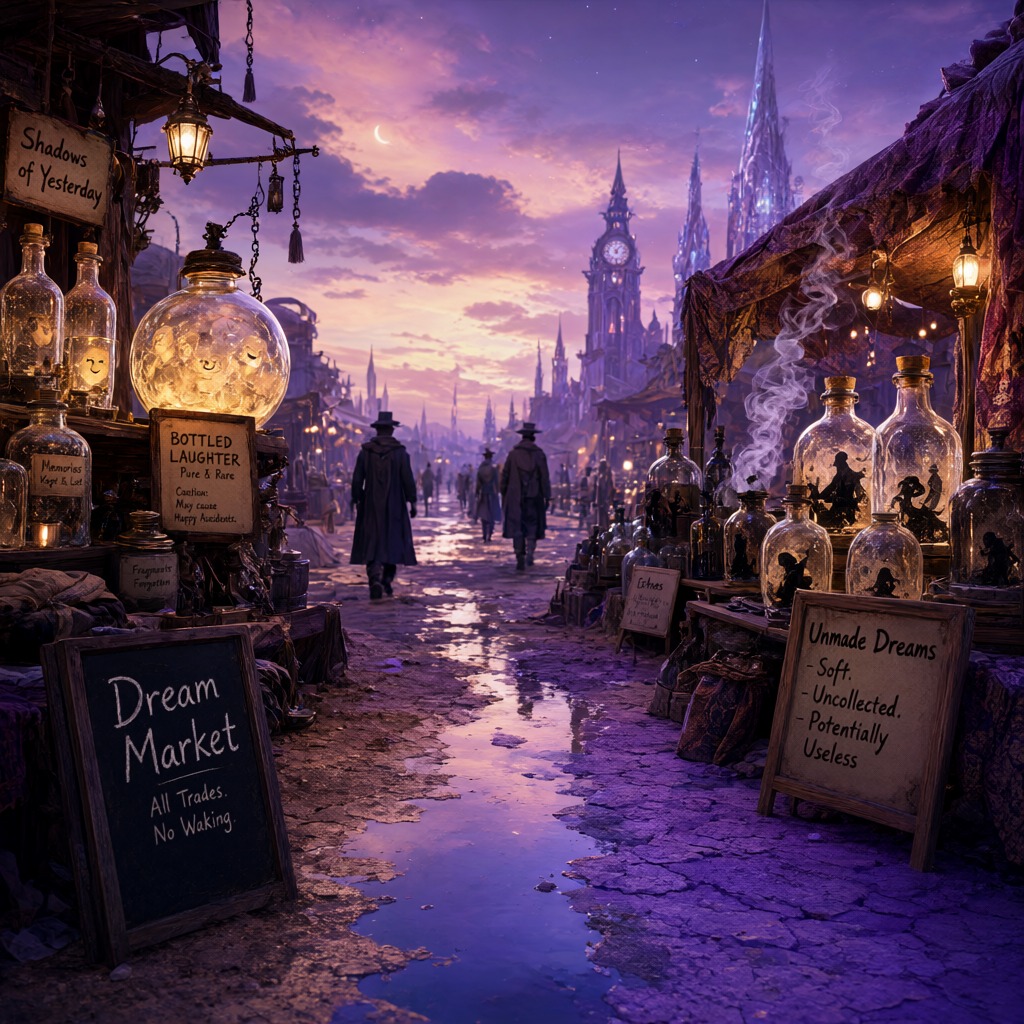}{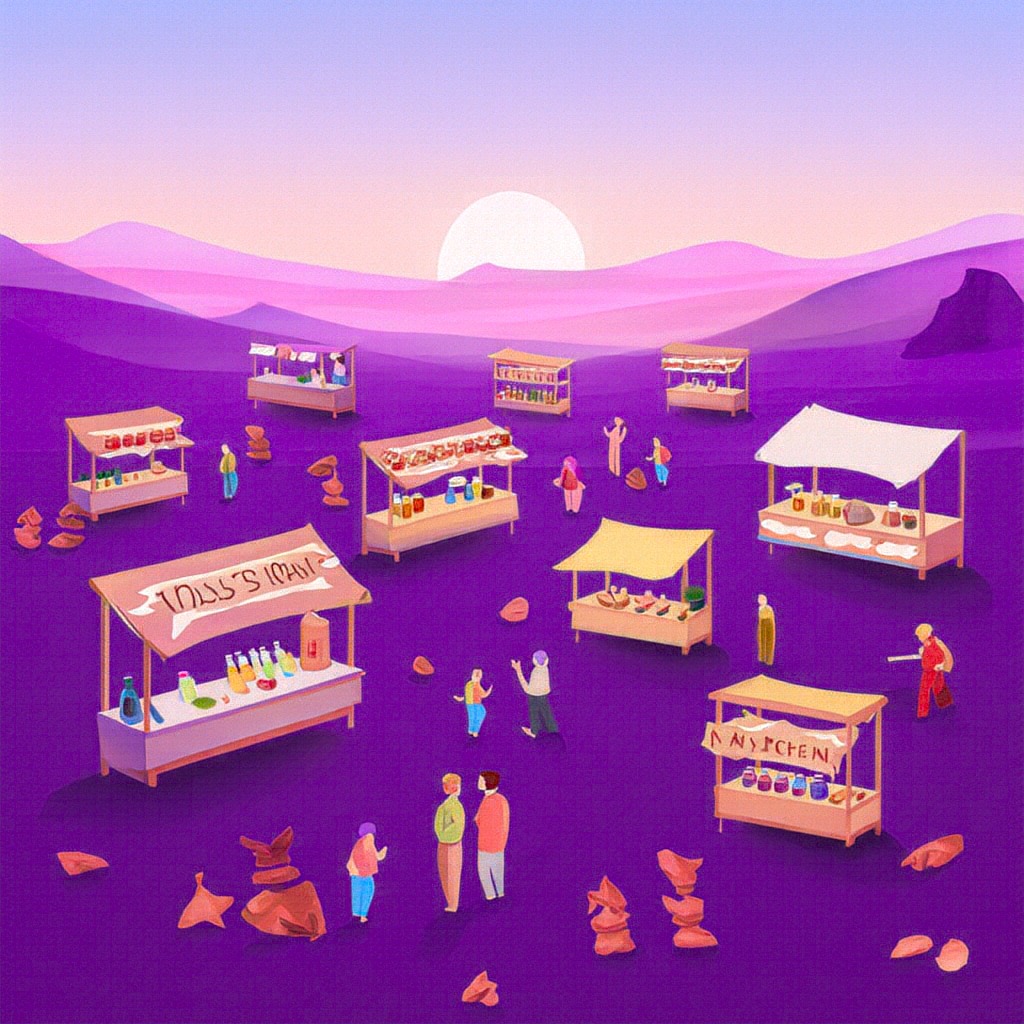}{A dream market unfolds in a violet desert; stalls are constructed from fragments of memory, selling bottled laughter, shadows of yesterday, and unmade dreams; customers walk on quicksand yet never sink.} &
\cmprow{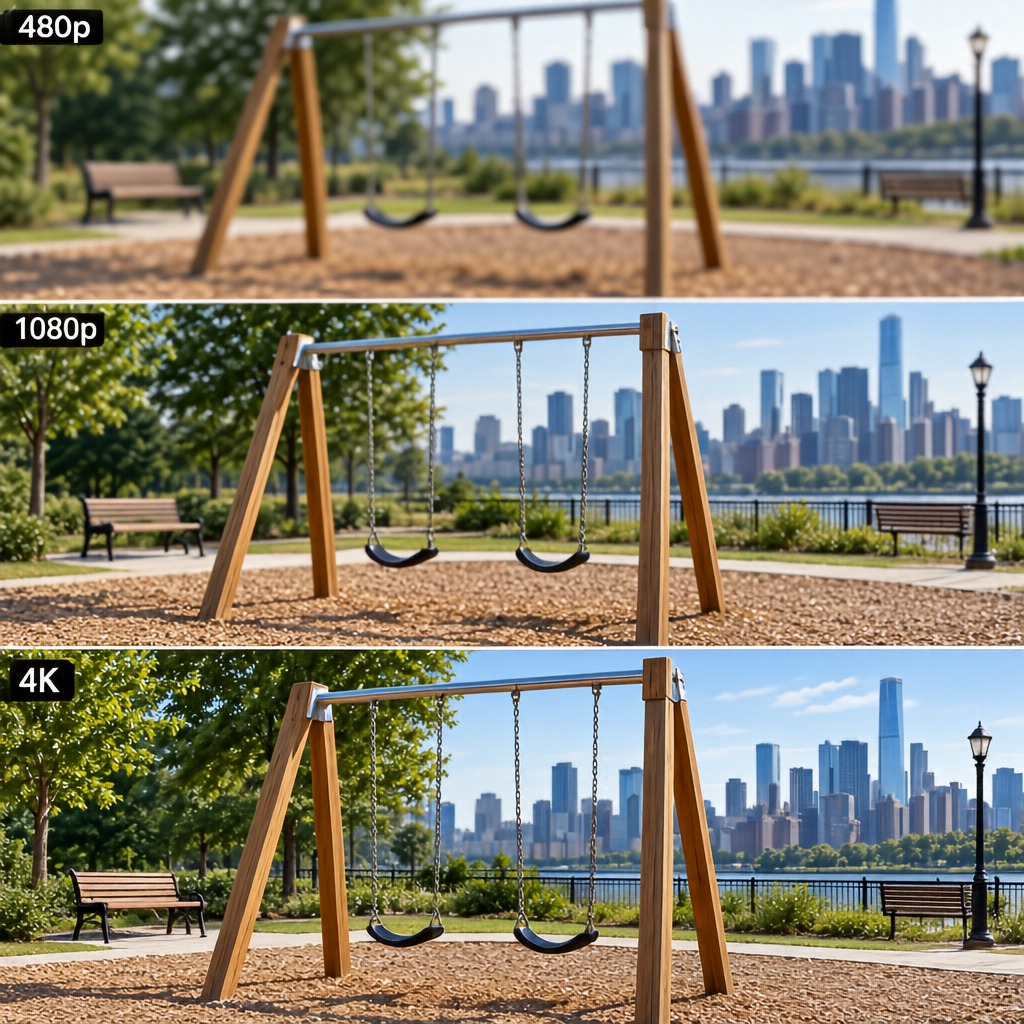}{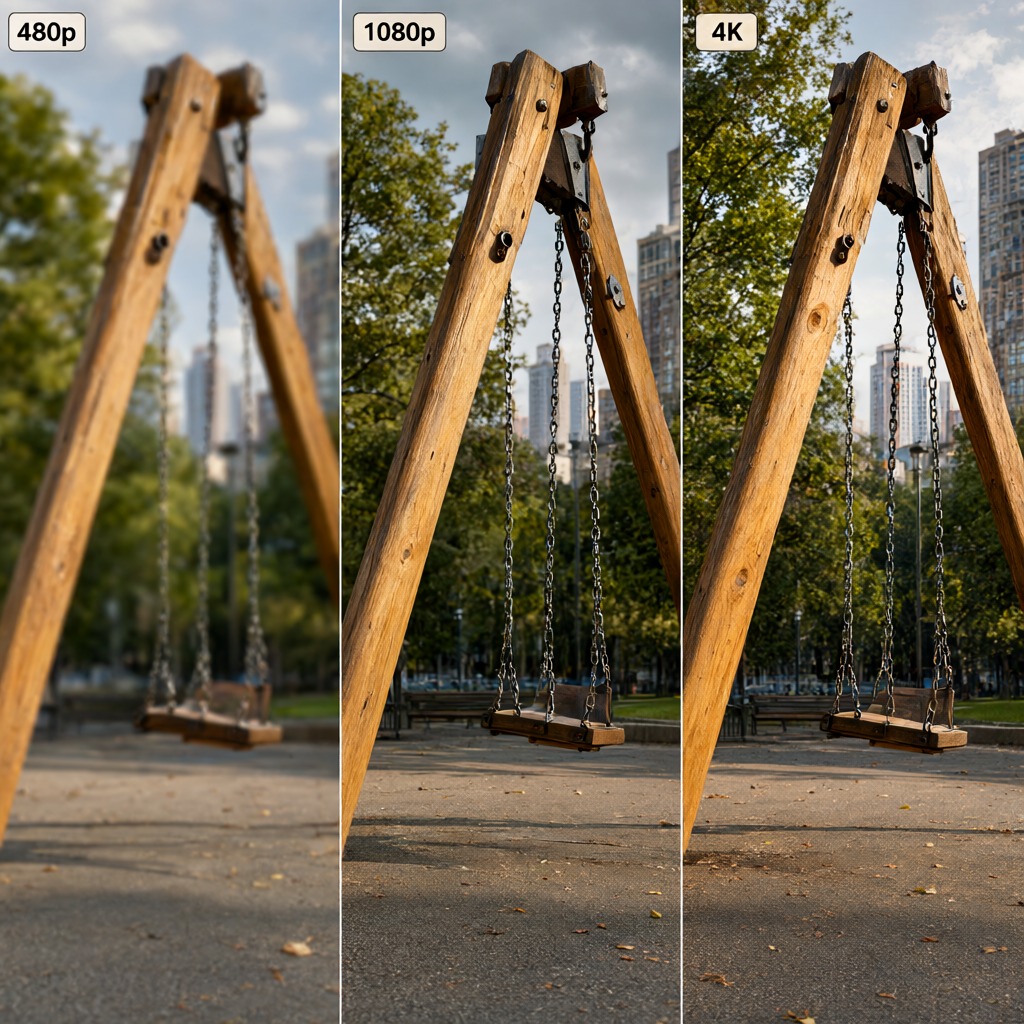}{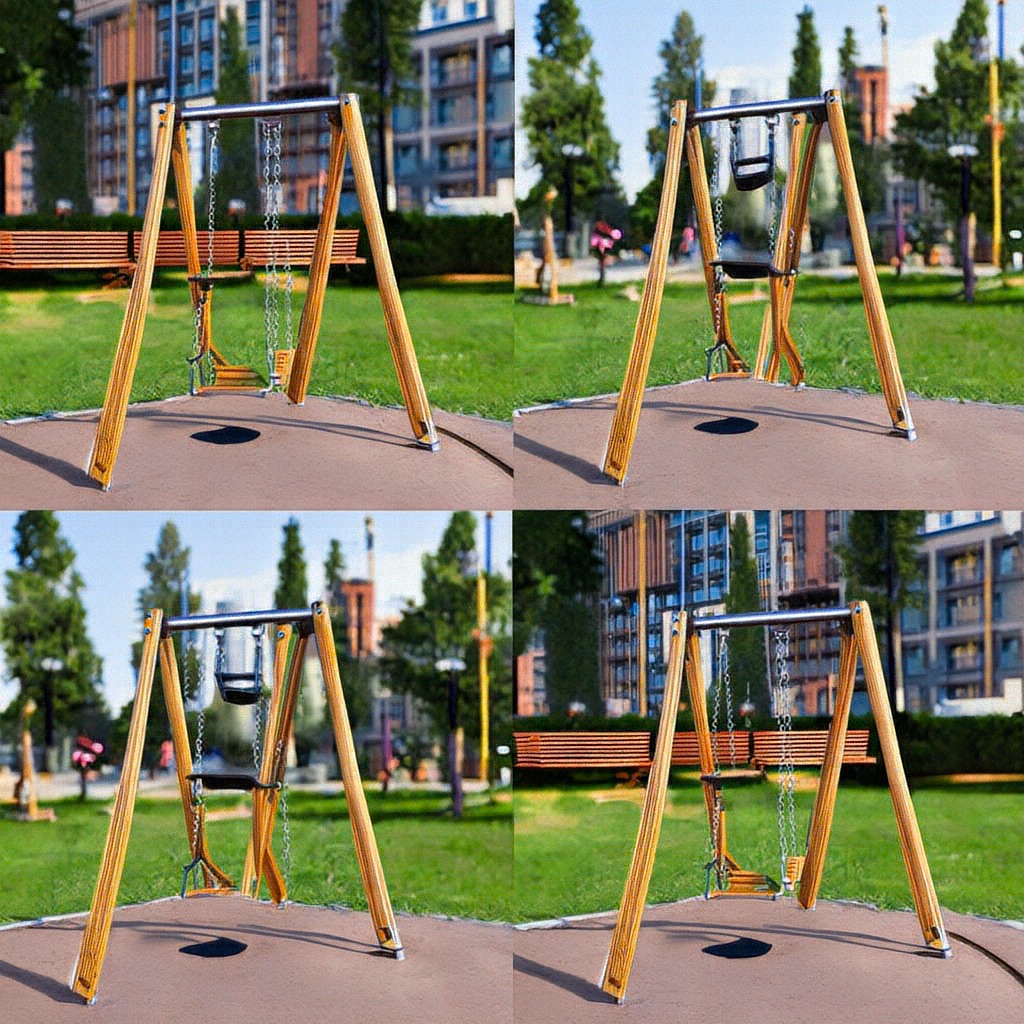}{A swing set in an urban park with clear wooden textures, natural metallic shine, and appropriate background blur, rendered in 480p, 1080p, and 4K resolutions.} \\
\end{tabular}}
\end{minipage}}
  \caption{\textbf{Visual comparison on text-to-image generation.} Each group of three images shows one prompt from QBench, printed below it, at $1024{\times}1024$. Qwen-Image-2.1 and AsymFLUX.2 klein use the settings of their QBench evaluation in Table~\ref{tab:t2i-headline}; $^\dagger$with the official prompt rewriter. Repr-MoE uses guidance scale 4.}
  \label{fig:t2i-compare}
\end{figure}

\begin{figure}[p]
  \centering
  \includegraphics[width=\textwidth]{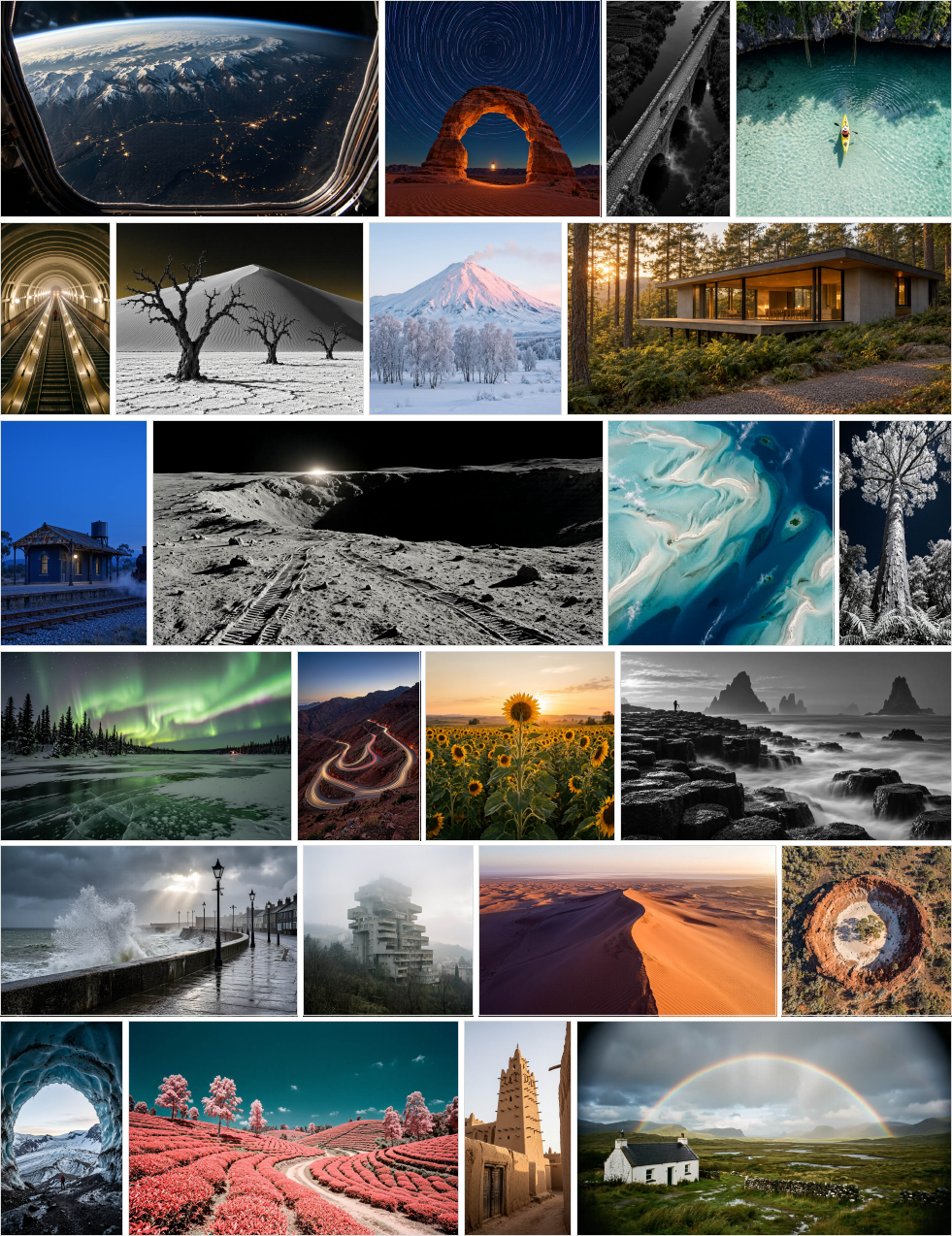}
  \caption{\textbf{Visual results of Repr-MoE on text-to-image generation: photographs of landscapes and architecture.}}
  \label{fig:t2i-gallery-places}
\end{figure}

\begin{figure}[p]
  \centering
  \includegraphics[width=\textwidth]{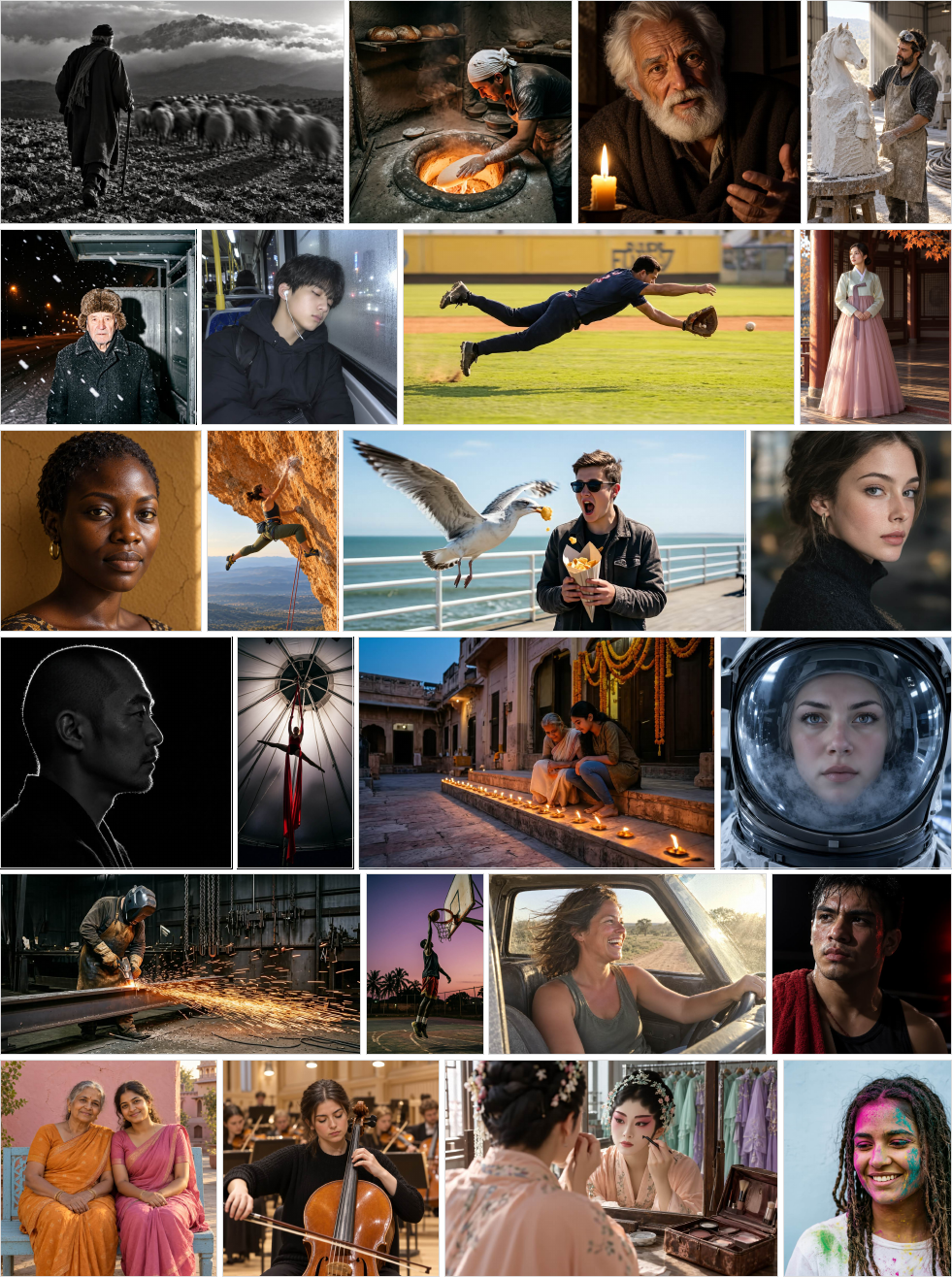}
  \caption{\textbf{Visual results of Repr-MoE on text-to-image generation: photographs of people.}}
  \label{fig:t2i-gallery-people}
\end{figure}

\begin{figure}[p]
  \centering
  \includegraphics[width=\textwidth]{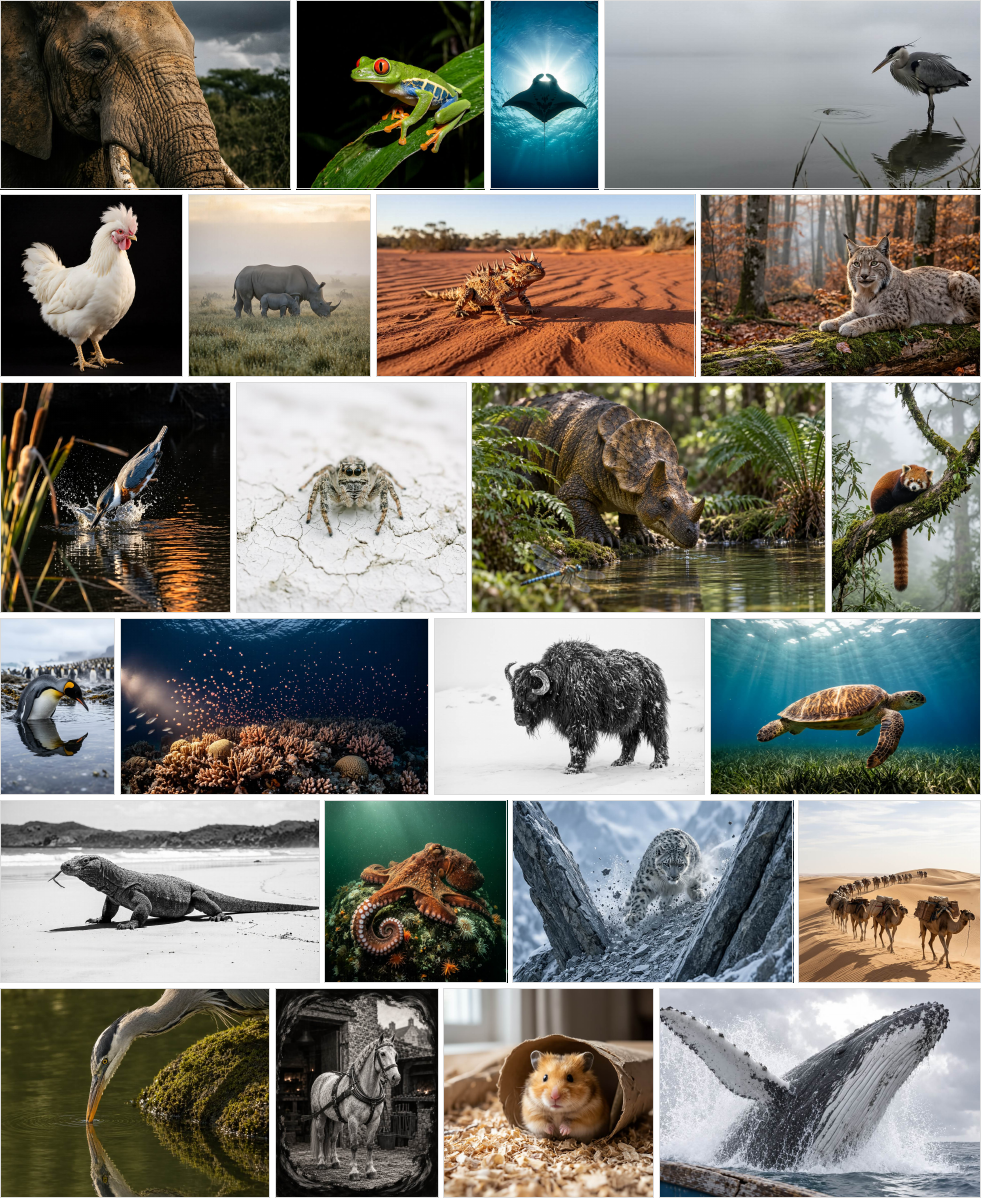}
  \caption{\textbf{Visual results of Repr-MoE on text-to-image generation: photographs of animals.}}
  \label{fig:t2i-gallery-animals}
\end{figure}

\begin{figure}[p]
  \centering
  \includegraphics[width=\textwidth]{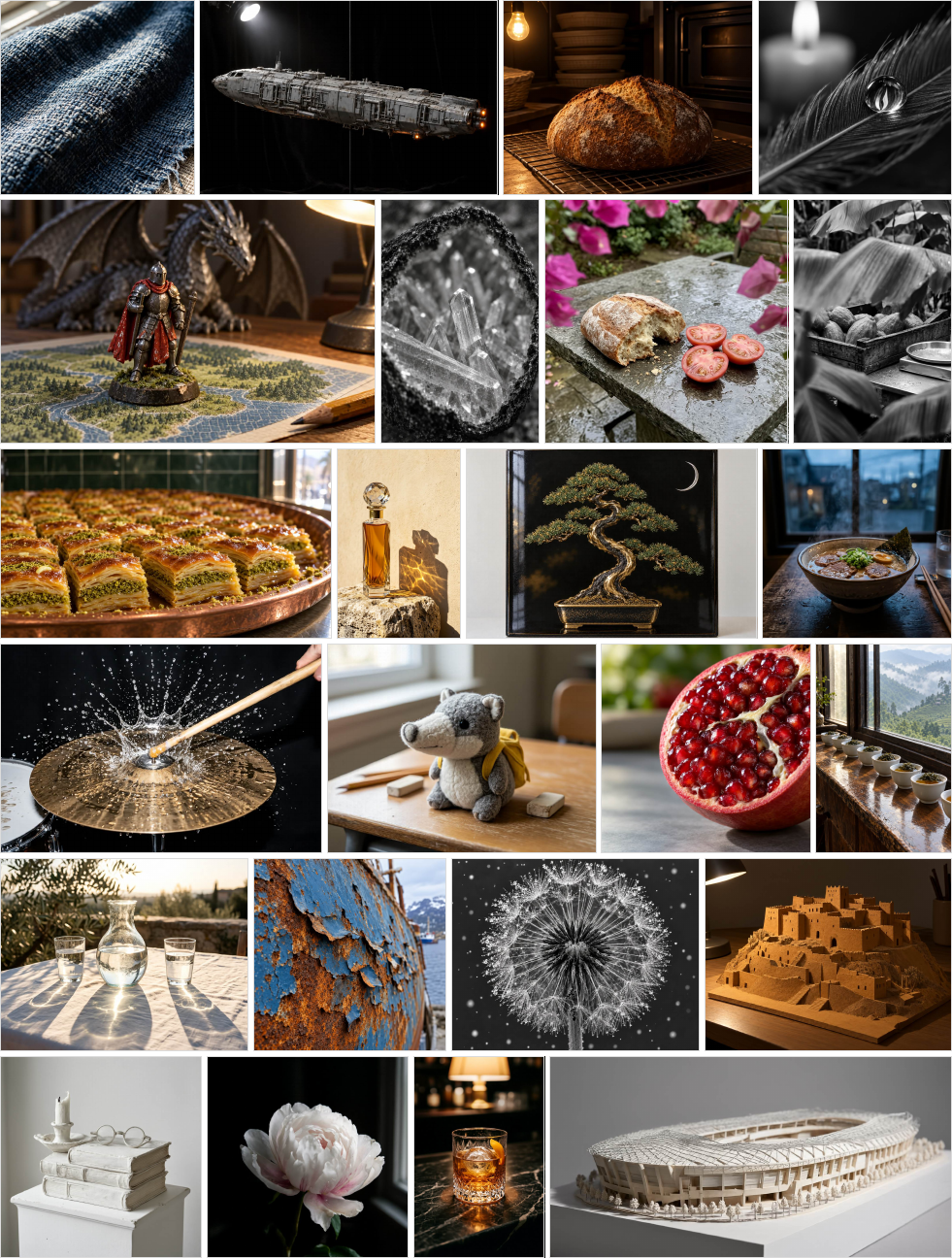}
  \caption{\textbf{Visual results of Repr-MoE on text-to-image generation: still life and close-ups.}}
  \label{fig:t2i-gallery-still}
\end{figure}

\begin{figure}[p]
  \centering
  \includegraphics[width=\textwidth]{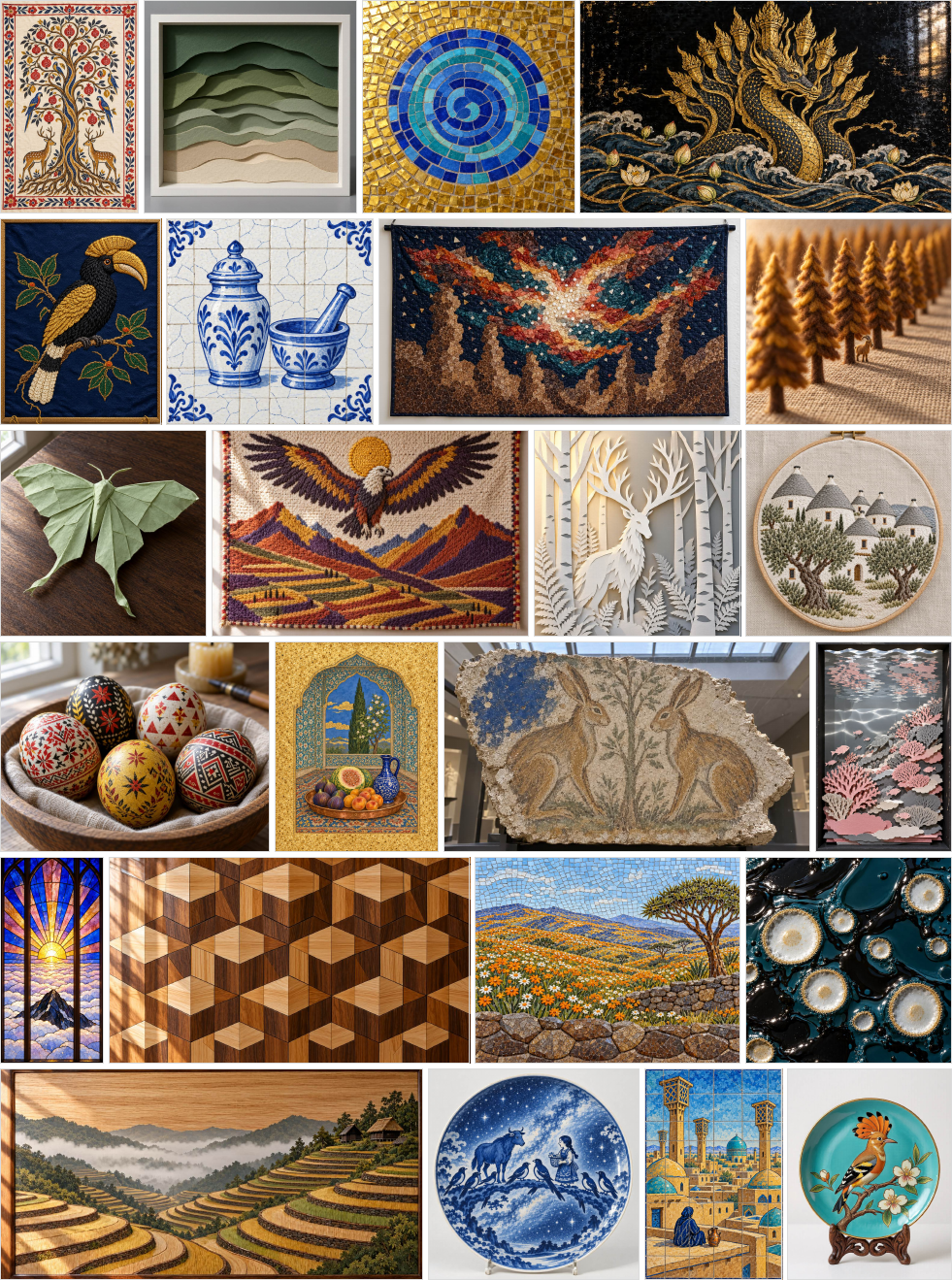}
  \caption{\textbf{Visual results of Repr-MoE on text-to-image generation: craft and handmade work.}}
  \label{fig:t2i-gallery-craft}
\end{figure}

\begin{figure}[p]
  \centering
  \includegraphics[width=\textwidth]{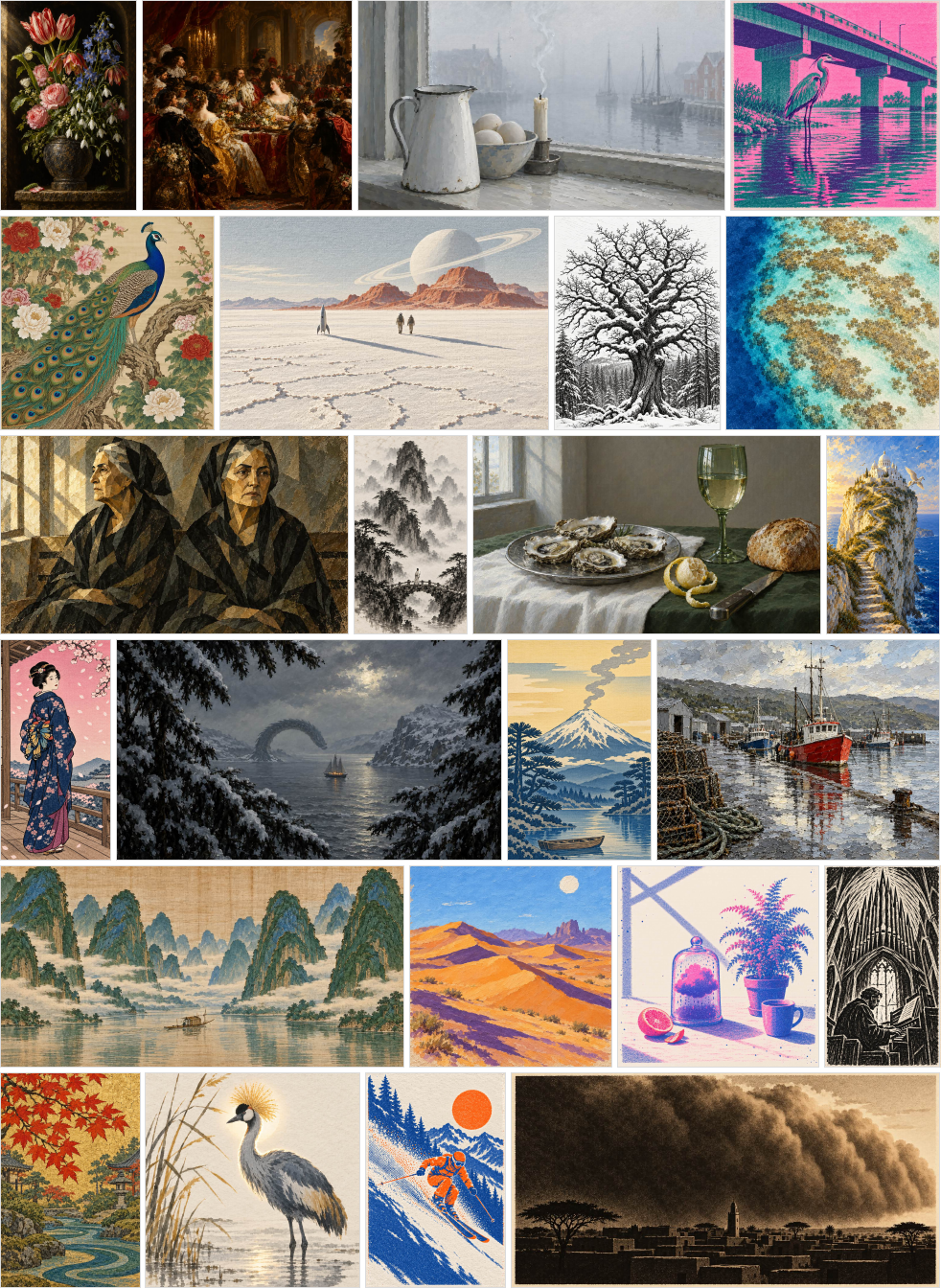}
  \caption{\textbf{Visual results of Repr-MoE on text-to-image generation: painting and printmaking.}}
  \label{fig:t2i-gallery-painting}
\end{figure}

\begin{figure}[p]
  \centering
  \includegraphics[width=\textwidth]{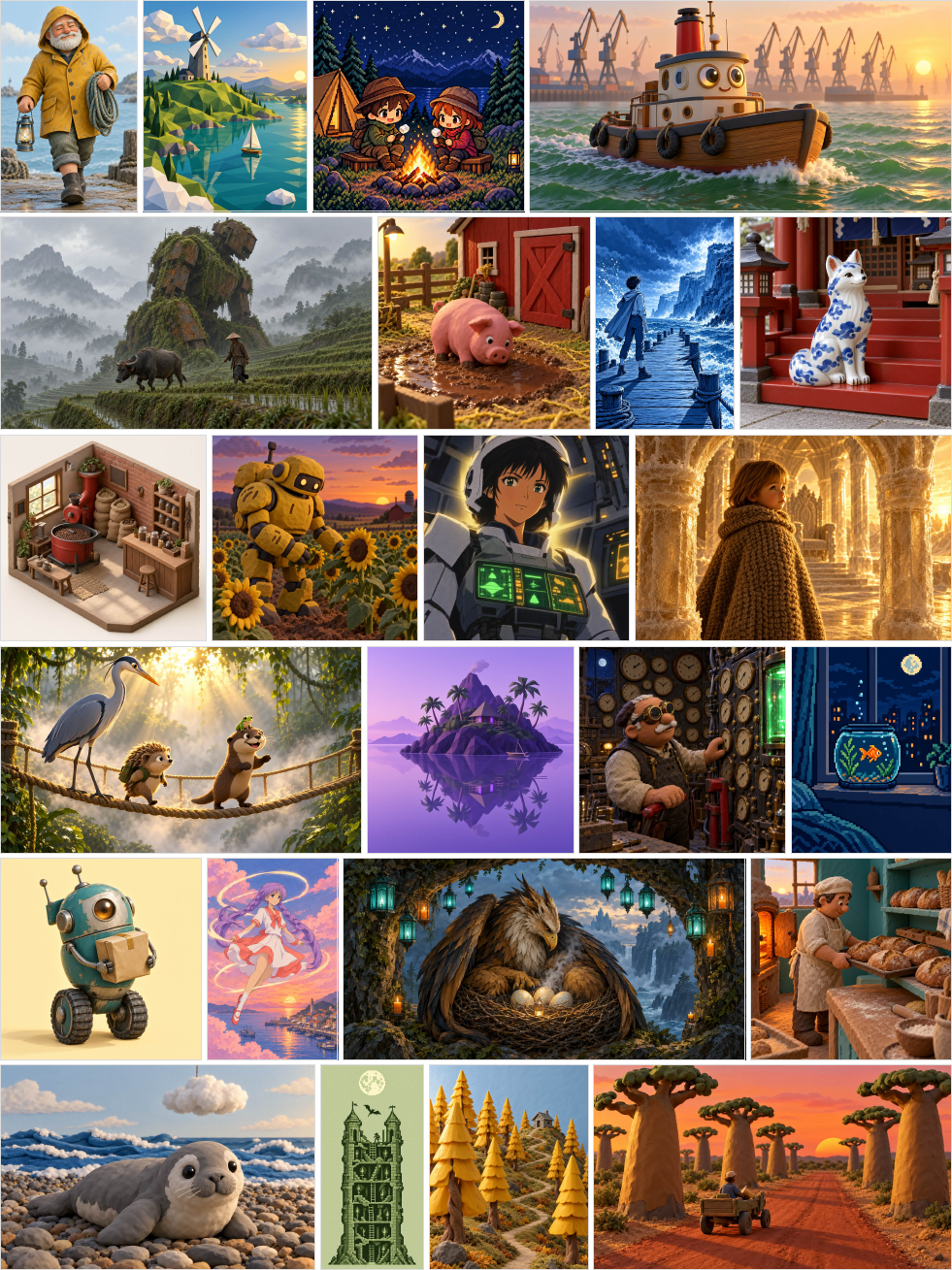}
  \caption{\textbf{Visual results of Repr-MoE on text-to-image generation: illustration and 3D rendering.}}
  \label{fig:t2i-gallery-illustration}
\end{figure}

\begin{figure}[p]
  \centering
  \includegraphics[width=\textwidth,height=\dimexpr\textheight-102pt\relax,keepaspectratio]{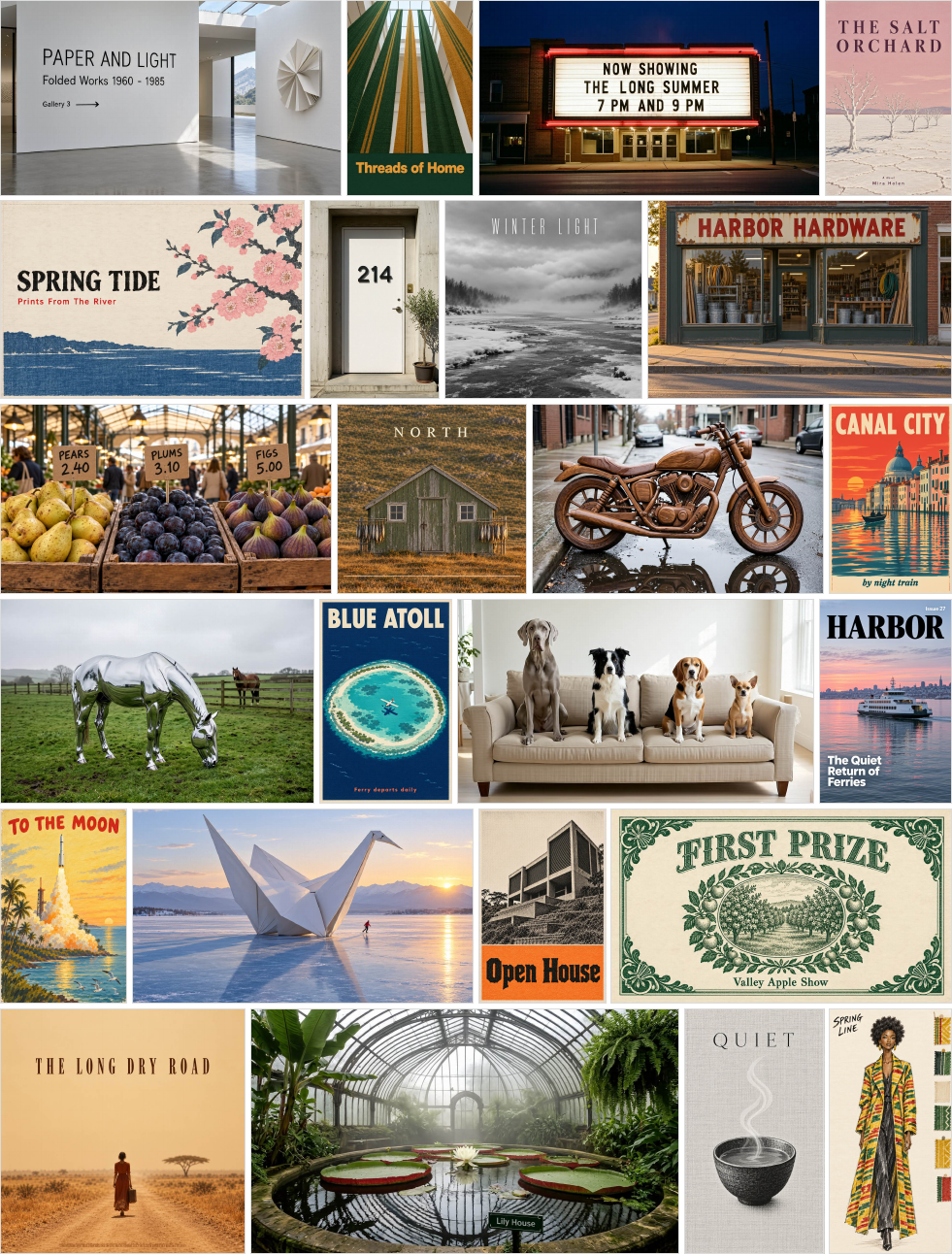}
  \caption{\textbf{Visual results of Repr-MoE on text-to-image generation: text rendering and prompt following.} The prompts specify the lettering. In reading order, the largest string requested for each image is ``PAPER AND LIGHT'', ``Threads of Home'', ``THE LONG SUMMER'', ``THE SALT ORCHARD'', ``SPRING TIDE'', ``214'', ``WINTER LIGHT'', ``HARBOR HARDWARE'', ``PEARS 2.40'' with ``PLUMS 3.10'' and ``FIGS 5.00'', ``NORTH'', ``CANAL CITY'', ``BLUE ATOLL'', ``HARBOR'', ``TO THE MOON'', ``Open House'', ``First Prize'', ``THE LONG DRY ROAD'', ``Lily House'', ``QUIET'', and ``SPRING LINE''. The four images without lettering come from prompts for a motorcycle made of walnut wood, a horse made of chrome, four dogs ordered by size, and a paper crane as tall as a house.}
  \label{fig:t2i-gallery-text}
\end{figure}

Fig.~\ref{fig:t2i-compare} compares Repr-MoE on QBench prompts with Qwen-Image-2.1, the highest-scoring latent model on QBench in Table~\ref{tab:t2i-headline}, and pixel-space AsymFLUX.2 klein.
Figs.~\ref{fig:t2i-gallery-places}--\ref{fig:t2i-gallery-text} show additional uncropped visual results of our pixel-space Repr-MoE (1.1B activated of 11.2B, Table~\ref{tab:t2i-headline}), grouped by subject and medium, at their generated aspect ratios using the sampling settings in Sec.~\ref{app:setup-t2i}.

\clearpage

\end{document}